%% file: acl_latex.tex
\documentclass[11pt]{article}

\usepackage[preprint]{acl}

\usepackage{times}
\usepackage{latexsym}
\usepackage{soul,color}
\usepackage{todonotes}
\usepackage[T1]{fontenc}

\usepackage[utf8]{inputenc}

\usepackage{microtype}

\usepackage{inconsolata}

\usepackage{graphicx}
\usepackage{url}
\usepackage{caption}
\usepackage{amsmath}
\usepackage{amsthm}
\usepackage{booktabs}
\usepackage{algorithm}
\usepackage{natbib}
\usepackage[switch]{lineno}
\usepackage{enumerate}
\usepackage{amssymb}
\usepackage{array}
\usepackage[noend]{algpseudocode}
\usepackage{multirow}
\usepackage{makecell}
\usepackage{subcaption}
\usepackage{adjustbox}
\usepackage[table]{xcolor}
\usepackage{colortbl}

\usepackage{xurl} 

\definecolor{lightblue}{RGB}{230,245,255}

\definecolor{SimBase}{RGB}{115,140,205} 
\definecolor{DivBase}{RGB}{215,135,135} 
\usepackage{pgfmath}
\usepackage{arydshln} 
\usepackage{enumitem}
\usepackage{tcolorbox}
\tcbuselibrary{skins,breakable}
\usepackage{xcolor}
\usepackage{pifont}
\newcommand{\cmark}{\textcolor{green!60!black}{\ding{51}}}
\newcommand{\xmark}{\textcolor{red!70!black}{\ding{55}}}

\newcommand{\MINSATSD}{10}
\newcommand{\MAXSATSD}{85}

\newcommand{\SIMLO}{0.40}
\newcommand{\SIMHI}{0.66}
\newcommand{\SIMGAMMA}{1.6}
\newcommand{\SIMCONTR}{1.8}

\newcommand{\DIVLO}{0.40}
\newcommand{\DIVHI}{0.65}
\newcommand{\DIVGAMMA}{1.2}
\newcommand{\DIVCONTR}{1.6}

\newcommand{\HeatCapCSD}[6]{
  \pgfmathsetmacro{\HX}{min(1,max(0,(#1-#2)/(#3-#2)))}%
  \pgfmathsetmacro{\HS}{min(1,max(0,0.5+(\HX-0.5)*#6))}%
  \pgfmathsetmacro{\HY}{pow(\HS,#5)}%
  \pgfmathtruncatemacro{\HP}{round(\MINSATSD+(\MAXSATSD-\MINSATSD)*\HY)}%
  \edef\HSpec{#4!\HP!white}%
  \expandafter\cellcolor\expandafter{\HSpec}%
}

\newcommand{\Sim}[1]{%
  \HeatCapCSD{#1}{\SIMLO}{\SIMHI}{SimBase}{\SIMGAMMA}{\SIMCONTR}#1%
}
\newcommand{\Div}[1]{%
  \pgfmathsetmacro{\Dinv}{\DIVHI-(#1)}%
  \HeatCapCSD{\Dinv}{0}{(\DIVHI-\DIVLO)}{DivBase}{\DIVGAMMA}{\DIVCONTR}#1%
}

\colorlet{SimHL}{SimBase!40!white}
\colorlet{DivHL}{DivBase!40!white}
\newcommand{\simhl}[1]{{\setlength{\fboxsep}{0.4pt}\colorbox{SimHL}{#1}}}
\newcommand{\divhl}[1]{{\setlength{\fboxsep}{0.4pt}\colorbox{DivHL}{#1}}}

\definecolor{REColor}{RGB}{115,140,205}
\definecolor{CVRColor}{RGB}{215,135,135}

\newcommand{\MINSATRE}{15}
\newcommand{\MAXSATRE}{85}

\newcommand{\RELO}{0.00}
\newcommand{\REHI}{1.20}
\newcommand{\CVRLO}{0.5}
\newcommand{\CVRHI}{1.00}

\newcommand{\REGAMMA}{2.1}
\newcommand{\RECONTR}{2.3}
\newcommand{\CVRGAMMA}{0.9}
\newcommand{\CVRCONTR}{5.0}

\newcommand{\HeatCapCRE}[6]{
  \pgfmathsetmacro{\HX}{min(1,max(0,(#1-#2)/(#3-#2)))}%
  \pgfmathsetmacro{\HS}{min(1,max(0,0.5+(\HX-0.5)*#6))}%
  \pgfmathsetmacro{\HY}{pow(\HS,#5)}%
  \pgfmathtruncatemacro{\HP}{round(\MINSATRE+(\MAXSATRE-\MINSATRE)*\HY)}%
  \edef\HSpec{#4!\HP!white}%
  \expandafter\cellcolor\expandafter{\HSpec}%
}

\newcommand{\RE}[1]{%
  \pgfmathsetmacro{\inv}{\REHI-(#1)}%
  \HeatCapCRE{\inv}{0}{(\REHI-\RELO)}{REColor}{\REGAMMA}{\RECONTR}#1%
}
\newcommand{\CVR}[1]{%
  \HeatCapCRE{#1}{\CVRLO}{\CVRHI}{CVRColor}{\CVRGAMMA}{\CVRCONTR}#1%
}
\colorlet{EffHL}{SimBase!30!white} 

\title{\textit{How You Ask Matters!} Adaptive RAG Robustness to Query Variations
}

\author{Yunah Jang$^{1}$, Megha Sundriyal$^{2}$, Kyomin Jung$^{1, \dagger}$ and Meeyoung Cha$^{2, \dagger}$ \\
  $^{1}$Seoul National University,
  $^{2}$Max Planck Institute for Security and Privacy, Germany
  \\
  \texttt{\{vn2209, kjung\}@snu.ac.kr}\\
  \texttt{\{megha.sundriyal, mia.cha\}@mpi-sp.org}
  }

\begin{document}
\maketitle

\begingroup
\renewcommand{\thefootnote}{\fnsymbol{footnote}}
\setcounter{footnote}{1}
\footnotetext[1]{Work done during an internship at Max Planck Institute for Security and Privacy.}
\footnotetext[2]{Co-corresponding authors.}
\endgroup

\begin{abstract}
\input{Text/1.Abstract}

\end{abstract}

\section{Introduction}
\input{Text/2.Introduction}

\section{Related Works}
\input{Text/3.RelatedWorks}

\section{Benchmark Construction}
\input{Text/4.DataConstruction}

\section{Experimental Setup}
\input{Text/5.Experimental_setup}

\section{Results}
\input{Text/6.Results}

\section{Human Query Analysis}
\input{Text/7.Analysis}
\label{sec:human_query_analysis}

\section{Conclusion}
\input{Text/8.Conclusion}

\newpage
\section*{Limitations}
\input{Text/9.Limitation}


\bibliography{custom}
\clearpage
\appendix

\input{Text/11.Appendix}

\end{document}

%% file: Text/1.Abstract.tex
Adaptive Retrieval-Augmented Generation (RAG) promises accuracy and efficiency by dynamically triggering retrieval only when needed and is widely used in practice.
However, real-world queries vary in surface form even with the same intent, and their impact on Adaptive RAG remains under-explored.
We introduce the first large-scale benchmark of diverse yet semantically identical query variations, combining human-written and model-generated rewrites. 
Our benchmark facilitates a systematic evaluation of Adaptive RAG robustness by examining its key components across three dimensions: answer quality, computational cost, and retrieval decisions.
We discover a critical robustness gap, where small surface-level changes in queries dramatically alter retrieval behavior and accuracy.
Although larger models show better performance, robustness does not improve accordingly.
These findings reveal that Adaptive RAG methods are highly vulnerable to query variations that preserve identical semantics, exposing a critical robustness challenge.

%% file: Text/2.Introduction.tex
Retrieval-Augmented Generation (RAG) reduces hallucinations in large language models (LLMs) by grounding outputs in retrieved evidence, thereby improving factual accuracy in downstream tasks~\cite{nearestneighbor, rag_seq2seq, selfrag}. 
While effective in reducing factual errors, conventional RAG often retrieves irrelevant or noisy documents, which can mislead reasoning and degrade answer quality~\cite{workingmemory,rowen, noise}. 

\begin{figure}[t!]
\centering
\includegraphics[width=\linewidth]{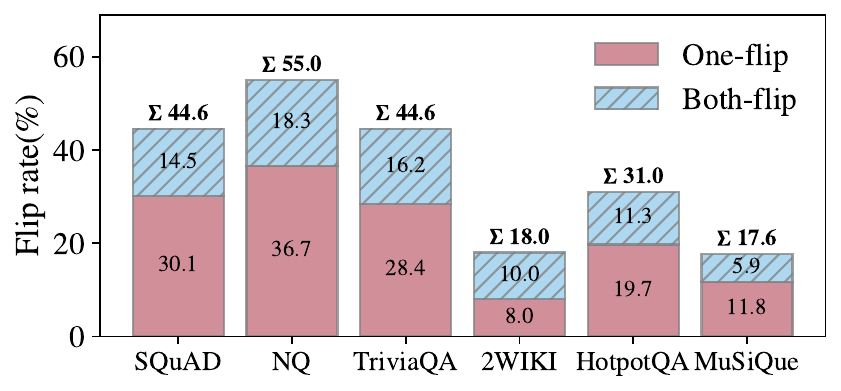}
\caption{
\textbf{Retrieval-decision flip rates} of the Qwen-32B model, measuring how often the model changes its retrieval judgment under meaning-preserving human query rewrites. 
Higher rates signal instability.
Flips are categorized as \textit{One-flip} (only one rewrite flips) and \textit{Both-flip} (both rewrites flip). 
}
\label{fig:preliminary}

\end{figure}

To address these limitations, Adaptive RAG dynamically decides \emph{when} and \emph{what} to retrieve based on the query and evolving context~\cite{adaptive_rag, knowing_youdontknow}.
If the model can answer confidently from parametric knowledge, it may skip retrieval and avoid interference from noisy documents.
While simple factoid questions may need only a few retrievals, multi-hop questions often benefit from iterative subqueries that accumulate evidence across hops~\cite{2wikimultihopqa, musique, ircot}.
Prior adaptive methods have therefore improved both accuracy and efficiency by allocating retrieval and LLM calls according to query difficulty and intermediate states~\cite{dragin, flare, rowen}.
These gains have driven growing adoption of Adaptive RAG in real-world applications~\cite{agenticrag_suvey}.

\begin{figure*}[t!]
\centering
\includegraphics[width=\linewidth]{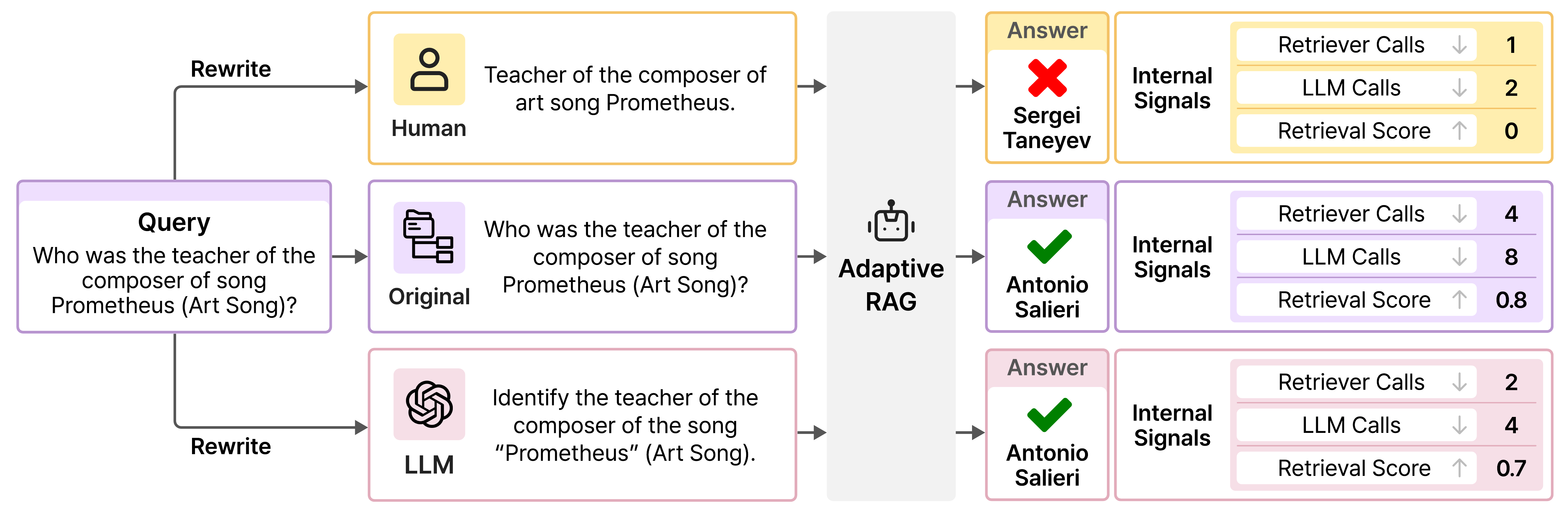}
\caption{\textbf{Example of Adaptive RAG responses} under human, original, and LLM-generated query rewrites, highlighting differences in answer correctness, computation overhead, and retrieval score.}
\label{fig:intro_figure}
\end{figure*}

In real-world settings, users express the same intent in diverse ways, including paraphrases, stylistic variations, and typos or other noise~\cite{robust_research, qerag, out_of_style}.
To assess decision stability, we measure how often the model’s retrieval decision (retrieve or not) flips between the original query and two human rewrites.
As shown in Figure~\ref{fig:preliminary}, human rewrites cause decision flips in up to 55\% of cases, revealing severe instability in retrieval decisions.
Despite this, robustness to real-world query variation remains underexplored, which is especially concerning in high-stakes settings where factual correctness and user trust are critical~\cite{rag_domainspecific, rag_systematic}.

This paper presents the first empirical study of Adaptive RAG robustness under realistic query variations.
To facilitate this, we introduce a benchmark spanning controlled linguistic paraphrases and naturally written human queries.
We evaluate on six widely used question-answering (QA) datasets covering both single-hop and multi-hop questions and analyze how surface-level differences affect both intermediate retrieval trajectories and final answers (Figure~\ref{fig:intro_figure}).
For each question, we include six model-generated variants and two human rewrites, yielding 27K QA pairs spanning diverse levels of complexity.
Across these query variations, we comprehensively assess model behavior along three dimensions: answer quality, computational cost, and retrieval decisions. This work presents the following key findings:\footnote{The sample dataset and code are provided as supplementary materials. We will release them publicly upon acceptance.}
\begin{itemize}
    \item Improved performance does not imply greater robustness; even strong baselines show substantially greater inconsistency in answer semantics and computational cost.
    \item Comprehensive experiments reveal that all evaluated methods and models are consistently vulnerable to both human rewrites and spelling errors.
    \item Semantically identical queries with different surface forms can trigger different intermediate trajectories,  subquery generation, retrieval, and the final answer.
\end{itemize}

%% file: Text/3.RelatedWorks.tex
\paragraph*{Adaptive RAG.}
Adaptive RAG implements dynamic, model-driven decisions about when to retrieve and what to retrieve~\cite{rowen}. 
We broadly categorize prior methods into two main streams: logit-based and generation-based methods. 
Logit-based methods monitor model confidence and trigger retrieval when the model appears uncertain, for example, using token-level probabilities as in~\citet{flare}  or uncertainty estimates based on token probabilities and attention weights as in~\citet{dragin}.
Generation-based methods, on the other hand, let the model decide on the fly when it needs retrieval by generating search queries during the generation process~\cite{searcho1}. 
Our work focuses on approaches that use a single model with a retriever and do not rely on external classifiers, such as \citet{adaptive_rag, knowing_youdontknow}.

\input{Tables/examples_main}

\paragraph*{Query Robustness.}
Prior work studied the robustness of RAG systems. More broadly, traditional IR research has shown that query variations can affect retrieval rankings ~\cite{QUERYVARIATION_8, QUERYVARIATION11, QUERYVARIATION12}, compromise evaluation reliability ~\cite{QUERYVARIATION_9}, and reveal substantial human query variability ~\cite{QUERYVARIAITON_10_uqv100, QUERYVARIATION13}.

In the context of QA and RAG, prior studies have demonstrated that linguistic perturbations, including query-entry errors, grammatical errors, typos, and stylistic variations, can noticeably degrade system performance~\cite{out_of_style, query_perturbation_gem, rare}. 
Collectively, these results indicate that even small modifications to a query may lead to substantial changes in retrieved evidence and downstream answers~\cite{qerag, robust_research, User_variability}.
Despite these advances, existing work has two important limitations: (1) it primarily considers conventional RAG, without addressing the added complexity of adaptive RAG; and (2) it does not investigate the impact of real human query reformulations across end-to-end RAG frameworks.

%% file: Tables/examples_main.tex
\begin{table*}[t!]
\centering
\small
\setlength{\tabcolsep}{6pt}
\renewcommand{\arraystretch}{1.1}

\begin{tabular}{
  >{\centering\arraybackslash}m{0.07\textwidth}
  >{\centering\arraybackslash}m{0.15\textwidth}
  m{0.68\textwidth}
}
\specialrule{1.2pt}{1.5pt}{1.5pt}
\rowcolor{lightblue}
\multicolumn{3}{c}{
  \textbf{Original Query:} Who is the child of the director of film Mukhyamantri (1996 Film)?
} \\
\specialrule{1.2pt}{1.5pt}{1.5pt}

\textbf{Category} & \textbf{Variation} & \textbf{Sample Rewrites} \\
\midrule

\multirow[c]{2}{*}[-0.35ex]{\parbox[c]{0.065\textwidth}{\centering\footnotesize\bfseries Style}}
& Formality\,{\scriptsize (Form.)}\,$\downarrow$
& Yo, quick one — who’s the kid of the director behind the 1996 flick Mukhyamantri? \\
\cmidrule{2-3}
& Readability\,{\scriptsize (Read.)}\,$\downarrow$
& By whom is the progeny of the director of Mukhyamantri (1996 Film) constituted? \\
\midrule

\multirow[c]{2}{*}[-0.35ex]{\parbox[c]{0.065\textwidth}{\centering\footnotesize\bfseries Type}}
& Declarative {\scriptsize (Decl.)}
& I want to know who is the child of the director of the film Mukhyamantri (1996 Film).\\
\cmidrule{2-3}
& Imperative {\scriptsize (Impr.)}
& Name the child of the director of film Mukhyamantri (1996 Film). \\
\midrule

\multirow[c]{2}{*}[-0.35ex]{\parbox[c]{0.065\textwidth}{\centering\footnotesize\bfseries Error}}
& Spelling\, {\scriptsize (Spell.)}\,$\downarrow$
& Who is ght childer ar the directer of filme Mukhyamantri (1996 Film )?\\
\cmidrule{2-3}
& Grammar\, {\scriptsize (Gram.)}\,$\downarrow$
& Who are the childs of the director from film Mukhyamantri 1996? \\
\bottomrule
\end{tabular}

\caption{\textbf{Examples of query rewrites} grouped by category, illustrating stylistic, structural, and error-based variations. $\downarrow$ denotes reduction in corresponding writing variation.}
\label{tab:query_rewrites}
\end{table*}

%% file: Text/4.DataConstruction.tex
\subsection{Dataset}
We evaluate on six QA datasets: Natural Questions (NQ)~\cite{natural_questions}, TriviaQA~\cite{triviaqa}, and SQuAD v1.1~\cite{squad} for single-hop QA, and 2WikiMultiHopQA (2WIKI)~\cite{2wikimultihopqa}, MuSiQue~\cite{musique}, and HotpotQA~\cite{hotpotqa} for multi-hop QA.

\subsection{Query Variation Generation}
We design a taxonomy of machine-generated queries that captures realistic query variation while remaining interpretable and controllable for diagnostic analysis.
It comprises three categories: sentence style, sentence type, and errors, as shown in Table~\ref{tab:query_rewrites}.
While prior work has mainly studied style and error variations~\cite{qerag, query_perturbation_gem, out_of_style}, we additionally incorporate sentence type variations, yielding 21k queries.
Following prior work~\cite{adarague, adaptive_rag, ircot}, we evaluate all systems on the same 500-sample set using GPT-5-mini, with prompts in Appendix~\ref{app:prompts} and validation details in Appendix~\ref{app:query details}.

\paragraph{Sentence Style.}
Human queries exhibit diverse linguistic styles, often characterised by less formality, simplified language, or omission of polite expressions~\cite{web_query, web_query2}.
Following prior work~\cite{out_of_style}, we construct natural query variations with reduced formality and readability.
We define the two variation types as follows:

\begin{itemize}
\item \textbf{Formality}: We generate informal versions using simpler vocabulary, contractions, and more casual expressions. 
\item \textbf{Readability}: We reduce the readability of the queries, making them more difficult to understand. Although this may sacrifice clarity, it more accurately reflects how users often communicate in less structured contexts. \end{itemize}

\paragraph{Sentence Type.}
Sentence type variations capture the grammatical form of a query, reflecting how it is structured.
They are commonly categorized as interrogative, imperative, and declarative~\cite{sentence_type}.
Because most QA datasets mainly consist of interrogative questions, especially wh-questions (e.g., who, what, when)~\cite{qa_dataset, declartive_detection}, we additionally construct declarative and imperative variants beyond the original interrogative form.

\begin{itemize}
\item \textbf{Declarative sentence.} These queries are phrased as statements or assertions, often expressing knowledge, desire, or intention.
\item \textbf{Imperative sentence.} These queries are phrased as commands or requests, typically used when the user wants the system to perform a task.
\end{itemize}

\paragraph{Error.}
We focus on two common error types in human-generated queries: spelling and grammatical errors~\cite{error_typo_grammar, qerag}.
Such errors often arise from hurried typing, typographical mistakes, and limited command of grammar, and can affect retrieval performance~\cite{misspelling1, misspelling2, misspelling3}.

\begin{itemize}
\item \textbf{Spelling errors.} We inject spelling errors into the original queries using the Python library \textit{nlpaug}\footnote{\href{https://pypi.org/project/nlpaug/0.0.5/}{pypi.org/project/nlpaug/0.0.5}}.
\item \textbf{Grammatical errors.} We generate grammatical errors by prompting LLM to introduce grammatical mistakes. We then use LanguageTool to verify that each generated query contains at least one grammatical error.
\end{itemize}

\subsection{Human Rewrite Annotation}
Human queries vary substantially in form and often combine multiple variation types~\cite{queryrobustness_human, dmqr_human, human_vocab}. 
Yet existing robustness evaluations of end-to-end RAG systems largely overlook this dimension by excluding human-written queries~\cite{out_of_style, query_perturbation_gem, qerag}.
We therefore collect two human rewrites for each of 600 queries (100 per dataset across six QA benchmarks), yielding 1,200 rewrites from 12 annotators and 22.2k total queries in our benchmark.
Detailed guidelines are provided in Appendix~\ref{app:human_annotation_details}.

%% file: Text/5.Experimental_setup.tex
\subsection{Methods}
We consider two representative Adaptive RAG paradigms: logit-based control (LB) and generation-based control (GB). 
For LB, we adopt DRAGIN~\cite{dragin}, which determines when and what to retrieve from token-level signals such as probabilities and attention weights. 
For GB, we adopt Search-o1~\cite{searcho1}, which performs retrieval planning during generation, deciding whether to search, how to formulate queries, and how to generate the final answer. 
Unlike LB, which relies on threshold-based control, GB handles retrieval control end-to-end without hand-tuned thresholds or auxiliary classifiers.

\captionsetup[subfigure]{font=footnotesize}
\begin{figure*}[t]
  \centering

  \includegraphics[width=0.65\textwidth]{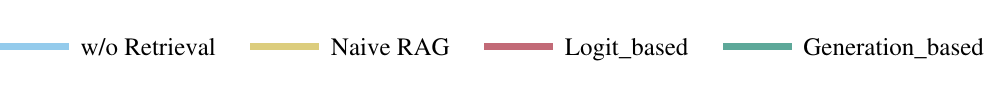}
  \vspace{-0.4em}

  \begin{subfigure}[t]{0.24\textwidth}
    \centering
    \includegraphics[width=\linewidth]{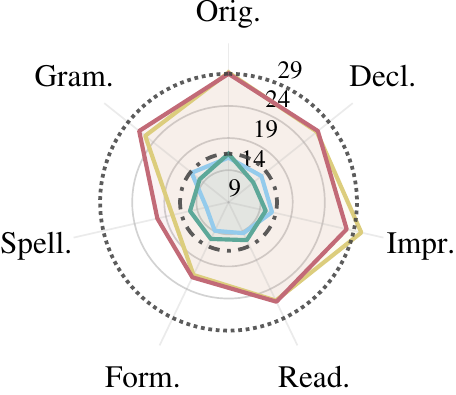}
    \caption{SQuAD / Llama-3.1}
  \end{subfigure}\hfill
  \begin{subfigure}[t]{0.24\textwidth}
    \centering
    \includegraphics[width=\linewidth]{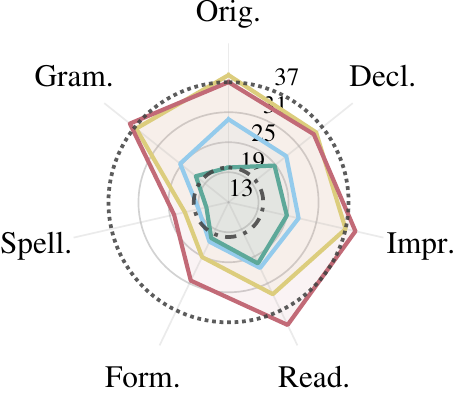}
    \caption{2WIKI / Llama-3.1}
  \end{subfigure}\hfill
  \begin{subfigure}[t]{0.24\textwidth}
    \centering
    \includegraphics[width=\linewidth]{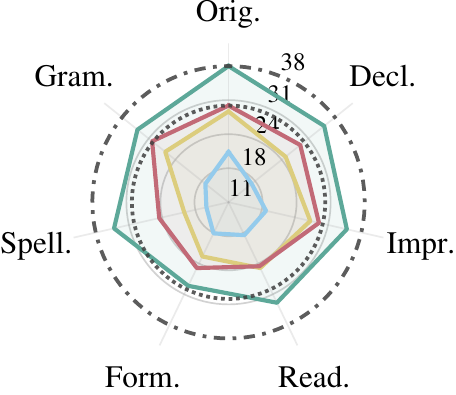}
    \caption{SQuAD / QwQ}
  \end{subfigure}\hfill
  \begin{subfigure}[t]{0.24\textwidth}
    \centering
    \includegraphics[width=\linewidth]{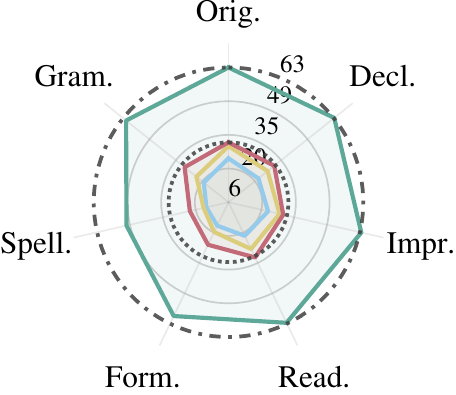}
    \caption{2WIKI / QwQ}
  \end{subfigure}
  \caption{\textbf{InAccuracy results} across six perturbation types for a given dataset / model pair.}
  \label{fig:inaccuracy_four}
\end{figure*}

\subsection{Metrics}

Adaptive RAG involves three main components: retrieval decisions, LLM/retrieval calls, and final answer generation.
Accordingly, we analyze all three aspects to obtain a thorough understanding of system robustness. 

\paragraph{Answer Robustness.}
We evaluate final answer quality using InAccuracy, following \citet{inacc, adarague}, which tests whether the gold answer appears in the generated answer.
We also assess semantic consistency across query variations using similarity and diversity.

\begin{enumerate}
\item \textbf{InAccuracy.}
For each instance $i$ and perturbation type $v$, let $\hat{a}_{i,v}$ be the generated answer and $g_i$ the gold answer.
We report the dataset-level mean for each perturbation type $v$:
\begin{equation*}
\begin{aligned}
\mathrm{InAcc}_{v}=\frac{1}{N}\sum_{i=1}^{N}\mathbf{1}\!\left[g_i \subseteq \hat{a}_{i,v}\right]
\end{aligned}
\end{equation*}

\item \textbf{Similarity.}
Let $\phi(\cdot)$ denote an embedding function (SBERT), and let $e_{i,v}=\phi(\hat{a}_{i,v})$.
We compute the cosine similarity between the answer to the original query, $\hat{a}_{i,\mathrm{orig}}$, and that to variant $v$, $\hat{a}_{i,v}$:
\begin{equation*}
\begin{aligned}
\mathrm{Sim}_{v}=\frac{1}{N}\sum_{i=1}^{N}\cos(e_{i,v},e_{i,\mathrm{orig}})
\end{aligned}
\end{equation*}

\item \textbf{Diversity (Div.).}
Let $\mathcal{V}$ be the set of query variants for instance $i$, including the original, and let $K=|\mathcal{V}|$.
We define answer spread as
\begin{gather*}
\mathrm{Div}_{i}=\frac{2}{K(K-1)}\sum_{v<w}\bigl(1-\cos(e_{i,v},e_{i,w})\bigr)\\
\mathrm{Div}=\frac{1}{N}\sum_{i=1}^{N}\mathrm{Div}_{i}
\end{gather*}
\end{enumerate}

\paragraph{Computation Robustness.}
In adaptive RAG, computation is measured by call counts, namely retrieval rounds and LLM invocations~\cite{adarague, dragin}.
We evaluate robustness for both the retriever and the LLM in terms of (i) cost changes relative to the original query and (ii) cost consistency across query variations.
Let $c_{i,v}$ denote the call count for instance $i$ under variant $v$, where $v=0$ denotes the original query.

\begin{enumerate}
\item \textbf{Relative Error (RE).}
For each perturbation type $v$, we compute the deviation of each instance $i$ from the original-query call count and report the dataset-level mean.
\begin{equation*}
\mathrm{RE}_{v}=\frac{1}{N}\sum_{i=1}^{N}\mathrm{RE}_{i,v} = \frac{1}{N}\sum_{i=1}^{N} \frac{\left|c_{i,v}-c_{i,0}\right|}{\max(1,c_{i,0})}
\end{equation*}

\item \textbf{Coefficient-of-Variation Robustness (CVR).}
Let $\mathcal{V}$ be the set of variants, including the original, and let $r_i=(c_{i,v})_{v\in\mathcal{V}}$.
We define instance-wise variability as
\begin{equation*}
\mathrm{CV}_{i}=\frac{\mathrm{std}(r_i)}{\mathrm{mean}(r_i)},\quad
\mathrm{CVR}=\frac{1}{N}\sum_{i=1}^{N}\frac{1}{1+\mathrm{CV}_{i}}
\end{equation*}
\end{enumerate}

\noindent
We report RE by perturbation type and CVR over all perturbations for both \textbf{retriever calls} and \textbf{LLM calls}.

\input{Tables/Robustness_ACC_deatailed}
\input{Tables/Robustness_efficiency}

\paragraph{Retrieval Decision Robustness.}
We assess whether the system retrieves only when external evidence is needed under query variation.
Let $\textit{act}$ denote whether retrieval is triggered ($\textit{act}=1$ if triggered, $\textit{act}=0$ otherwise).
We define $\textit{need}$ using a no-retrieval baseline: if the model answers correctly without retrieval, then $\textit{need}=0$; otherwise, $\textit{need}=1$.
Following~\citet{adarague}, we define:

\begin{itemize}
\item \textbf{Overconfidence.} Retrieval is needed but not triggered ($\textit{need}=1, \textit{act}=0$).
\item \textbf{Underconfidence.} Retrieval is triggered although it is not needed ($\textit{need}=0, \textit{act}=1$).
\end{itemize}

\subsection{Implementation Details}
We use two LLMs, QwQ-32B~\cite{qwq32b} and Llama 3.1-8B-Instruct~\cite{grattafiori2024llama}, both widely used in Adaptive RAG.
For retrieval, we employ BM25~\cite{bm25} as the sparse retriever over a Wikipedia-based document pool following~\citet{adarague}.
We also use Contriever~\cite{contriever} to test whether the same vulnerability persists with dense retrieval.
Appendix~\ref{app:implementation_details} provides further details and explains our model choices.
\input{Tables/dense_retriever_main}

%% file: Tables/Robustness_ACC_deatailed.tex
\begin{table}[t]
\centering
\scriptsize
\setlength{\tabcolsep}{3.0pt}
\renewcommand{\arraystretch}{1.08}

\begin{adjustbox}{width=\columnwidth,center}
\begin{tabular}{l l cccccc c}
\toprule[1.2pt]
\multirow{2}{*}{\textbf{Dataset}} &
\multirow{2}{*}{\textbf{Mtd.}} &
\multicolumn{6}{c}{\textbf{Similarity} $\uparrow$} &
\multirow{2}{*}{\textbf{Div.} $\downarrow$}\\
\cmidrule(lr){3-8}
& & Form. & Read. & Decl. & Impr. & Spell. & Gram. & \\
\midrule

\multicolumn{9}{l}{\textbf{\textit{Llama-3.1}}} \\
\addlinespace[2pt]

\multirow{2}{*}{SQuAD}

& LB & \Sim{0.551} & \Sim{0.539} & \Sim{0.593} & \textbf{\Sim{0.634}} & \Sim{0.552} & \Sim{0.599} & \Div{0.469} \\
& GB & \Sim{0.426} & \Sim{0.408} & \textbf{\Sim{0.433}} & \Sim{0.430} & \Sim{0.411} & \Sim{0.414} & \Div{0.585} \\
\addlinespace[3pt]

\cdashline{1-9}
\addlinespace[3pt]

\multirow{2}{*}{2WIKI}

& LB & \Sim{0.560} & \Sim{0.628} & \Sim{0.649} & \textbf{\Sim{0.655}} & \Sim{0.537} & \Sim{0.627} & \Div{0.414} \\
& GB & \Sim{0.423} & \textbf{\Sim{0.458}} & \Sim{0.300} & \Sim{0.403} & \Sim{0.313} & \Sim{0.385} & \Div{0.667} \\

\midrule

\multicolumn{9}{l}{\textbf{\textit{QwQ}}} \\
\addlinespace[2pt]

\multirow{2}{*}{SQuAD}
& LB & \Sim{0.573} & \Sim{0.579} & \Sim{0.615} & \textbf{\Sim{0.631}} & \Sim{0.544} & \Sim{0.584} & \Div{0.440} \\
& GB & \Sim{0.552} & \Sim{0.558} & \Sim{0.579} & \textbf{\Sim{0.589}} & \Sim{0.552} & \Sim{0.579} & \Div{0.459} \\
\addlinespace[3pt]

\cdashline{1-9}
\addlinespace[3pt]

\multirow{2}{*}{2WIKI}

& LB & \Sim{0.500} & \Sim{0.545} & \Sim{0.551} & \textbf{\Sim{0.565}} & \Sim{0.462} & \Sim{0.519} & \Div{0.511} \\
& GB & \Sim{0.520} & \Sim{0.557} & \Sim{0.554} & \textbf{\Sim{0.559}} & \Sim{0.499} & \Sim{0.548} & \Div{0.465} \\

\bottomrule[1.2pt]
\end{tabular}
\end{adjustbox}

\caption{
\textbf{Answer similarity and diversity} across query variations. 
Darker shades indicate more desirable outcomes: higher \simhl{similarity} and lower \divhl{diversity}. We compare the results between logit-based (LB) and generation-based (GB) method.}
\label{tab:robustness_acc_detailed}
\end{table}

%% file: Tables/Robustness_efficiency.tex
\begin{table*}[t]
\centering
\resizebox{\textwidth}{!}{
\begin{tabular}{%
p{0.5cm}
p{1.3cm}
c
ccccccc @{\hspace{1.4em}} ccccccc}
\toprule[1.7pt]

 &  &  &
 \multicolumn{7}{c}{\textbf{SQuAD}}
 & \multicolumn{7}{c}{\textbf{2WIKI}} \\
\cmidrule(lr){4-10}\cmidrule(lr){11-17}

 & \multirow{2}{*}{\textbf{Target}}
 & \multirow{2}{*}{\textbf{Mtd.}}
 & \multicolumn{6}{c}{RE $\downarrow$} & \multirow{2}{*}{CVR $\uparrow$}
 & \multicolumn{6}{c}{RE $\downarrow$} & \multirow{2}{*}{CVR $\uparrow$} \\
\cmidrule(lr){4-9}\cmidrule(lr){11-16}

 &  &  & Form. & Read. & Decl. & Impr. & Spell. & Gram.
    &  & Form. & Read. & Decl. & Impr. & Spell. & Gram. & \\
\midrule

\multirow{2}{*}{\rotatebox{90}{\parbox[c]{1.9cm}{\textbf{Llama-3.1}}}}
 & \multirow{2}{*}{\parbox[c]{1.3cm}\centering \textbf{RTR}}
 & LB & \RE{0.472} & \RE{0.592} & \RE{0.416} & \textbf{\RE{0.338}} & \RE{0.431} & \RE{0.439} & \CVR{0.734}
      & \RE{0.853} & \RE{0.859} & \textbf{\RE{0.455}} & \RE{0.846} & \RE{0.928} & \RE{0.941} & \CVR{0.710} \\
 &  & GB & \RE{0.092} & \RE{0.094} & \textbf{\RE{0.086}} & \RE{0.090} & \RE{0.100} & \RE{0.092} & \CVR{0.937}
      & \RE{0.468} & \RE{0.416} & \textbf{\RE{0.402}} & \RE{0.542} & \RE{0.496} & \RE{0.435} & \CVR{0.742} \\
\cmidrule(lr){2-17}
 & \multirow{2}{*}{\textbf{LLM}}
 & LB & \RE{0.441} & \RE{0.592} & \RE{0.396} & \textbf{\RE{0.313}} & \RE{0.416} & \RE{0.411} & \CVR{0.767}
      & \RE{1.001} & \RE{1.001} & \textbf{\RE{0.427}} & \RE{1.007} & \RE{1.093} & \RE{1.130} & \CVR{0.742} \\
 &  & GB & \RE{0.056} & \RE{0.059} & \textbf{\RE{0.048}} & \RE{0.055} & \RE{0.061} & \RE{0.051} & \CVR{0.963}
      & \RE{0.447} & \RE{0.404} & \textbf{\RE{0.401}} & \RE{0.520} & \RE{0.477} & \RE{0.418} & \CVR{0.745} \\
      
\midrule
\multirow{2}{*}{\rotatebox{90}{\parbox[c]{1.5cm}{\textbf{QwQ}}}}
 & \multirow{2}{*}{\textbf{RTR}}
 & LB & \RE{0.494} & \RE{0.450} & \textbf{\RE{0.405}} & \RE{0.415} & \RE{0.495} & \RE{0.441} & \CVR{0.740}
      & \RE{0.560} & \RE{0.502} & \RE{0.446} & \textbf{\RE{0.443}} & \RE{0.565} & \RE{0.511} & \CVR{0.719} \\
 &  & GB & \RE{0.360} & \RE{0.322} & \textbf{\RE{0.298}} & \RE{0.315} & \RE{0.361} & \RE{0.330} & \CVR{0.762}
      & \RE{0.788} & \RE{0.808} & \textbf{\RE{0.780}} & \RE{0.825} & \RE{0.787} & \RE{0.884} & \CVR{0.689} \\
\cmidrule(lr){2-17}
 & \multirow{2}{*}{\textbf{LLM}}
 & LB & \RE{0.477} & \RE{0.466} & \textbf{\RE{0.389}} & \RE{0.399} & \RE{0.491} & \RE{0.424} & \CVR{0.760}
      & \RE{0.547} & \RE{0.481} & \textbf{\RE{0.426}} & \RE{0.430} & \RE{0.565} & \RE{0.493} & \CVR{0.747} \\
 &  & GB & \RE{0.265} & \RE{0.230} & \textbf{\RE{0.225}} & \RE{0.238} & \RE{0.250} & \RE{0.243} & \CVR{0.858}
      & \RE{1.078} & \RE{1.142} & \RE{1.092} & \RE{1.203} & \textbf{\RE{1.049}} & \RE{1.219} & \CVR{0.716} \\
\bottomrule[1.7pt]
\end{tabular}
}
\caption{
\textbf{Computation robustness} across different model-generated query variations.
Darker shading indicates more desirable outcomes: lower \simhl{RE} and higher \divhl{CVR}. We compare the results between logit-based (LB) and generation-based (GB) method. RTR stands for retriever and bold denotes maximum across the variations.
}
\label{tab:robustness_efficiency}
\end{table*}

%% file: Tables/dense_retriever_main.tex
\begin{table}[t]
\centering
\small
\renewcommand{\arraystretch}{1.15}
\setlength{\tabcolsep}{5pt}

\begin{tabular}{llccc}
\toprule[1.2pt]
\textbf{Model} & \textbf{Mtd.} & \textbf{Orig.} & \textbf{Human} & \textbf{Spell.} \\
\midrule
Llama-3.1-8B & LB & \textbf{37.6} & 17.9 & 23.0 \\
Llama-3.1-8B & GB & \textbf{ 20.2} & 12.1 & 9.8 \\
QwQ-32B & LB &  \textbf{26.6} & 20.0 & 21.0 \\
QwQ-32B & GB &  \textbf{59.2} & 43.4 & 45.8 \\
\bottomrule[1.2pt]
\end{tabular}
\caption{\textbf{Contriever (dense retriever) results.} We report accuracy only for Original, Human, and Spelling variation under logit-based (LB) and generation-based (GB) methods.}
\label{tab:dense_retriever_main}
\end{table}

%% file: Text/6.Results.tex
We present results on two representative datasets, SQuAD (single-hop) and 2WIKI (multi-hop), and report results on the remaining datasets in the Appendix.

\subsection{Answer Robustness}
\paragraph{Model performance drops substantially under query variations.}
Figure~\ref{fig:inaccuracy_four} shows that spelling errors yield the largest accuracy drop across dataset--model pairs, highlighting strong sensitivity to surface-form noise.
Generation-based methods benefit from larger models and more challenging datasets.
For example, on 2WIKI with QwQ, accuracy declines from 63.4\% on the original queries to 50.6\% under spelling errors, with drops of up to 12.8\% across variants.
Results for all six datasets are reported in Appendix~\ref{app:answer_robustness}.

\paragraph{Generation-based methods yield less consistent outputs.}
Table~\ref{tab:robustness_acc_detailed} confirms the same trend: imperative rewrites achieve the highest answer similarity to the original, whereas spelling errors consistently yield the lowest similarity.
Generation-based methods also show higher diversity (lower is better), suggesting less stable outputs across perturbations.

\paragraph{Dense retriever shows performance degradation.}
We further examine whether the vulnerability of Adaptive RAG to query variations persists under dense retrieval using Contriever.
Table~\ref{tab:dense_retriever_main} presents the accuracy results on 2WIKI. We observe substantial performance degradation for both human-written and spelling-error queries, indicating that the robustness gap persists in semantic retrieval as well.
The full results are provided in Table~\ref{tab:contriever_results}.

\subsection{Computation Robustness}

\begin{figure}[t!]
\centering
\hspace*{8mm}\includegraphics[width=0.95\linewidth]{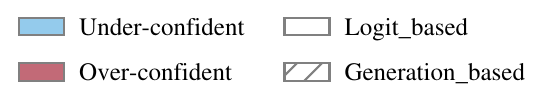}
\includegraphics[width=0.8\linewidth]{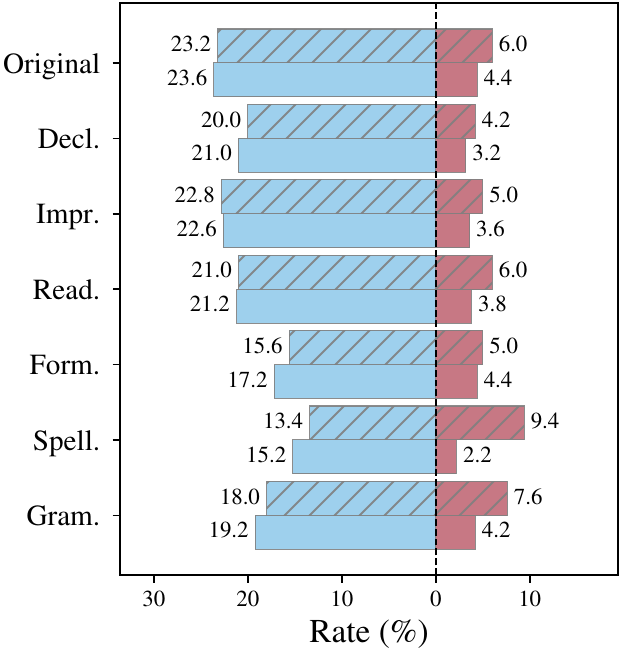}
\caption{\textbf{Under- and over-confident rates} on model query variations.}

\label{fig:under_over_2wiki}
\end{figure}

\begin{figure*}[t!]
  \centering
  \includegraphics[width=\linewidth]{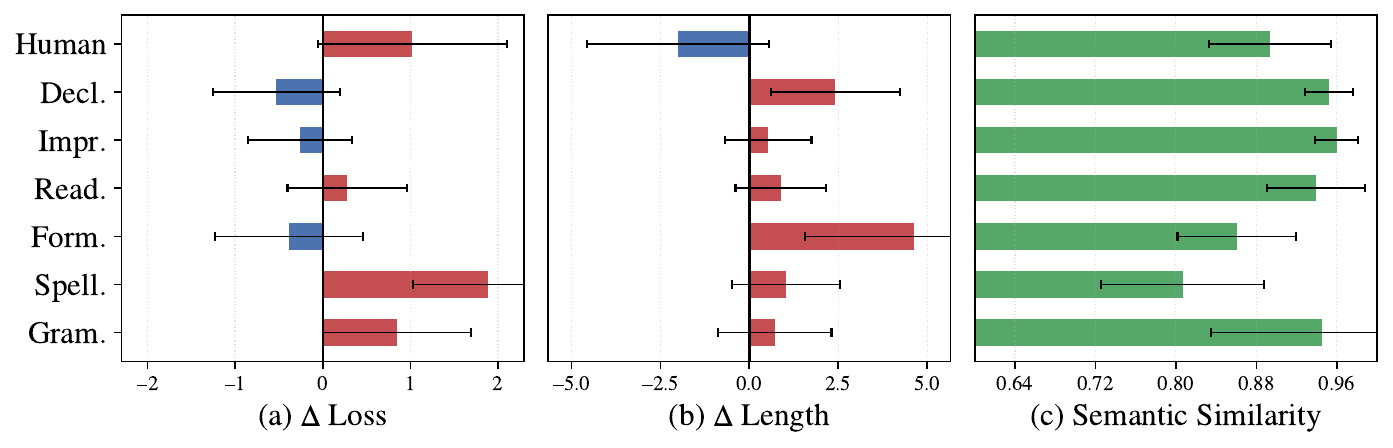}
  \caption{\textbf{Query analysis.} 
  From left to right: (a) change in LM loss relative to the original query ($\Delta$Loss; nats/token), (b) change in query length ($\Delta$Length; words), and (c) semantic similarity to the original query (cosine similarity). 
  }
  \label{fig:query_analysis-2wikimultihopqa}
\end{figure*}

\captionsetup[subfigure]{font=footnotesize}
\begin{figure}[t]
  \centering
  \hspace*{7mm}\includegraphics[width=0.36\textwidth]{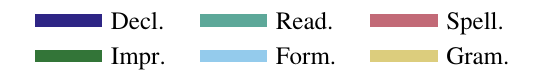}
  \begin{subfigure}[t]{0.24\textwidth}
    \centering
    \includegraphics[width=\linewidth]{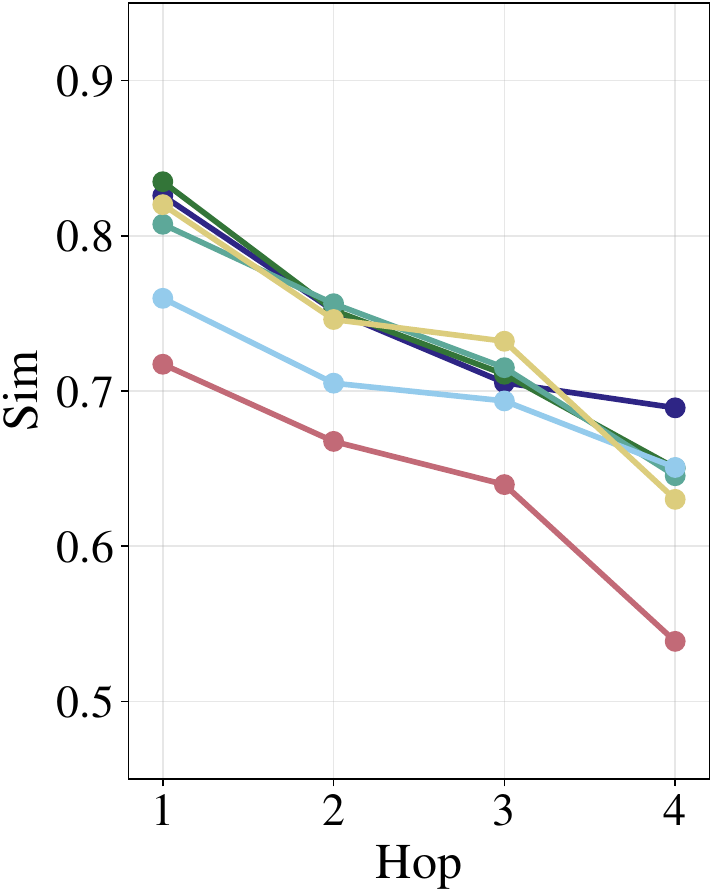}
    \caption{Subquery Drift}
  \end{subfigure}\hfill
  \begin{subfigure}[t]{0.24\textwidth}
    \centering
    \includegraphics[width=\linewidth]{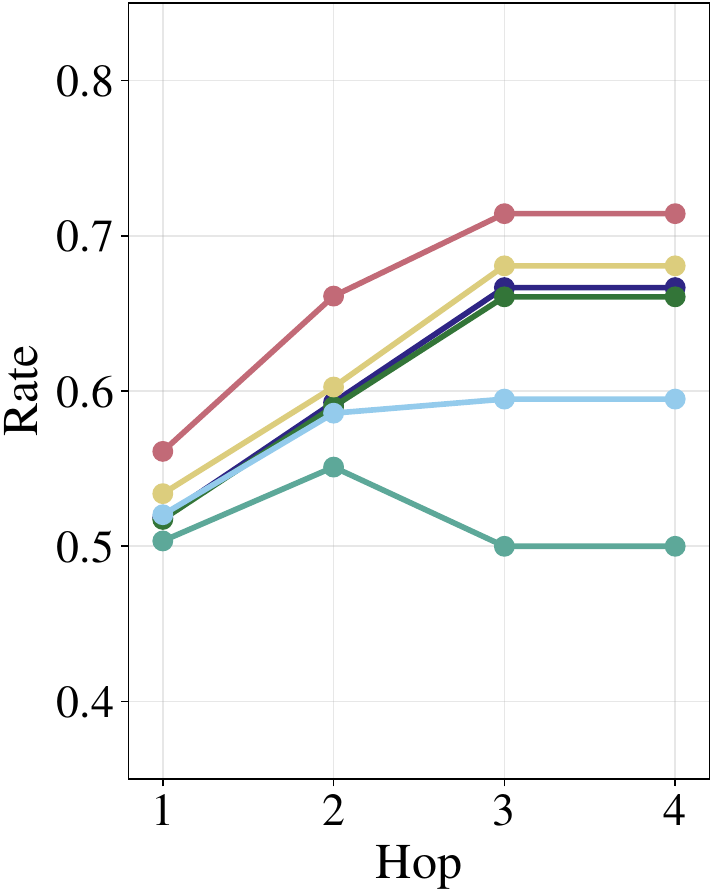}
    \caption{Retrieval Failure Rate}
  \end{subfigure}\hfill
  \caption{\textbf{Subquery drift effect on retrieval quality.} 2WIKI Dragin results (a) Subquery similarity to the original decreases across turns. (b) Retrieval failure rate (no gold document in top-$k$) across turns. 
}
\label{fig:subquery_trajectory_model}
\end{figure}

\paragraph{Computation robustness worsens on harder questions.}
We evaluate computational robustness using RE and CVR, which capture whether the model invokes the retriever and the LLM consistently across query variations.
Table~\ref{tab:robustness_efficiency} shows that retrieval-call counts are most sensitive on 2WIKI, with larger RE under most perturbations.
Spelling and grammatical errors produce the highest RE, indicating that surface-level noise most strongly changes the retrieval-call budget.
LLM-call patterns are similar.
On SQuAD, RE remains moderate and CVR stays high for Llama (up to 0.963 with the GB method), suggesting stable call counts under rewrites.
On 2WIKI, RE rises and CVR falls, with the largest degradation under spelling errors.

\paragraph{Computational robustness does not imply higher accuracy.}
Figure~\ref{fig:inaccuracy_four} shows that the generation-based method attains strong accuracy on harder datasets with larger models.
However, Table~\ref{tab:robustness_efficiency} shows the opposite pattern for computational robustness.
For 2WIKI with QwQ, computational robustness degrades substantially.
These results indicate that high accuracy does not necessarily imply robust computation.

\subsection{Retrieval Decision Robustness}
\paragraph{Query errors invoke over-confidence.}
We investigate the retrieval decision correctness using under- and over-confidence metrics.
Under-confident instances incur additional costs, while over-confidence harms accuracy by skipping necessary retrievals.
Figure~\ref{fig:under_over_2wiki} shows spelling-error appear on high over-confident cases, reaching up to 9.4\%, particulary in larger models.

\paragraph{Drifted subqueries miss gold documents.}
We study retrieval-decision robustness in multi-hop Adaptive RAG.
At hop $t$, the model generates a subquery $q_t$ and retrieves a top-$k$ document set $D_t$.
To examine how query variations perturb this process, we track (i) semantic drift in intermediate subqueries and (ii) hop-level retrieval failure rates.
We measure drift using mean subquery similarity (MSS), the average semantic similarity between each variant subquery and its corresponding original subquery across hops.
We measure retrieval failures using retrieval failure rate (RFR), defined as the proportion of hops where the top-$k$ retrieved documents contain no gold document, i.e., $\mathrm{recall@}k(D_t^{\mathrm{var}}, D_t^{\mathrm{Gold}})=0$.

As subqueries drift from the original trajectory (lower MSS), the top-$k$ retrieved sets increasingly miss gold evidence (higher RFR).
Figure~\ref{fig:subquery_trajectory_model} shows this pattern: subquery similarity decreases while RFR increases across hops, suggesting that subquery drift directly harms retrieval quality.

%% file: Text/7.Analysis.tex
\captionsetup[subfigure]{font=footnotesize}
\begin{figure}[t]
  \centering
  \includegraphics[width=0.36\textwidth]{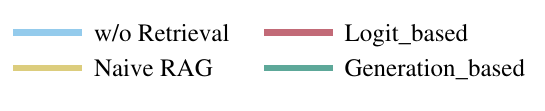}
  \vspace{-0.4em}
  \begin{subfigure}[t]{0.49\linewidth}
    \centering
    \includegraphics[width=\linewidth]{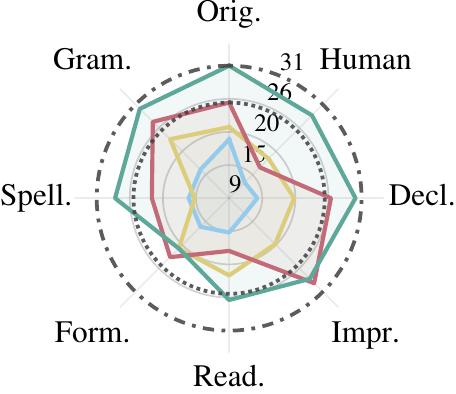}
    \caption{SQuAD}
  \end{subfigure}  \hfill
  \begin{subfigure}[t]{0.49\linewidth}
    \centering
    \includegraphics[width=\linewidth]{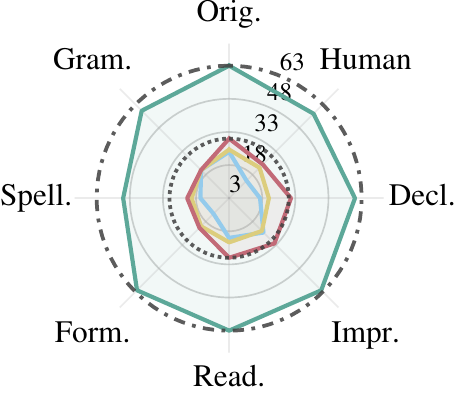}
    \caption{2WIKI}
  \end{subfigure}
  \caption{\textbf{InAccuracy results} with human rewrites.
  }
  \label{fig:inaccuracy_human_two}
\end{figure}

\paragraph{Human rewrites are semantically faithful but harder for the model.}
Figures~\ref{fig:query_analysis-2wikimultihopqa} (b) and (c) show that human rewrites are substantially shorter than other query variants, often compressing or omitting surface cues, while still preserving much of the original semantic intent.
However, Figure~\ref{fig:query_analysis-2wikimultihopqa} (a) shows that they induce substantially higher token-level loss, indicating that the model assigns much lower probability to these rewrites despite their semantic faithfulness.

\input{Tables/subquery_failure_example}
\input{Tables/Robustness_Acc_human}

In Figure~\ref{fig:inaccuracy_human_two}, human-authored rewrites lead to accuracy degradation compared to original query, in some cases comparable to that caused by spelling-errors. 
Table~\ref{tab:robustness_answer_human_2wiki} further shows reduced answer stability relative to the original queries. 
This degradation is similar to that observed under simple surface-level perturbations, indicating that robustness failures are not limited to synthetic noise.
To identify the source of errors, we trace a causal chain from properties of human rewrites to subquery drift, retrieval failures, and ultimately reduced end-to-end answer accuracy.

\label{sec:human_query_retrieval_decision}

\paragraph{Human query length and uncertainty are associated with subquery drift.}
Shorter and more uncertain human rewrites are reflected in subquery drift.
For Search-o1 on 2WIKI, query loss is negatively correlated with mean subquery similarity (MSS; Spearman’s $\rho$=-0.15), while query length is positively correlated with MSS ($\rho$=0.29). 
These results suggest that higher-loss, more compressed human phrasing destabilizes intermediate subqueries, leading to lower MSS.

\paragraph{Semantic drift causes retrieval trajectory collapse and evidence failures.}
Once intermediate subqueries drift, retrieval becomes unstable.
For Search-o1 on human rewrites, MSS is strongly negatively correlated with trajectory collapse ($\rho=-0.75$), showing that retrieved documents diverge from those of the original queries.
MSS is also negatively correlated with RFR ($\rho=-0.35$), indicating that gold documents are more often missing from the retrieved set.
These results suggest that human rewrites alter retrieval trajectories and increase failures to retrieve necessary evidence.

\begin{figure}[t!]
\centering
\includegraphics[width=0.8\linewidth]{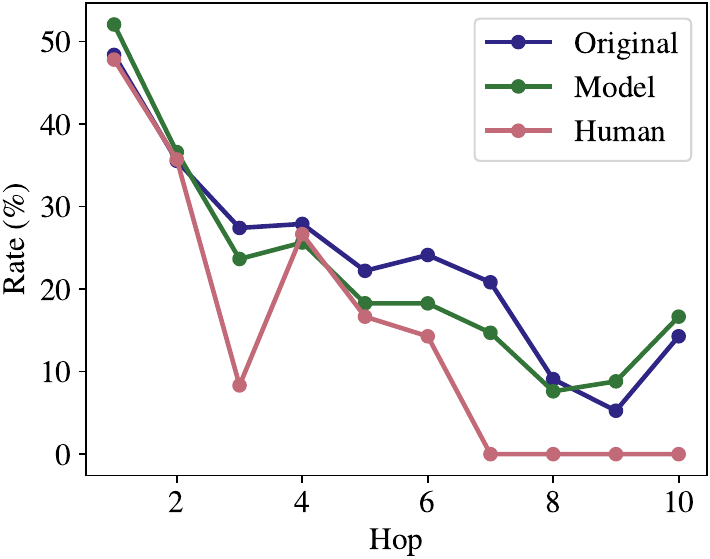}
\caption{
\textbf{Gold document inclusion rate across hops.} For each hop, we measure how often the top-$k$ retrieved results contain at least one gold document.}
\label{fig:human_query_drift}
\end{figure}

Figure~\ref{fig:human_query_drift} illustrates this pattern: the gold-document inclusion rate for human queries declines markedly for later hops relative to the original and model-generated queries.
Table~\ref{tab:subquery_failure} provides an example from QwQ outputs.
In the first few hops, generated subqueries remain highly similar to the original query (similarity $\approx 0.99$--$1.00$), but they begin to drift from hop 3 onward, leading to retrieval failures as reflected by the hit metric.

\paragraph{Retrieval errors propagate.}
Failures in intermediate retrieval steps translate into end-to-end accuracy degradation. 
Retrieval failure is negatively correlated with the performance gap $Y$ ($\rho=-0.20$ for Search-o1). 
When the system fails to retrieve gold documents, it lacks the necessary evidence, and downstream generation rarely recovers.

Overall, our findings indicate that human rewrites preserve semantic meaning yet increase the difficulty for Adaptive RAG systems.
Higher query loss and more compressed phrasing amplify subquery drift, which destabilizes the retrieval trajectory and increases retrieval failure, ultimately degrading final answer accuracy.

%% file: Tables/subquery_failure_example.tex
\begin{table*}[t]
\centering
\small
\setlength{\tabcolsep}{4pt}
\resizebox{\textwidth}{!}{%
\begin{tabular}{r p{6cm} p{6cm} c c c}
\toprule[1.3pt]
\textbf{} &
\textbf{Original Query} &
\textbf{Human Query} &
&
&
\\
\midrule
&
Which film has the director born earlier, The Adventures Of Priscilla, Queen Of The Desert or Harvest: 3,000 Years? &
Who was born first director of The Adventures Of Priscilla, Queen Of The Desert or Harvest: 3,000 Yearst &
& & \\
\midrule
\textbf{t} &
\textbf{Original Subqueries} &
\textbf{Human Subqueries} &
\textbf{Sim} &
\textbf{Hit(O)} &
\textbf{Hit(H)} \\
\midrule
1 &
director of the adventures of priscilla, queen of the desert &
director of the adventures of priscilla queen of the desert &
0.99 & \cmark & \cmark \\

2 &
director of harvest 3000 years &
director of harvest 3000 years &
1.00 & \cmark & \cmark \\

3 &
``harvest: 3,000 years'' director birth year &
harvest 3000 years movie director &
0.85 & \cmark & \xmark \\

4 &
giuliano carnimeo birth year &
harvest 3000 years korean movie director &
0.13 & \cmark & \xmark \\
\bottomrule[1.3pt]
\end{tabular}
}
\caption{
\textbf{Per-hop comparison of original and human subqueries.}
Sim is the semantic similarity between the two subqueries at hop $t$.
Hit(O) indicates whether the retrieved top-$k$ set contains any ground-truth supporting evidence for the original (O) and human (H) query.}
\label{tab:subquery_failure}
\end{table*}

%% file: Tables/Robustness_Acc_human.tex
\begin{table}[t]
\centering
\setlength{\tabcolsep}{3.0pt}
\renewcommand{\arraystretch}{1.08}

\begin{adjustbox}{width=\columnwidth,center}
\begin{tabular}{l ccccccc c}
\toprule[1.5pt]
\multirow{2}{*}{\textbf{Mtd.}} &
\multicolumn{7}{c}{\textbf{Similarity} $\uparrow$} &
\multirow{2}{*}{\textbf{Div.} $\downarrow$}\\
\cmidrule(lr){2-8}
& Human & Form. & Read. & Decl. & Impr. & Spell. & Gram. & \\
\midrule

\addlinespace[2pt]

LB & \Sim{0.518} & \Sim{0.528} & \Sim{0.570} & \Sim{0.573} & \Sim{0.612} & \Sim{0.522} & \Sim{0.572} & \Div{0.477} \\
GB & \Sim{0.529} & \Sim{0.542} & \Sim{0.534} & \Sim{0.568} & \Sim{0.598} & \Sim{0.520} & \Sim{0.558} & \Div{0.460} \\

\bottomrule[1.5pt]
\end{tabular}
\end{adjustbox}

\caption{
\textbf{Answer Robustness with human queries.} \simhl{Similarity} to the original query and \divhl{diversity} on 2WIKI (QwQ-32B) including human rewrites.
}
\label{tab:robustness_answer_human_2wiki}
\end{table}

%% file: Text/8.Conclusion.tex
In this work, we introduced a large-scale benchmark of query variations to evaluate the robustness of Adaptive RAG systems.
We studied three dimensions of robustness: answer robustness, computation robustness, and retrieval decision robustness, covering both final outputs and intermediate reasoning steps.
We showed that both logit-based and generation-based adaptive methods are sensitive to query perturbations, which degrade accuracy, efficiency, and stability.
Human rewrites were especially harmful, inducing drift in intermediate subqueries and retrieval-trajectory collapse, leading to substantial end-to-end performance drops.
Finally, we found that human rewrites differ systematically from model-generated variations, highlighting a realism gap in current robustness evaluations.

%% file: Text/9.Limitation.tex
Although our study offers a systematic view of Adaptive RAG behavior under realistic query variation, several limitations remain. First, our human rewrites are constructed as meaning-preserving paraphrases of existing questions.
While this setting captures a common and controlled form of user rephrasing, it does not cover broader interactive behaviors such as multi-turn follow-up questions, conversational reformulations, or deliberately adversarial inputs. 
Moreover, these rewrites may differ from naturally occurring first-shot user queries in real-world settings.
Second, our experiments focus on a specific set of retrievers, LLMs, and Adaptive RAG controllers.
Despite the scale of our analysis, other retrieval architectures or routing mechanisms may exhibit different robustness–efficiency trade-offs, making broader system coverage an important direction for future work. 
Finally, although the human rewrites were carefully collected and anonymized, they capture only a limited subset of the diversity found in real-world query phrasing. 
Extending human-written evaluation across languages, domains, and user populations would further strengthen the robustness of this benchmark.

%% file: Text/11.Appendix.tex
\section{Implementation Details}
\label{app:implementation_details}

We conduct our experiments with two LLMs: Llama~3.1-8B-Instruct and QwQ-32B. We select these models to cover both a representative instruction-tuned baseline and a complementary reasoning-oriented model. Prior work commonly evaluates adaptive RAG with mid-sized backbones (approximately 3B--13B)~\cite{flare, dragin, adaptive_rag}, often using Llama~3 8B variants~\cite{knowing_youdontknow,adarague, marina-etal-2025-llm}; thus, we include Llama~3.1-8B-Instruct. 
We further include QwQ-32B, a reasoning-focused model from the Qwen family that has also been adopted in agentic search settings~\cite{searcho1}. This combination allows us to assess query-variation robustness across different model families, scales, and capability profiles.

We use two retrievers: BM25~\cite{bm25} as the sparse retriever and Contriever~\cite{contriever} as the dense retriever. Our primary results are based on BM25, since it serves as the default retrieval component in many prior adaptive RAG systems and benchmarks~\cite{dragin,flare,marina-etal-2025-llm}.

We generate responses using nucleus sampling with temperature $\mathcal{T}=0.7$ and top-$p=0.95$, and report all main results from three multiple runs. To quantify variance, we additionally report results in Appendix~\ref{appendix:multi_run} showing the standard deviation of the accuracy. All experiments are run locally on 2$\times$ NVIDIA A100 GPUs with an Intel(R) Xeon(R) Silver 4210R CPU @ 2.40GHz.

For the DRAGIN method, following prior work~\cite{dragin}, we tune the retrieval threshold by evaluating several high-performing candidate values on the Natural Questions (NQ) dataset and select 0.6.
We then fix this setting and apply the same configuration to all remaining datasets to ensure a consistent evaluation of computational cost across benchmarks.
For the Search-o1 method, maximum search limit hyperparameter is set 15 for the multi-hop QA and 5 for the single-hop QA. 
Rest of the hyperparamer follows the ~\cite{searcho1} setting. 
All of the embedding similarity is calculated by the \texttt{sentence-transformers/all-mpnet-base-v2}, which is suitable for clustering and semantic search task.

\section{Prompts}
\label{app:prompts}
\input{Tables/prompt_formality}

\input{Tables/prompt_readability}
\input{Tables/prompt_type}

\input{Tables/prompt_grammar}

\textbf{Readability Variation.} We utilize the prompt from ~\citet{out_of_style} shown in Table~\ref{tab:readability_prompt} to reduce the readability of the original query.

\section{Benchmark Construction}
\subsection{Model-generated Queries}
\label{app:query details}
Style and sentence-type variations are generated using GPT-5-mini\footnote{gpt-5-mini-2025-08-07} with the prompts provided in Appendix~\ref{app:prompts}.
For the error categories, we use external tools to inject or verify errors. Spelling errors are injected using the \texttt{SpellingAug} function in \texttt{nlpaug}.
For the grammar error category, we use LanguageTool\footnote{\url{https://languagetool.org}}
 to verify that each query contains at least one grammatical error. If a query is classified as having no grammar errors, we regenerate it using GPT-5-mini to inject a grammatical error.

For all query variations, the co-authors manually inspected sampled queries. 
The annotation process was conducted by two co-authors, both fluent in English. 
Annotators were instructed to focus on the quality of the rewritten queries, whether the intended perturbation was correctly applied while preserving the original meaning.
We randomly sampled 100 examples across query variations, and the two annotators marked 96\% and 98\% as passing, indicating that the rewrites are largely meaning-preserving and adhere well to the intended query categories.

\subsection{Human Annotation Details}
\label{app:human_annotation_details}
Annotation guidelines are provided in Table~\ref{tab:human_instruction}. 
Annotators were shown the original query and its corresponding answer and were instructed to generate alternative formulations as if they were posing the question themselves.
The answer was provided as minimal grounding to reduce semantic drift during rewriting and to preserve the original intent of the query. 
Accordingly, the resulting human rewrites should be interpreted as controlled, semantically equivalent, human-like reformulations rather than fully naive first-shot user queries.

We recruited 12 volunteer annotators to produce human query rewrites. 
All volunteers were fluent in English and between 20 and 40 years old; with 5 female and 7 male participants. 
All annotators were graduate students or postdoctoral researchers with research experience in NLP or closely related fields. 
They reported high familiarity with AI tools (e.g., LLM-based assistants) and used them frequently in everyday tasks.
Participation was unpaid and voluntary. 
Prior to the annotation process, all annotators provided informed consent for their rewrites to be collected and used for research purposes.

\section{Dense Retrieval Results}
We compare sparse retrieval with the dense retriever Contriever (ref) on the SQuAD and 2WikiMultiHopQA datasets. Although sparse retrievers are known to be sensitive to surface-form variations and dense retrievers are generally more robust, performance under spelling errors and human variations remains low, indicating that the issue persists even with dense retrieval.
\input{Tables/Dense_retriever}

\section{Flip Rates Across Temperatures}
\input{Tables/fliprate_diff_temp}
We extend the preliminary decoding-sensitivity diagnostic by varying the sampling temperature to assess whether flip rates in the retrieval decision are driven by a particular sampling regime. Specifically, we evaluate temperatures $\{0.0, 0.3, 0.7, 1.0\}$ on the SQuAD dataset which showed high flip rates in our initial analysis.
As shown in Table~\ref{tab:flip_rates_temperature}, flip rates remain consistently high across all temperatures, including near-deterministic decoding at $T=0.0$. This indicates that the observed robustness gap is not primarily attributable to stochastic sampling, but instead reflects instability in the retrieval decision itself under small decoding variations.

\section{Closed-source LLMs}
\input{Tables/Closed_llm}
To further assess generalizability beyond open-source LLMs, we additionally evaluate a closed-source model, GPT-4o mini, under the same protocol on 2WIKI and SQuAD. we note that we test the model under generation-based method, as the logit-based method is not applicable to closed-source LLMs. As shown in Table~\ref{tab:closed_model_results}, GPT-4o mini also exhibits a noticeable accuracy drop under human rewrites compared to the original queries (while retrieval counts remain comparable), aligning with our main observation that semantically equivalent rewrites can degrade RAG performance. This suggests that the robustness issue is not confined to a specific open model family, but can also persist in closed-source deployments.

\section{Additional Accuracy Metrics}
\input{Tables/f1}
We validate our use of the InAccuracy metric by comparing it against token-level F1. While InAccuracy evaluates whether the final model output contains the gold answer, token-level F1 measures token overlap between the predicted answer and the reference, capturing partial correctness even when the match is not exact. Table~\ref{tab:inacc_vs_f1} reports both metrics for Llama-3.1-8B outputs on SQuAD and 2WikiMultiHopQA under the DRAGIN setting. Overall, token-level F1 exhibits the same degradation patterns as InAccuracy from original queries to human rewrites, providing complementary evidence that the observed robustness gaps are not an artifact of a particular evaluation metric.

\section{Computational Cost}
\input{Tables/cost_additional}
In the main paper, we discuss computational cost using LLM calls and retriever calls, which are the primary overhead signals in adaptive RAG systems. Here, we additionally report token usage as a complementary cost-related measure.

We define token usage as the total number of tokens in the model's full generated output. As shown in Table~\ref{tab:token_usage}, generation based method(Search-o1) consistently generates much longer outputs than logit-based method(DRAGIN), while no clear systematic trend appears across query variation types. This additional analysis is consistent with the call-based findings reported in the main paper.

\section{Helpfulness and Harmfulness of Retreivals}
\begin{figure*}[t!]
\centering
\includegraphics[width=0.6\linewidth]{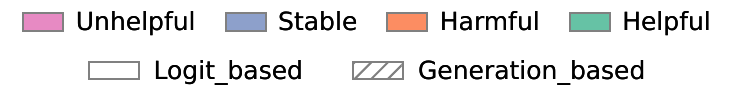}\par\vspace{-2pt}
\includegraphics[width=\linewidth]{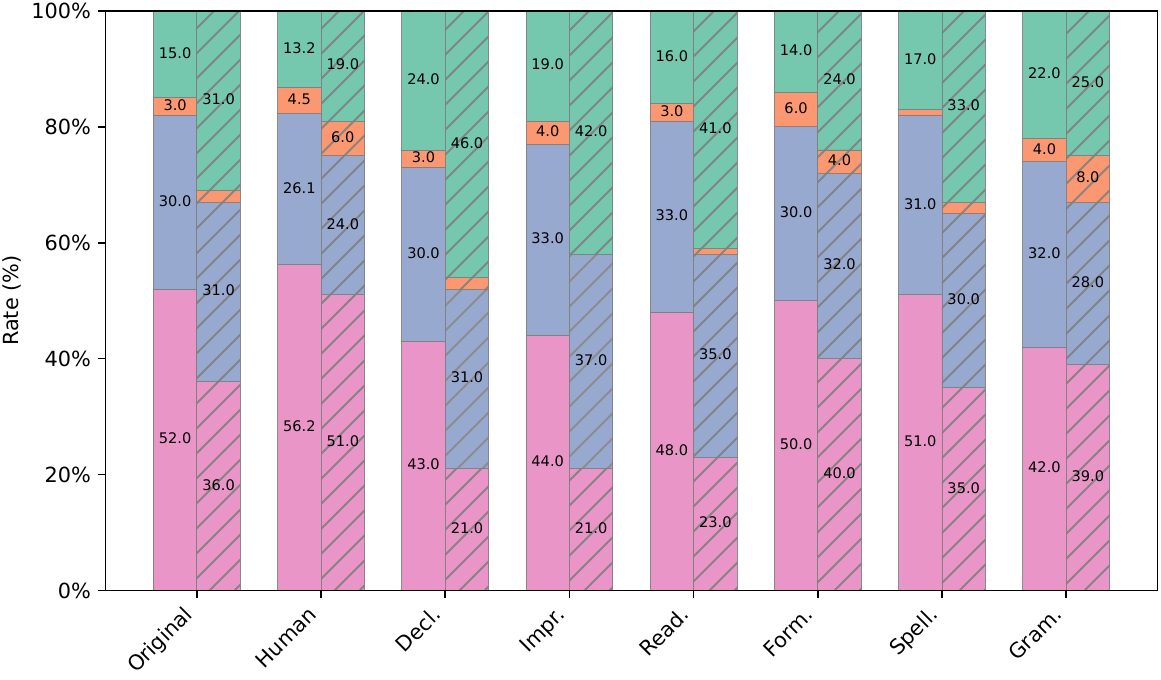}
\caption{Helpfulness and harmfulness distribution on NQ using the QwQ-32B model. }
\label{fig:helpfulness_nq}
\end{figure*}

We report the proportions of unhelpful, stable, harmful, and helpful outcomes. We compare accuracy across three settings: no retrieval, logit-based retrieval, and generation-based retrieval. A case is considered helpful when the method changes an originally incorrect answer into a correct one. Conversely, it is considered harmful when the model initially answers correctly but becomes incorrect after applying the method. Stable refers to cases where the model remains correct throughout, while unhelpful refers to cases where it remains incorrect throughout.
In Figure~\ref{fig:helpfulness_nq}, we observe that human-written queries show somewhat higher harmful rates, particularly when they contain grammatical errors. In contrast, declarative and imperative queries exhibit relatively high helpful rates.

\section{Multi-Run Stability Analysis}
\label{appendix:multi_run}
\input{Tables/multi_run_llama}
\input{Tables/multi_run_qwq}

To assess the stability of our findings, we repeat the full evaluation protocol over three independent runs. All runs use the same decoding configuration (temperature $\mathcal{T}=0.7$, top-$p=0.95$) and identical experimental settings, with only the sampling randomness varied across runs.

Tables~\ref{tab:multi_run_llama_variation_cols} and~\ref{tab:multi_run_qwq_variation_cols} report the mean and standard deviation of accuracy over three runs for Llama-3.1-8B and QwQ-32B, respectively. Across settings, the run-to-run variance remains small relative to the performance gap between original queries and their human rewrites. Notably, the accuracy drop under semantically equivalent rewrites is consistent across runs, and spelling-error and human-rewrite conditions consistently exhibit substantial degradation. These results suggest that our findings are robust and not attributable to a particular random seed.

\section{Robustness Results}
\label{app:robustness_results}
In this section, we report full results for all six datasets and model variants in our benchmark.

\subsection{Answer Robustness}
\label{app:answer_robustness}
\begin{figure*}[t!] 
\centering
\includegraphics[width=\linewidth]{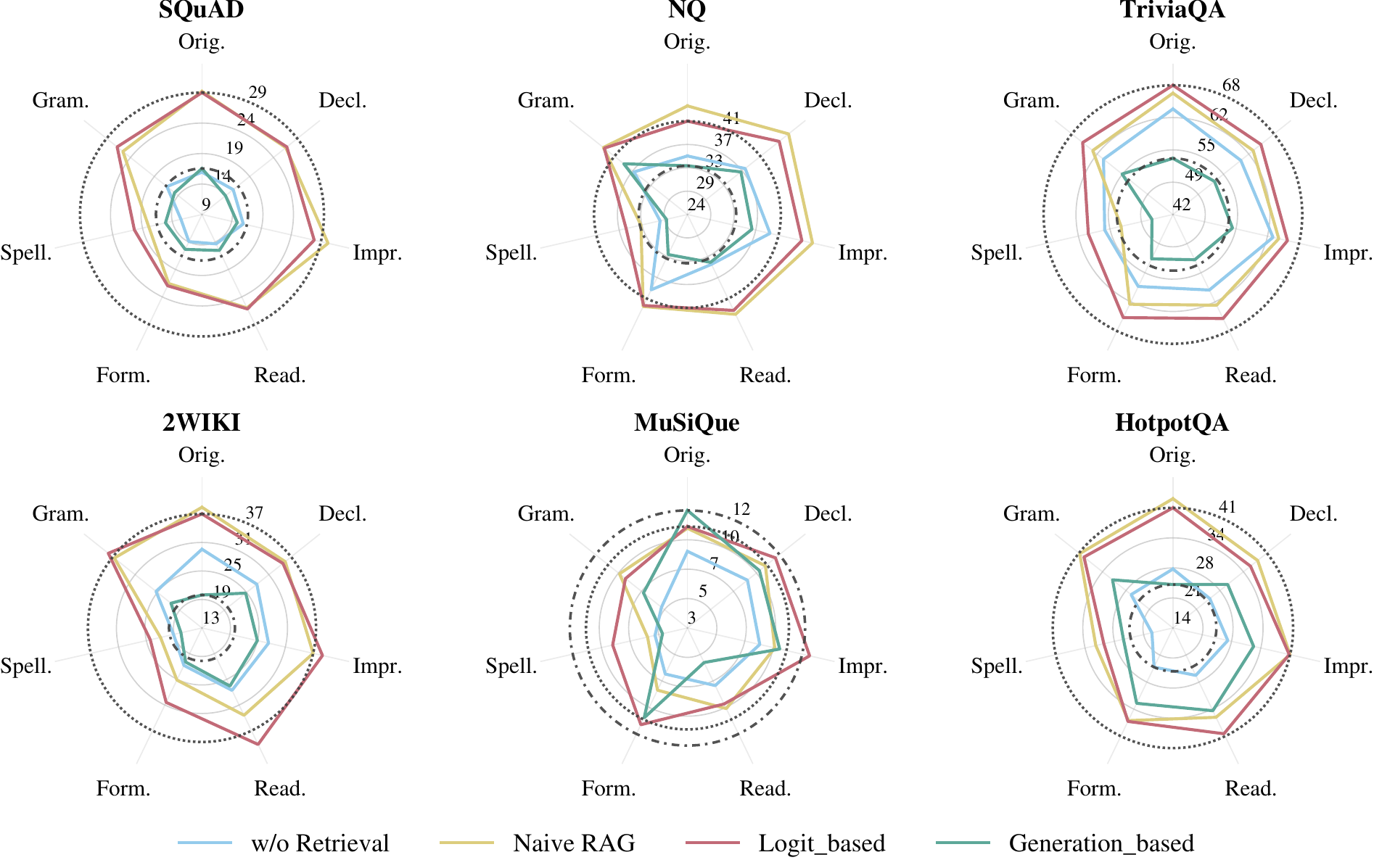}
\caption{\textbf{InAccuracy (Llama-3.1-8B).}
Performance across datasets for four Adaptive RAG methods. Each panel reports InAccuracy on seven query variations: the original query and six model-generated perturbations.}
\label{fig:radar_full_llama31_8b_nohuman}
\end{figure*}

\begin{figure*}[t!]
\centering
\includegraphics[width=\linewidth]{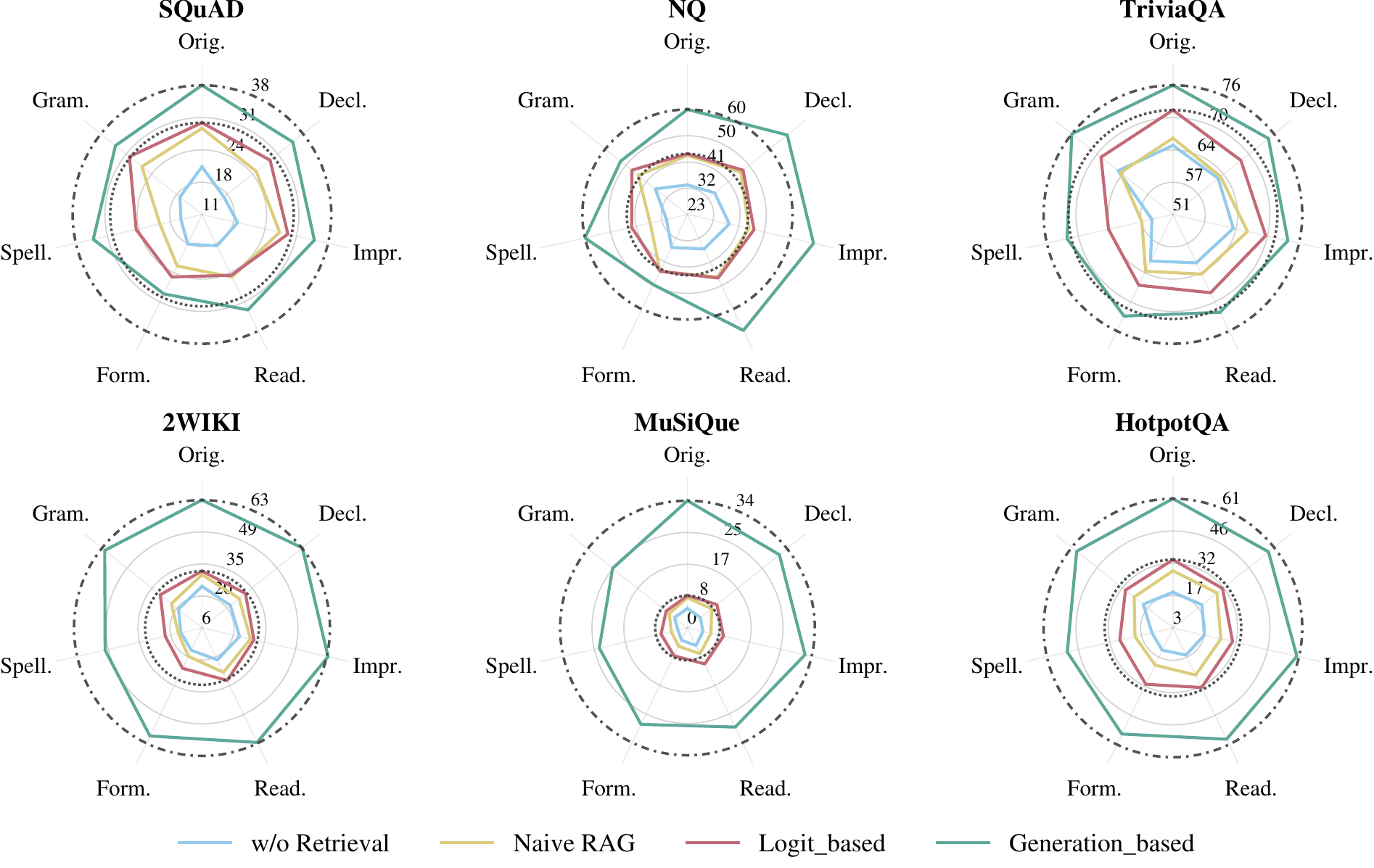}
\caption{\textbf{InAccuracy (QwQ-32B).}
Performance across datasets for four Adaptive RAG methods. Each panel reports InAccuracy on seven query variations: the original query and six model-generated perturbations).}
\label{fig:radar_full_qwq32b_nohuman}
\end{figure*}

\begin{figure*}[t!]
\centering
\includegraphics[width=\linewidth]{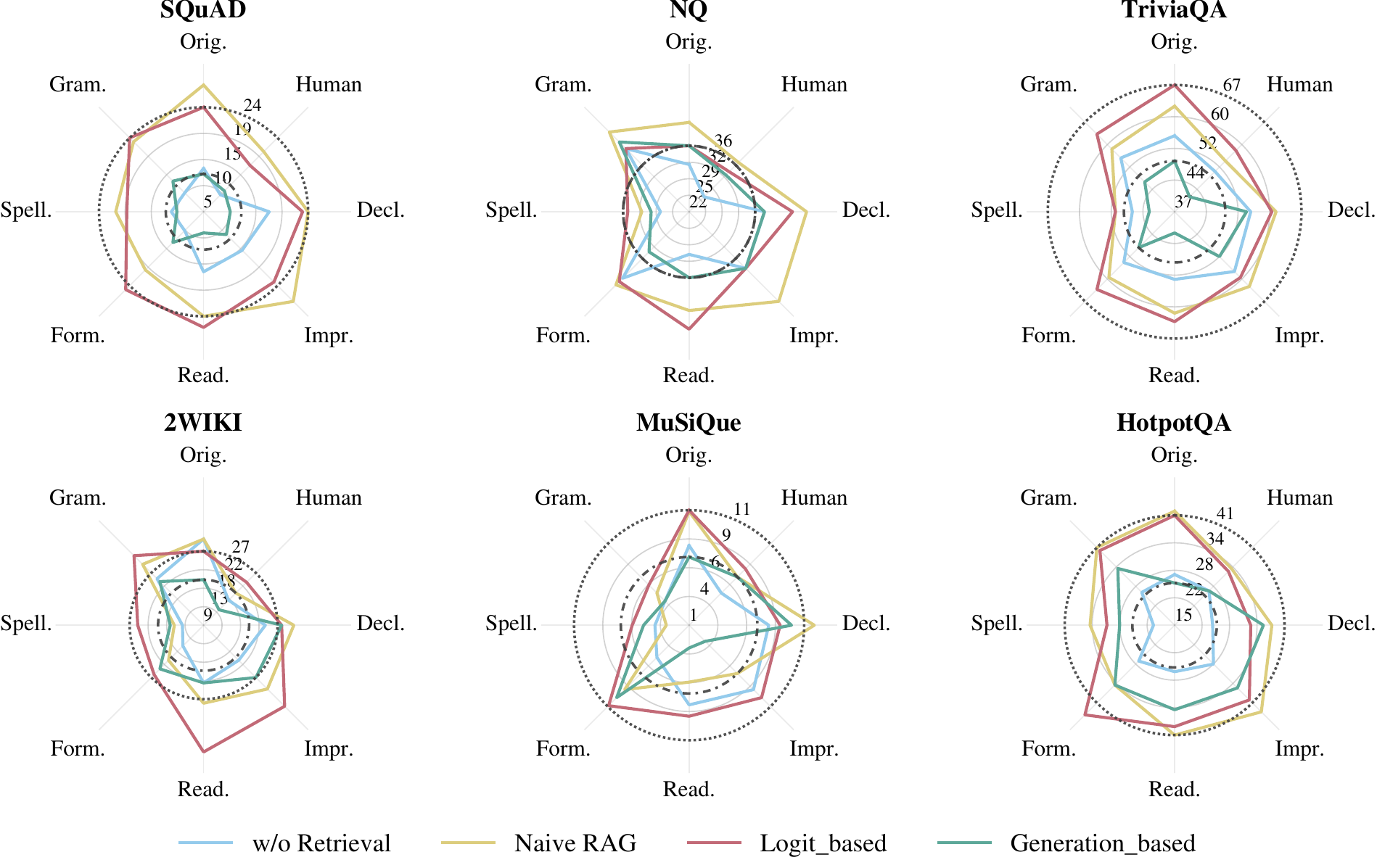}
\caption{\textbf{ InAccuracy (Llama-3.1-8B) including human rewrites.}
Performance across datasets for four Adaptive RAG methods, evaluated on a matched subset of instances that have human rewrites.
For each panel, InAccuracy is reported for eight query variations (original, human rewrite, and six model-generated perturbations), all computed on the same human-rewrite subset.}
\label{fig:radar_full_llama31_8b_withhuman}
\end{figure*}

\begin{figure*}[t!]
\centering
\includegraphics[width=\linewidth]{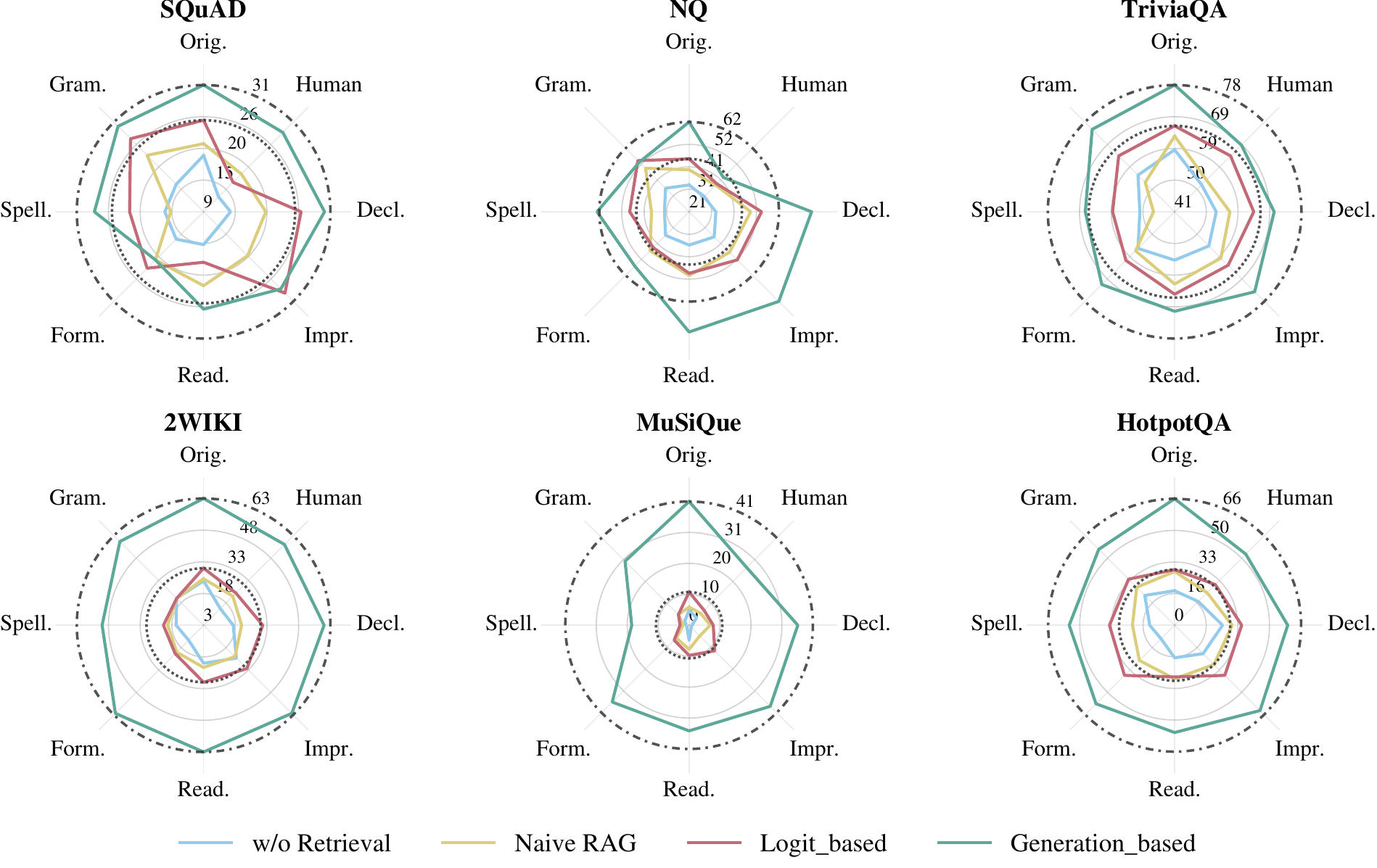}
\caption{\textbf{InAccuracy (QwQ-32B) including human rewrites.}
Performance across datasets for four Adaptive RAG methods, evaluated on a matched subset of instances that have human rewrites. For each panel, InAccuracy is reported for eight query variations (original, human rewrite, and six model-generated perturbations), all computed on the same human-rewrite subset.}
\label{fig:radar_full_qwq32b_withhuman}
\end{figure*}

\paragraph{InAccuracy.}
Figure~\ref{fig:radar_full_llama31_8b_nohuman} and~\ref{fig:radar_full_qwq32b_nohuman} show consistent degradation under spelling errors. In contrast, sentence-type variations (imperative and declarative) achieve performance comparable to the original, suggesting that these rewrites have little effect on accuracy.
For human rewrites, Figures~\ref{fig:radar_full_llama31_8b_withhuman} and~\ref{fig:radar_full_qwq32b_withhuman} show consistently lower performance across most datasets.

\paragraph{Answer Consistency.}
Tables~\ref{tab:robustness_simdiv_nohuman_llama} and~\ref{tab:robustness_simdiv_nohuman_qwq} report model-generated answer consistency across all six datasets. Full results including human query variations are shown in Tables~\ref{tab:robustness_simdiv_withhuman_llama} and~\ref{tab:robustness_simdiv_withhuman_qwq}.
Overall, human rewrites tend to yield answers that are less similar to those produced for the original queries.

\subsection{Computation Robustness}
\begin{figure*}[t!]
\centering
\includegraphics[width=\linewidth]{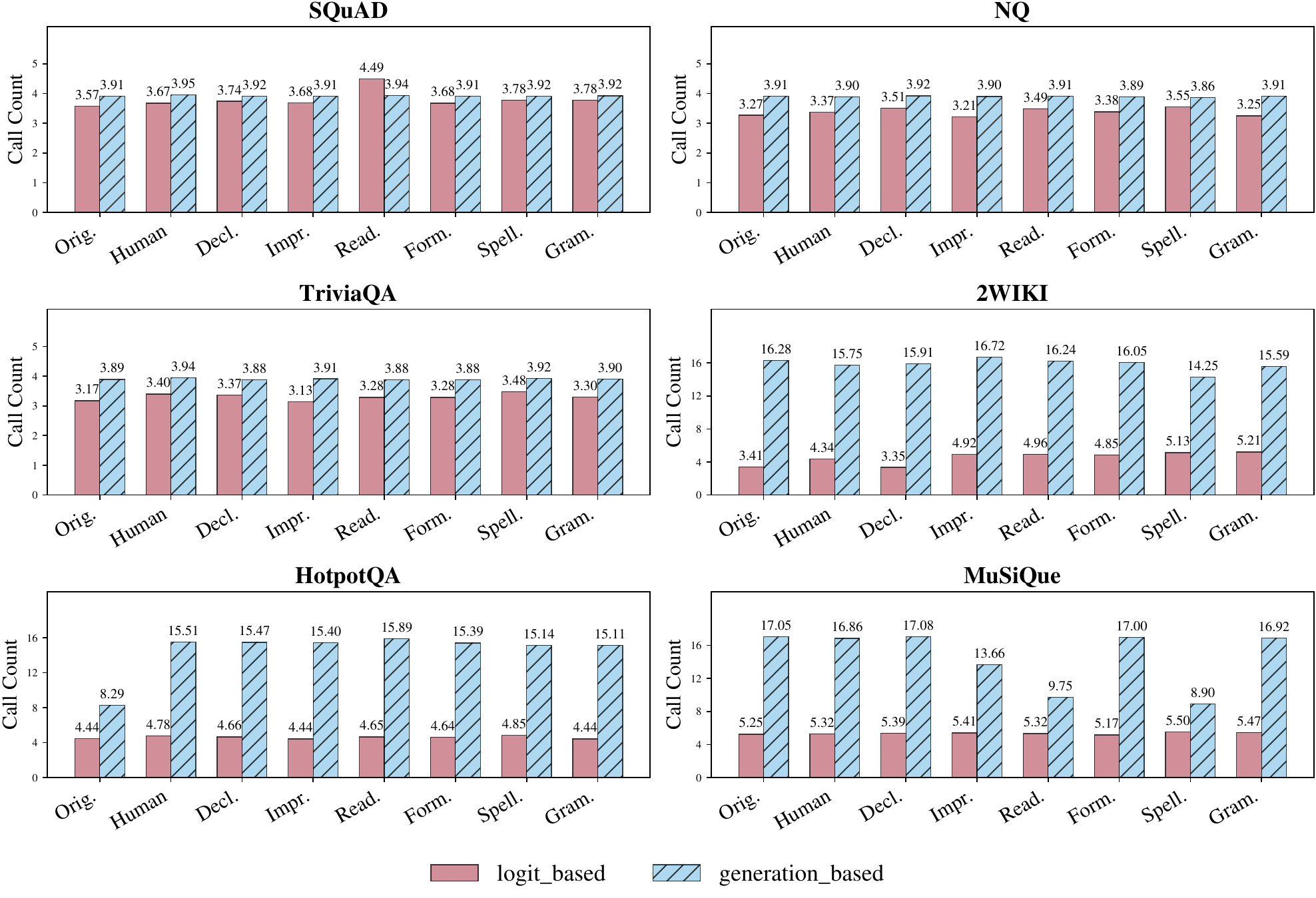}
\caption{\textbf{LLM Call} on Llama-3.1-8B on both model-generated and human queries.} 
\label{fig:llmcall_llama_human}
\end{figure*}

\begin{figure*}[t!]
\centering
\includegraphics[width=\linewidth]{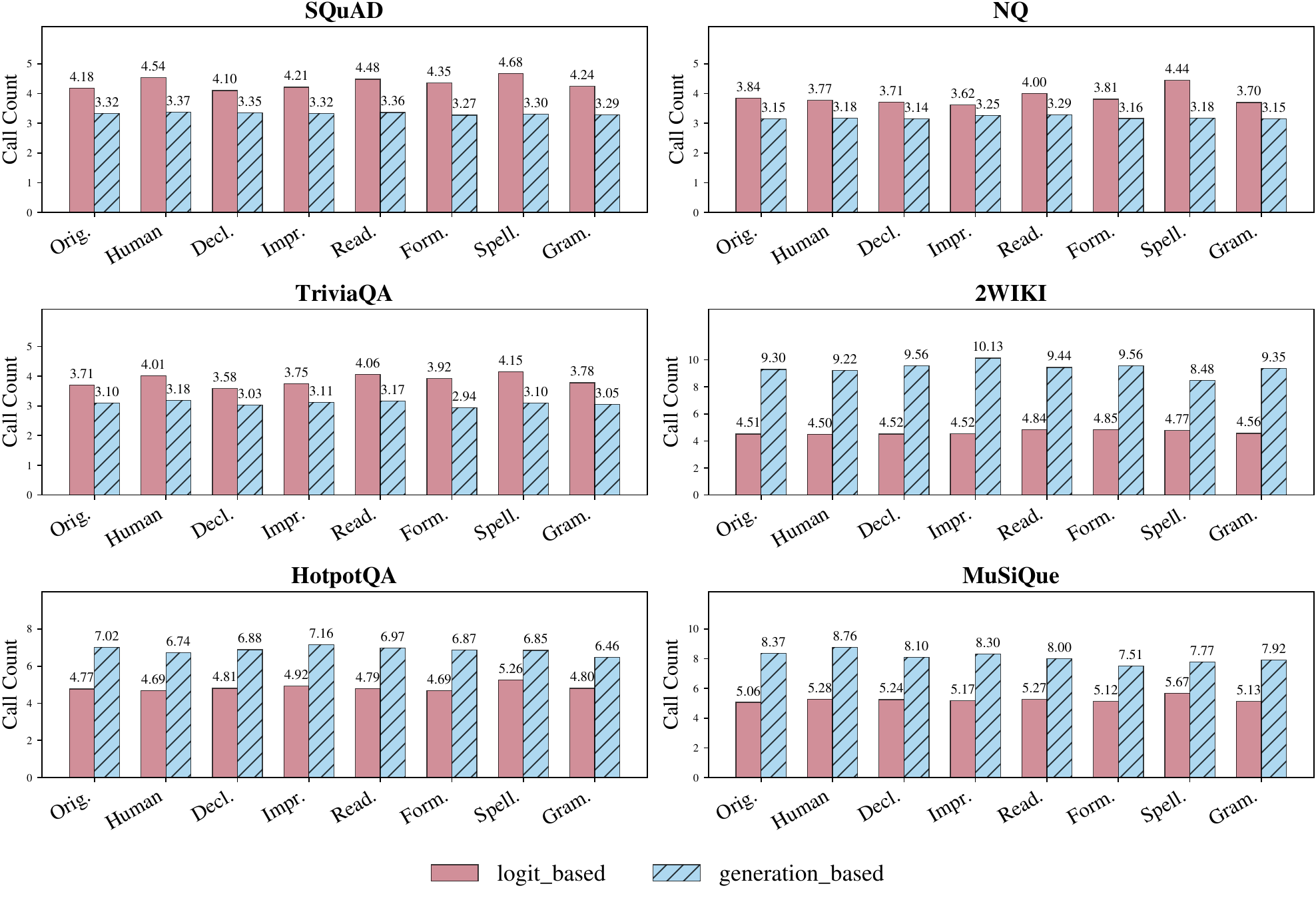}
\caption{\textbf{LLM Call} on QwQ-32B on both model-generated and human queries.}
\label{fig:llmcall_qwq_human}
\end{figure*}

\begin{figure*}[t!]
\centering
\includegraphics[width=\linewidth]{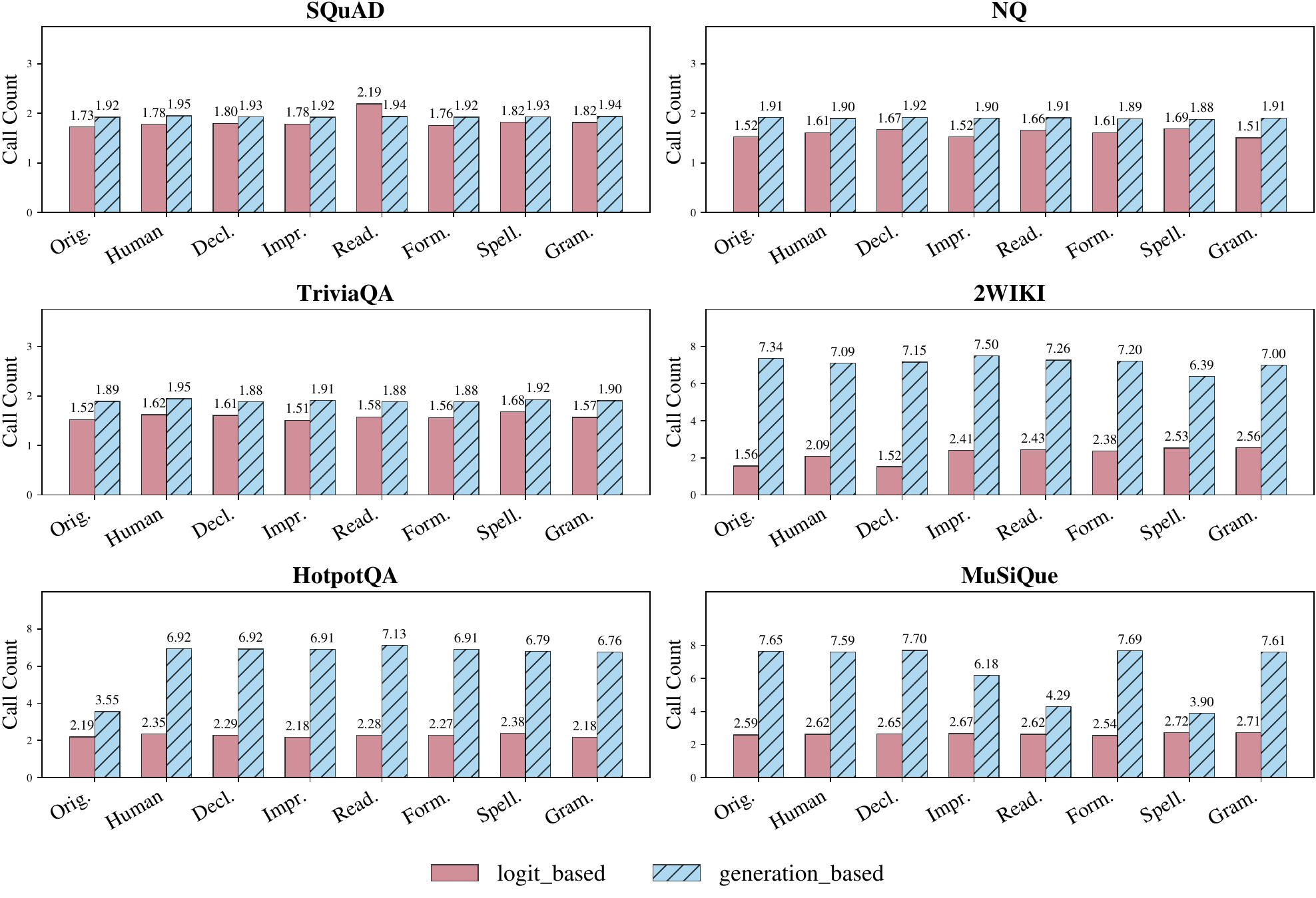}
\caption{\textbf{Retriever Call} on Llama-3.1-8B on both model-generated and human queries.} 
\label{fig:retrievercall_llama_human}
\end{figure*}

\begin{figure*}[t!]
\centering
\includegraphics[width=\linewidth]{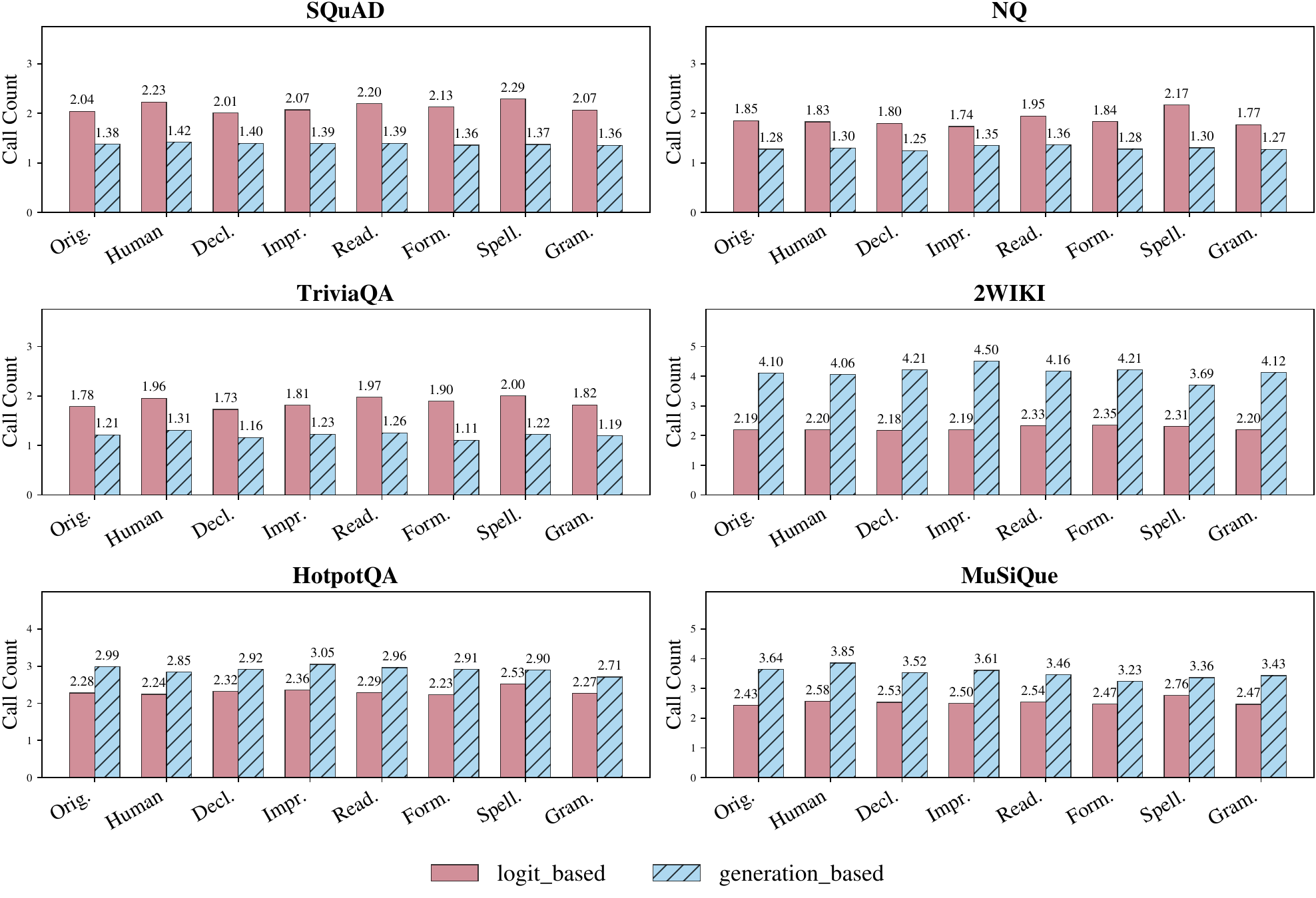}
\caption{\textbf{Retriever Call} on QwQ-32B on both model-generated and human queries.}
\label{fig:retrievercall_qwq_human}
\end{figure*}

We measure computation robustness in terms of the computational cost incurred by the LLM and the retriever.
We report the actual number of calls for each method--dataset pair.
Figures~\ref{fig:llmcall_llama_human} and~\ref{fig:llmcall_qwq_human} visualize the number of LLM calls across query variations.
Figures~\ref{fig:retrievercall_llama_human} and~\ref{fig:retrievercall_qwq_human} report the corresponding retriever calls for each model.
As dataset difficulty increases, the generation\_based method adaptively increases both LLM and retriever calls.

Robustness across model-generated query variations is reported in Tables~\ref{tab:robustness_efficiency_llama_nohuman} and~\ref{tab:robustness_efficiency_qwq_nohuman}.
Results including human queries are shown in Tables~\ref{tab:robustness_efficiency_llama} and~\ref{tab:robustness_efficiency_qwq}.
Overall, increased query complexity and higher call counts are associated with lower computational robustness.

\section{Self-Knowledge Results}
\label{app:self_knowledge_results}
\begin{figure*}[t!]
\centering
\includegraphics[width=\linewidth]{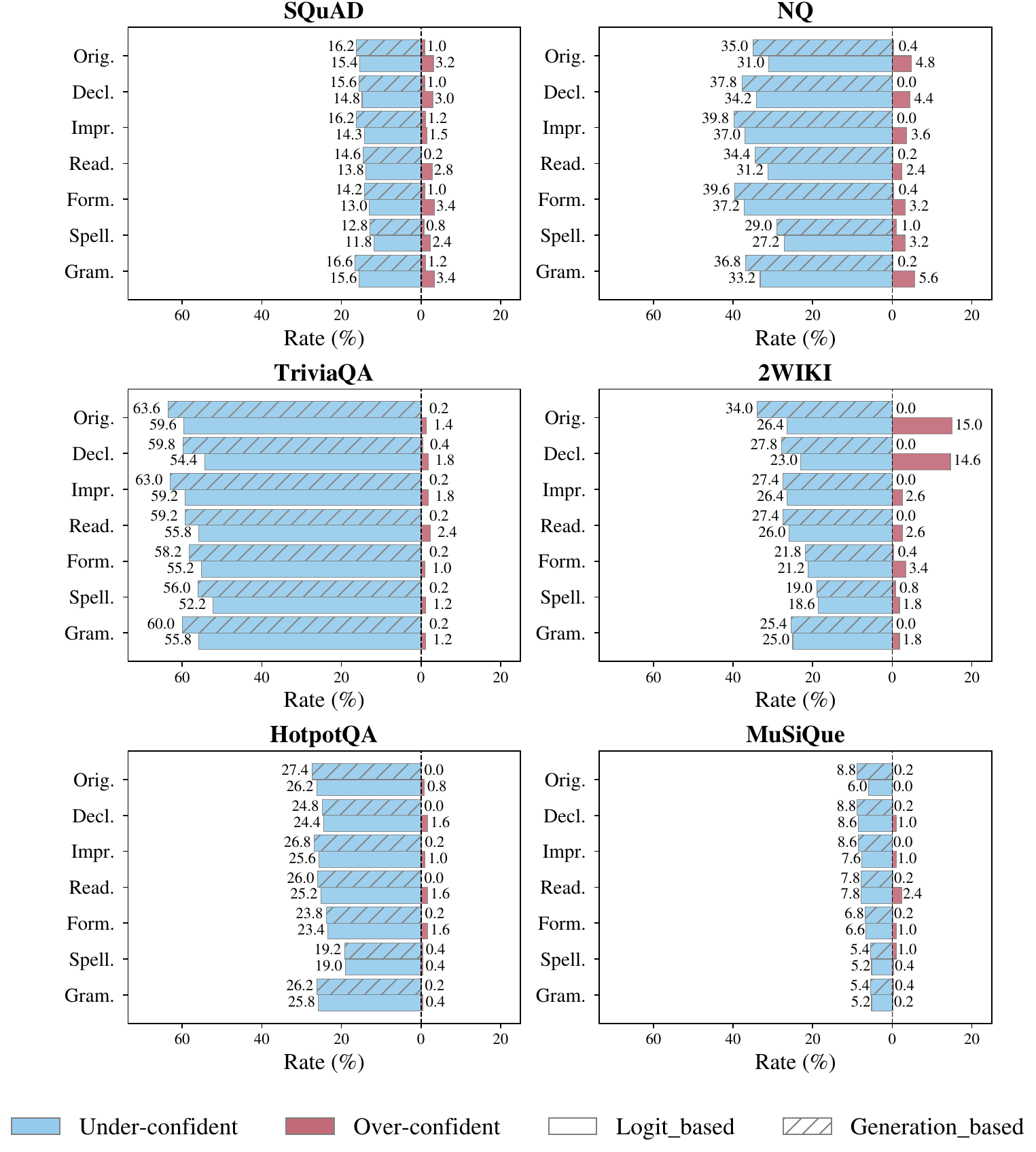}
\caption{\textbf{Under- and Over-confidence results}on Llama-3.1-8B on model-generated queries.} 
\label{fig:under_over_full_llama_nohuman}
\end{figure*}

\begin{figure*}[t!]
\centering
\includegraphics[width=\linewidth]{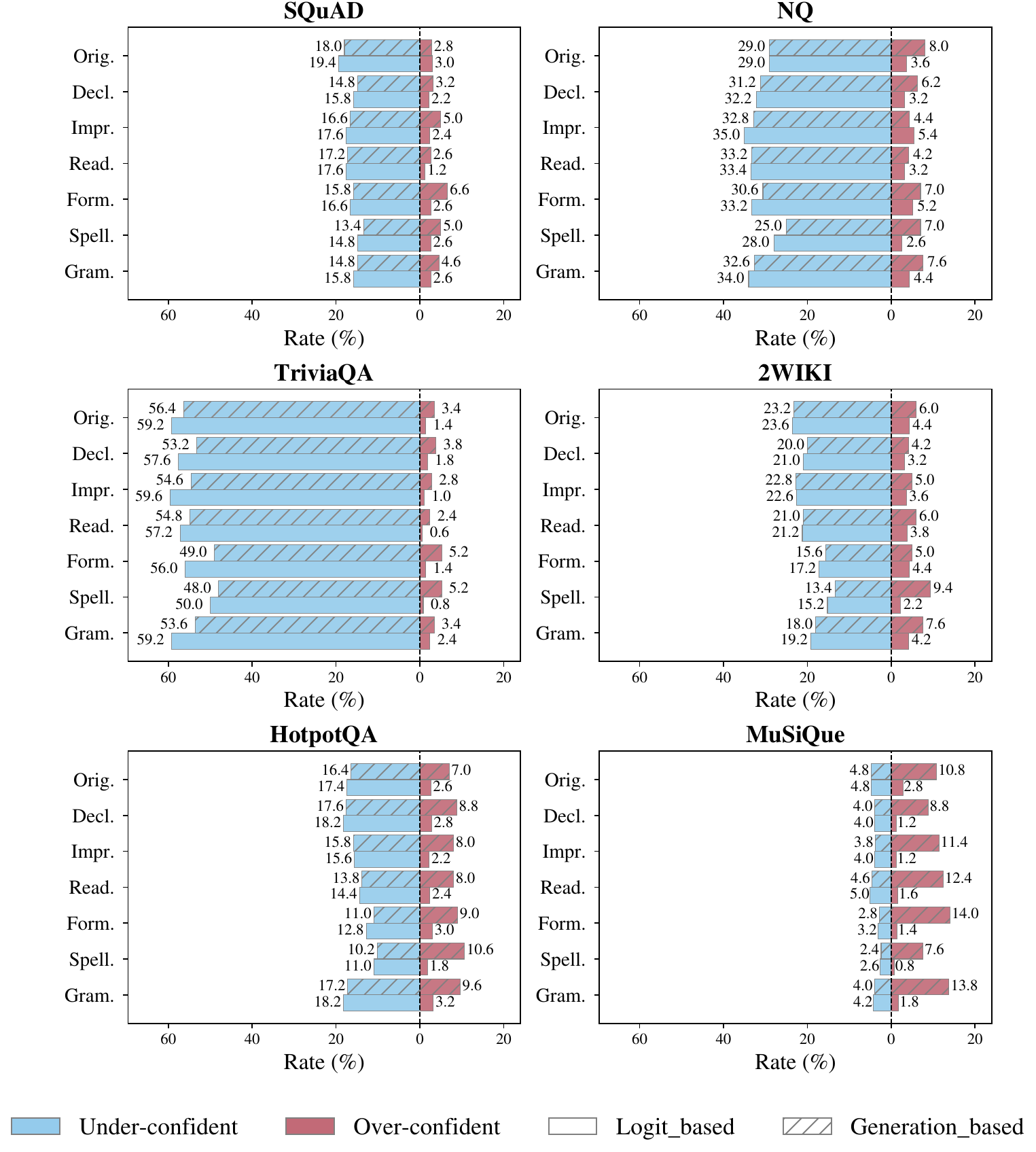}
\caption{\textbf{Under- and Over-confidence results}on QwQ-32B on model-generated queries.} 
\label{fig:under_over_full_qwq_nohuman}
\end{figure*}

\begin{figure*}[t!]
\centering
\includegraphics[width=\linewidth]{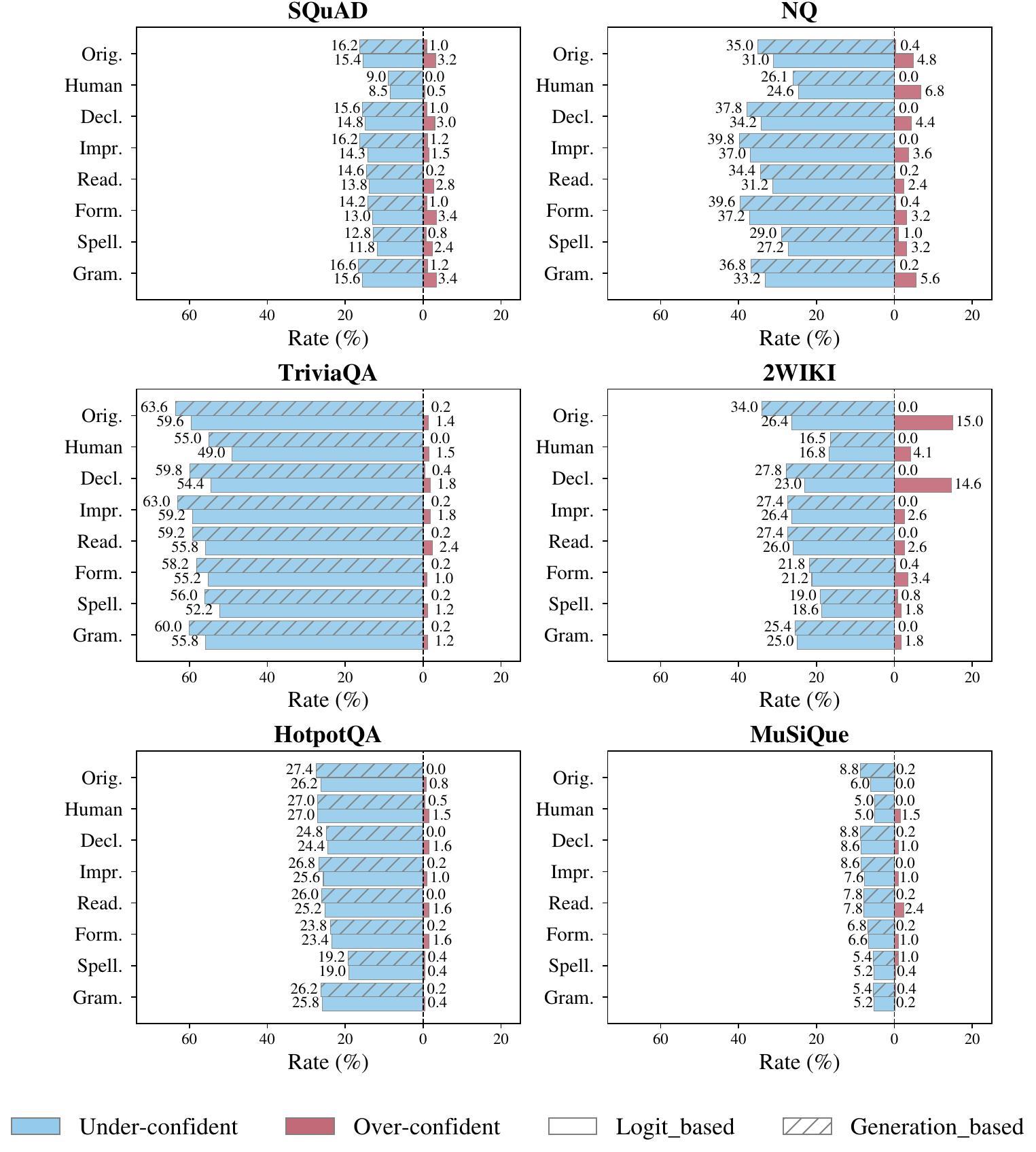}
\caption{\textbf{Under- and Over-confidence results}on Llama-3.1-8B on including human-written queries.} 
\label{fig:under_over_full_llama_human}
\end{figure*}

\begin{figure*}[t!]
\centering
\includegraphics[width=\linewidth]{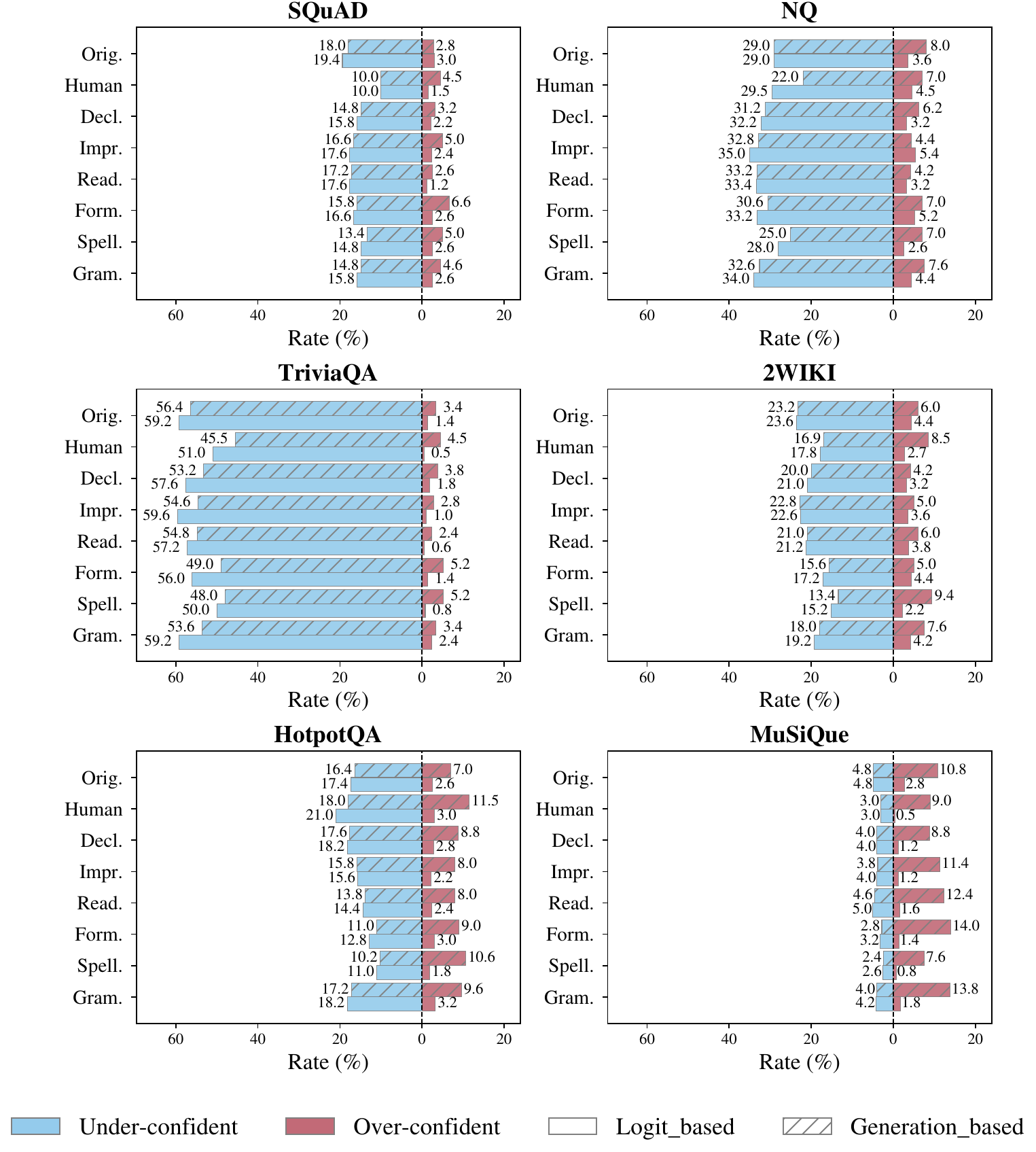}
\caption{\textbf{Under- and Over-confidence results} on Llama-3.1-8B on including human-written queries.} 
\label{fig:under_over_full_qwq_human}
\end{figure*}

We visualize the under- and over-confidence results for model-generated queries in Figures~\ref{fig:under_over_full_llama_nohuman} and~\ref{fig:under_over_full_qwq_nohuman}.
As dataset difficulty increases, the system exhibits more overconfident cases.
Figures~\ref{fig:under_over_full_llama_human} and~\ref{fig:under_over_full_qwq_human} report the results when including human rewrites as well.

\section{Human Query Analysis}
\label{app:human_query_analysis}
\begin{figure*}[t!]
\centering
\includegraphics[width=\linewidth]{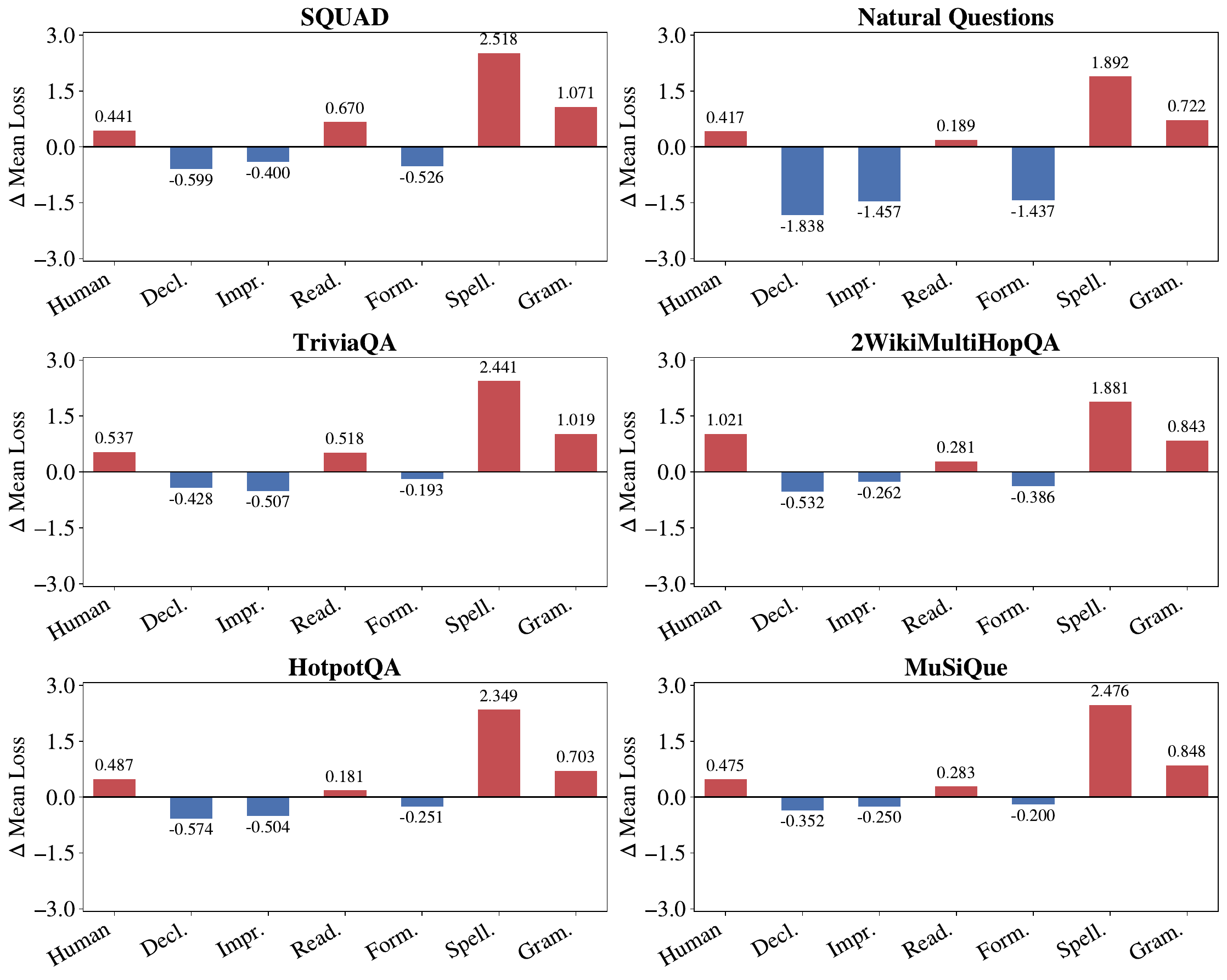}
\caption{Loss score across all datasets}
\label{fig:query_ppl_all}
\end{figure*}

\begin{figure*}[t!]
\centering
\includegraphics[width=\linewidth]{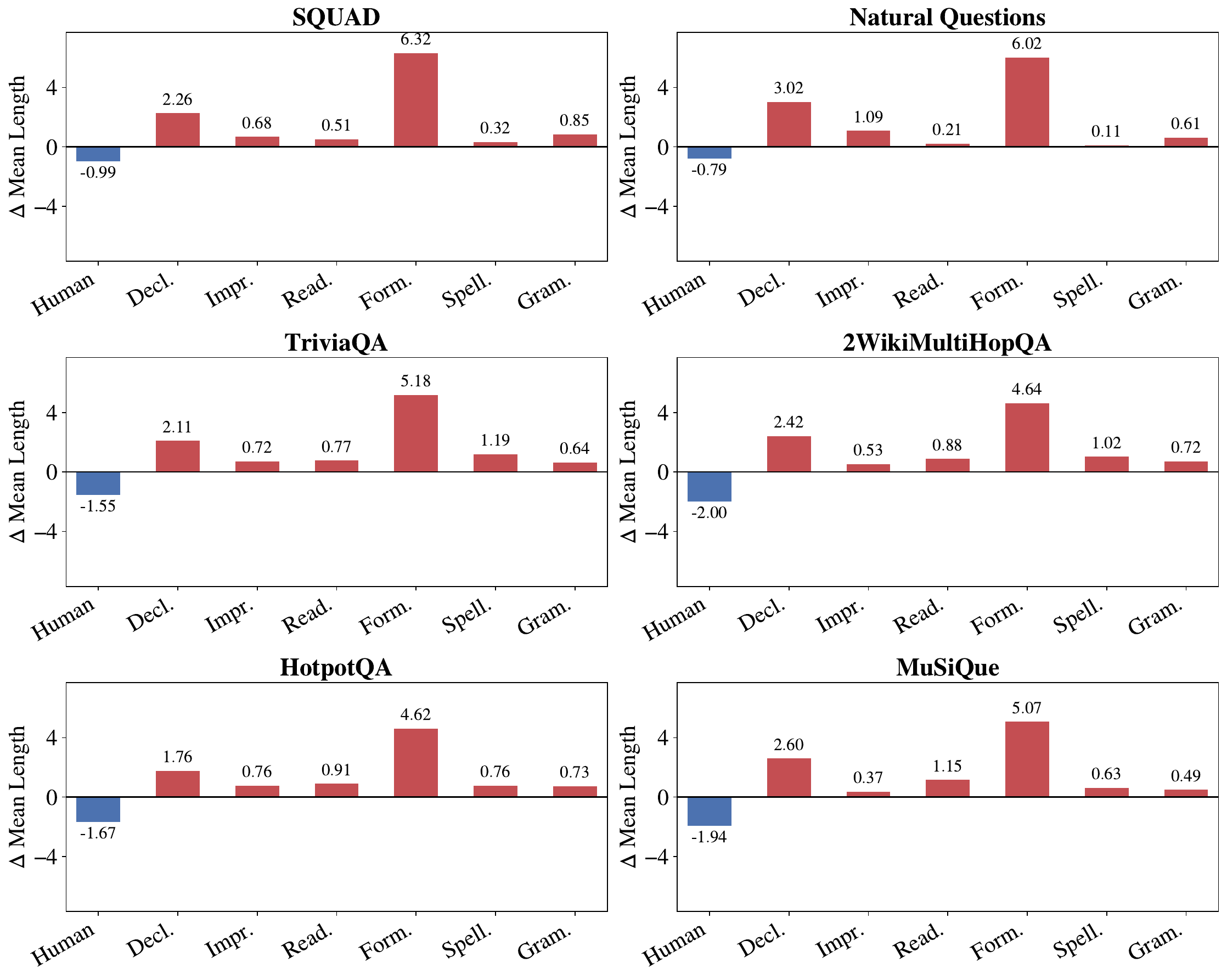}
\caption{Query length across all datasets}
\label{fig:query_length_all}
\end{figure*}

In this section, we visualize human query features in terms of loss, length, and semantic similarity.
As shown in Figure~\ref{fig:query_length_all}, human queries are consistently shorter than other query variations.
Consistent with their shorter length, Figure~\ref{fig:query_ppl_all} shows that human, readability, spelling, and grammar-error variations tend to have higher loss.
Lastly, semantic similarity to the original query shows a similar trend across datasets, indicating that human queries preserve the original query's meaning adequately.

\section{Usage of GenAI}
We used AI-assisted coding tools to support data analysis and visualization (e.g., drafting and debugging scripts for metric aggregation and figure/table generation). All code and reported results were reviewed and validated by the authors.

\input{Tables/human_instruction}

\vspace{5mm}

\input{Tables/robustness_accuracy_full_llama_appendix}
\input{Tables/robustness_accuracy_full_qwq_appendix}
\input{Tables/robustness_accuracy_human_full_llama_appendix}
\input{Tables/robustness_accuracy_human_full_qwq_appendix}

\input{Tables/Robustness_efficiency_full_llama_appendix}
\input{Tables/Robustness_efficiency_full_qwq_appendix}
\input{Tables/Robustness_efficiency_human_full_llama_appendix}
\input{Tables/Robustness_efficiency_human_full_qwq_appendix}

\begin{figure*}[t!]
\centering
\includegraphics[width=\linewidth]{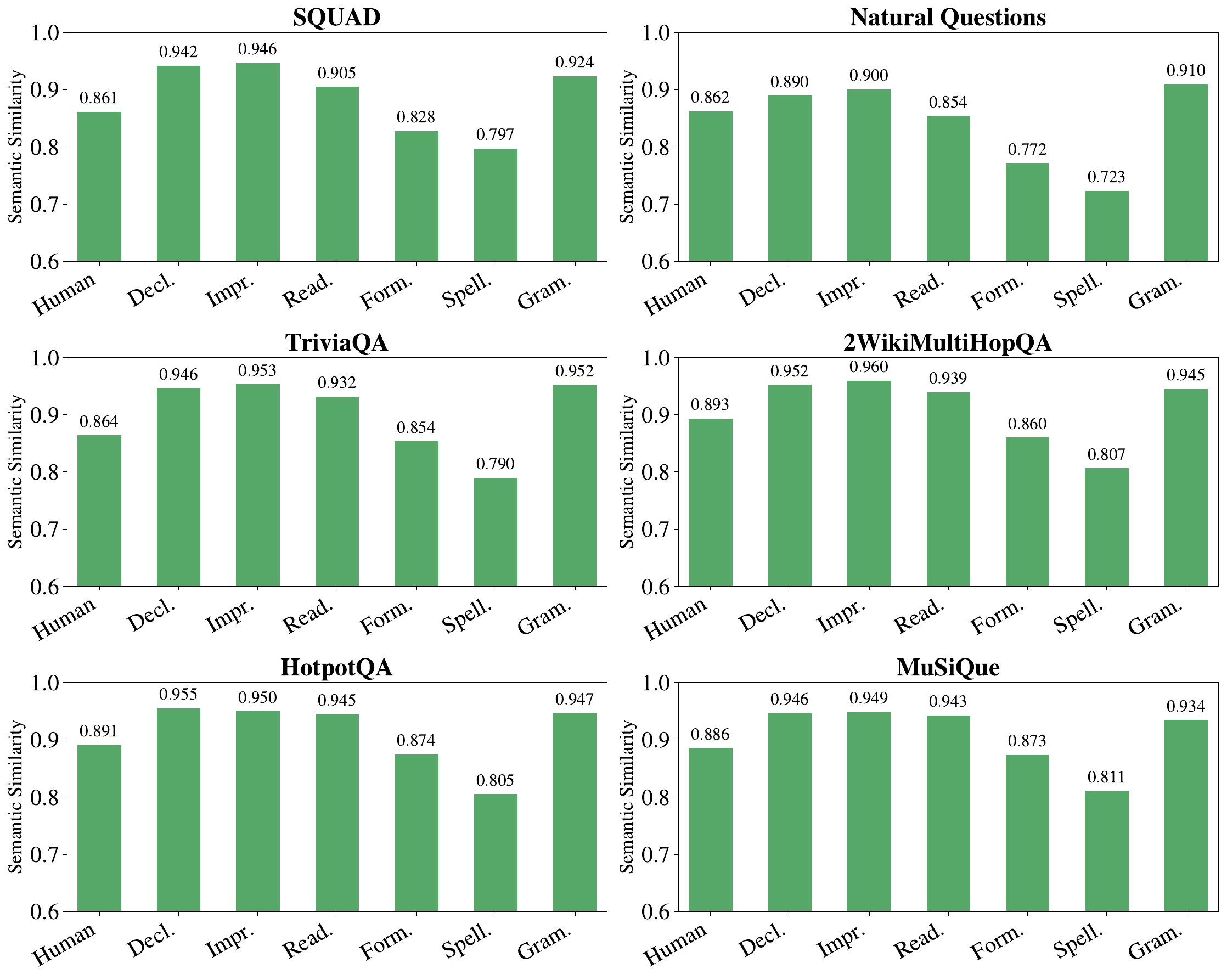}
\caption{Semantic similarity to original query across all datasets}
\label{fig:query_semantic_similarity_all}
\end{figure*}

%% file: Tables/prompt_formality.tex
\textbf{Formality Variation.} We adopt the lower-formality rewriting prompt proposed in prior work~\cite{out_of_style} to rewrite each query in a more casual style while preserving its original meaning.
\phantomsection
\begin{tcolorbox}[
  colback=white!95!gray,
  colframe=gray!50!black,
  rounded corners,
  title={\small Formality Variation Prompt Template}
]
\small
You are an AI assistant skilled at transforming formal queries into casual, everyday language.

Rewrite the following query so that it sounds very informal. Experiment with different colloquial openings, varied sentence constructions, and a mix of slang, idioms, and casual expressions throughout the sentence. Avoid using the same phrase repeatedly (e.g., 'hey, so like') and ensure the meaning remains unchanged.
\end{tcolorbox}

%% file: Tables/prompt_readability.tex
\begin{table*}[t]
\centering
\begin{tcolorbox}[
  colback=white!95!gray,
  colframe=gray!50!black,
  rounded corners,
  title={\small Readability Variation Prompt Template}
]
\small
1. Task Definition:\\
You are rewriting a query to make it significantly less readable while preserving the original semantic meaning as closely as possible.

2. Constraints \& Goals:\\

- Flesch Reading Ease Score: The rewritten text must have a Flesch score below 60 (preferably below 50).\\
- Semantic Similarity: The rewritten text must have SBERT similarity > 0.7 compared with the original query.\\
- Length: The rewritten text must remain approximately the same length as the original query (±10\%).\\
- Preserve Domain Terminology: Do not remove or drastically change domain-specific words, abbreviations, or technical terms (e.g., "IRS," "distance," etc.).\\
- Abbreviation: Do not expand abbreviations unless the original query already used the expanded form.\\
- No New Information: You must not add additional details beyond what the original query states.\\
- Question Format: Retain the form of a question if the original is posed as a question.

3. How to Increase Complexity:\\

- Lexical Changes: Use advanced or academic synonyms only for common words. For domain or key terms (e.g., "distance," "IRS," "tax"), keep the original term or use a very close synonym if necessary to maintain meaning.\\
- Syntactic Complexity: Introduce passive voice, nominalizations, embedded clauses, and parenthetical or subordinate phrases. Ensure the sentence flow is more formal and convoluted without changing the core meaning.\\
- Redundancy \& Formality: Employ circumlocution and excessively formal expressions (e.g., "due to the fact that" instead of "because") while avoiding any semantic drift.\\
- Dense, Indirect Construction: Favor longer phrases, indirect references, and wordiness. Avoid direct or simple phrasing.
\end{tcolorbox}
\caption{Prompt for lower-readability query generation.}
\label{tab:readability_prompt}
\end{table*}

%% file: Tables/prompt_type.tex
\textbf{Sentence Type Variation.}
This prompt rewrites the given query into a specified sentence type (declarative or imperative) while preserving meaning. We implement this prompt for both  declarative and imperative variations.
\phantomsection
\begin{tcolorbox}[
  colback=white!95!gray,
  colframe=gray!50!black,
  rounded corners,
  title={\small Sentence Type Variation Prompt Template}
]
\small
Task: Rewrite Query as \{\textit{sentence\_type}\} Sentence

Take the given query and rewrite it as a \{\textit{sentence\_type}\} sentence, changing only the sentence type while keeping the original meaning intact.

Do not add extra details or change the intent.
\end{tcolorbox}

%% file: Tables/prompt_grammar.tex
\textbf{Grammar Error Variation.} This prompt injects at least one grammar error to the original query. 
\phantomsection
\begin{tcolorbox}[
  colback=white!95!gray,
  colframe=gray!50!black,
  rounded corners,
  title={\small Grammar Error Variation Prompt Template}
]
\small
You are an AI assistant skilled at generating realistic grammatical errors in queries.\\
Rewrite the given query by introducing grammatical errors.\\
Constraints:\\
- Preserve the original meaning\\
- Introduce at least three grammatical errors per query\\
- Make it realistic and natural by varying the types of errors
\end{tcolorbox}

%% file: Tables/Dense_retriever.tex
\begin{table*}[t]
\centering
\small
\renewcommand{\arraystretch}{1.15} 
\setlength{\tabcolsep}{4.5pt}      

\begin{tabular}{llcccccccc}
\toprule[1.4pt]
\textbf{Dataset} & \textbf{Mtd.} & \textbf{Original} & \textbf{Human} & \textbf{Form.} & \textbf{Read.} & \textbf{Decl.} & \textbf{Impr.} & \textbf{Spell.} & \textbf{Gram.} \\
\midrule

\multicolumn{10}{l}{\textbf{\textit{Llama-3.1-8B}}} \\
\addlinespace[2pt]
2WIKI & LB    & 37.6 / 2.16 & 17.9 / 2.01 & 23 / 2.03 & 35 / 2.11 & 34 / 2.11 & 36 / 2.04 & 23.0 / 2.23 & 35.6 / 2.14 \\
2WIKI & GB & 20.2 / 8.24 & 12.1 / 7.27 & 19 / 7.15 & 18.6 / 7.41  & 19.4 / 7.36 & 20.4 / 7.46  & 9.8 / 6.53 & 21.2 / 7.03 \\
SQuAD & LB    & 27.6 / 1.69 & 14.5 / 1.82 & 19.4 / 1.80 & 25 / 1.81 & 25.4 / 1.77 & 27.0 / 1.81 & 17.6 / 1.77 & 24.4 / 1.80 \\
SQuAD & GB & 16.4 / 1.93 & 11.0 / 1.95 & 15.4 / 4.90 & 16.2 / 4.90  & 16.8 / 4.87 & 15.8 / 3.92 & 14.8 / 1.94 & 13.8 / 1.92 \\

\addlinespace[2pt]
\midrule
\addlinespace[2pt]

\multicolumn{10}{l}{\textbf{\textit{QwQ-32B}}} \\
\addlinespace[2pt]
2WIKI & LB    & 26.6 / 2.24 & 20.0 / 2.25 & 21.6 / 2.31 & 26.8 / 2.28 & 27.2 / 2.15 & 27.8 / 2.22 & 21.0 / 2.23 & 24.4 / 2.23 \\
2WIKI & GB & 59.2 / 4.93 & 43.4 / 4.96 & 55.2 / 4.85 & 57.2 / 4.91 & 59 / 4.75 & 56.4 / 5.11 & 45.8 / 4.40 & 54.6 / 4.68 \\
SQuAD & LB    & 27.2 / 2.03 & 21.5 / 2.08 & 23.6 / 2.14 & 33.3 / 2.13  & 27.2 / 1.99 & 30.4 / 1.99 & 22.4 / 2.26 & 27 / 2.00 \\
SQuAD & GB & 38.2 / 1.27 & 27.5 / 1.35 & 34.8 / 1.32  & 35.2 / 1.30 & 36.2 / 1.34 & 35.8 / 1.35 & 33.6 / 1.32 & 32.6 / 1.28 \\
\bottomrule[1.4pt]
\end{tabular}
\caption{ \textbf{Results for Dense Retriever (Contriever).} We report results as (accuracy / retrieval count), comparing results between NR (NR), logit-based (LB), and generation-based (GB) method.}
\label{tab:contriever_results}
\end{table*}

%% file: Tables/fliprate_diff_temp.tex
\begin{table}[t!]
\centering
\small
\setlength{\tabcolsep}{6pt}
\begin{tabular}{cccc}
\toprule
\textbf{Temp 0.0} & \textbf{Temp 0.3} & \textbf{Temp 0.7} & \textbf{Temp 1.0} \\
\midrule
44.60\% & 45.33\% & 45.45\% & 42.67\% \\
\bottomrule
\end{tabular}
\caption{Flip rates (\%) across different decoding temperatures on SQuAD.}
\label{tab:flip_rates_temperature}
\end{table}

%% file: Tables/Closed_llm.tex
\begin{table}[t]
\centering
\small
\setlength{\tabcolsep}{5pt}
\renewcommand{\arraystretch}{1.1}
\begin{tabular}{l l c c}
\toprule [1.3pt]
\textbf{Model} & \textbf{Dataset} & \textbf{Original} & \textbf{Human} \\
\midrule
Llama-3.1-8B  & 2WIKI & 20.0 / 7.34 & 14.2 / 7.09 \\
Llama-3.1-8B  & SQuAD & 16.8 / 1.92 & 10.5 / 1.95 \\
QwQ-32B       & 2WIKI & 61.0 / 4.32 & 57.1 / 4.06 \\
QwQ-32B       & SQuAD & 33.0 / 1.40 & 28.5 / 1.42 \\
\midrule
GPT-4o mini   & 2WIKI & 27.0 / 3.77 & 14.6 / 4.95 \\
GPT-4o mini   & SQuAD & 17.0 / 2.34 & 16.0 / 2.34 \\
\bottomrule[1.3pt]
\end{tabular}
\caption{\textbf{Closed- and open-source model results} We report results as (accuracy / retrieval count).}
\label{tab:closed_model_results}
\end{table}

%% file: Tables/f1.tex
\begin{table}[t]
\centering
\small
\setlength{\tabcolsep}{6pt}
\renewcommand{\arraystretch}{1.1}

\begin{tabular}{ l l cc}
\toprule
\textbf{Dataset}  & \textbf{Variation} & \textbf{F1} & \textbf{InAccuracy} \\
\midrule
2WIKI & Original & 0.396 & 37.0 \\
2WIKI & Human    & 0.283 & 23.8 \\
SQuAD & Original & 0.352 & 29.2 \\
SQuAD & Human    & 0.260 & 17.0 \\
\bottomrule
\end{tabular}
\caption{\textbf{Comparison between InAccuracy and token-level F1 scores} We report the scores for logit-based method with Llama-3.1-8B across datasets and query variations.}
\label{tab:inacc_vs_f1}
\end{table}

%% file: Tables/cost_additional.tex
\begin{table}[t]
\centering
\small\setlength{\tabcolsep}{4pt}
\begin{adjustbox}{width=\linewidth}
\begin{tabular}{llccc}
\toprule [1.3pt]
Model & Method & Original & Human & Spelling-error \\
\midrule
\multirow{2}{*}{Llama-3.1-8B}
& LB    & 177  & 229  & 268  \\
& GB & 2849 & 2349 & 2264 \\
\midrule
\multirow{2}{*}{QwQ-32B}
& LB    & 235  & 235  & 250  \\
& GB  & 4680 & 3759 & 4251  \\
\bottomrule[1.3pt]
\end{tabular}
\end{adjustbox}
\caption{\textbf{Token usage across query variation types on 2WIKI}. Token usage is measured as the total number of tokens in the model's full generated output.}
\label{tab:token_usage}
\end{table}

%% file: Tables/multi_run_llama.tex
\begin{table*}[t]
\centering
\small
\renewcommand{\arraystretch}{1.15}
\setlength{\tabcolsep}{3.5pt}
\resizebox{\textwidth}{!}{%
\begin{tabular}{llcccccccc}
\toprule[1.2pt]
\textbf{Dataset} & \textbf{Mtd.} &
\textbf{Orig.} & \textbf{Human} & \textbf{Form.} & \textbf{Read.} &
\textbf{Decl.} & \textbf{Impr.} & \textbf{Spell.} & \textbf{Gram.} \\
\midrule
\multicolumn{10}{l}{\textbf{\textit{Llama-3.1-8B}}} \\
\midrule

\multirow{3}{*}{SQuAD} & SR & $\mathbf{30.1}{\scriptstyle\pm1.1}$ & $19.3{\scriptstyle\pm1.0}$ & $22.2{\scriptstyle\pm0.9}$ & $26.3{\scriptstyle\pm0.7}$ & \underline{$26.6{\scriptstyle\pm0.5}$} & $29.3{\scriptstyle\pm0.9}$ & $18.8{\scriptstyle\pm1.0}$ & \underline{$26.7{\scriptstyle\pm0.9}$} \\
  & LB & $\mathbf{28.8}{\scriptstyle\pm0.5}$ & $17.2{\scriptstyle\pm0.3}$ & $22.2{\scriptstyle\pm0.6}$ & $26.3{\scriptstyle\pm2.0}$ & \underline{$26.9{\scriptstyle\pm0.7}$} & $28.3{\scriptstyle\pm2.1}$ & $21.2{\scriptstyle\pm0.7}$ & $25.9{\scriptstyle\pm0.9}$ \\
  & GB & $15.7{\scriptstyle\pm1.8}$ & $9.8{\scriptstyle\pm1.6}$ & $\mathbf{16.0}{\scriptstyle\pm2.0}$ & $15.1{\scriptstyle\pm1.7}$ & \underline{$15.8{\scriptstyle\pm2.0}$} & $14.8{\scriptstyle\pm0.5}$ & $14.1{\scriptstyle\pm1.8}$ & $15.3{\scriptstyle\pm0.9}$ \\
\midrule
\multirow{3}{*}{NQ} & SR & $35.3{\scriptstyle\pm0.1}$ & $28.0{\scriptstyle\pm0.0}$ & $\mathbf{40.1}{\scriptstyle\pm1.3}$ & $35.2{\scriptstyle\pm0.6}$ & \underline{$38.8{\scriptstyle\pm1.1}$} & $37.1{\scriptstyle\pm1.8}$ & $28.1{\scriptstyle\pm1.0}$ & $37.2{\scriptstyle\pm0.0}$ \\
  & LB & $41.8{\scriptstyle\pm0.3}$ & $33.6{\scriptstyle\pm0.6}$ & $44.7{\scriptstyle\pm1.8}$ & $43.8{\scriptstyle\pm0.6}$ & \underline{$46.1{\scriptstyle\pm0.8}$} & $\mathbf{47.2}{\scriptstyle\pm1.2}$ & $35.6{\scriptstyle\pm0.3}$ & $44.9{\scriptstyle\pm1.6}$ \\
  & GB & $33.4{\scriptstyle\pm0.3}$ & $32.8{\scriptstyle\pm0.0}$ & $32.6{\scriptstyle\pm0.3}$ & $34.5{\scriptstyle\pm0.7}$ & $35.5{\scriptstyle\pm1.7}$ & \underline{$37.8{\scriptstyle\pm1.8}$} & $29.7{\scriptstyle\pm1.9}$ & $\mathbf{38.5}{\scriptstyle\pm1.0}$ \\
\midrule
\multirow{3}{*}{TriviaQA} & SR & $\mathbf{66.3}{\scriptstyle\pm0.8}$ & $55.0{\scriptstyle\pm1.4}$ & $63.1{\scriptstyle\pm1.1}$ & $62.3{\scriptstyle\pm0.3}$ & $63.5{\scriptstyle\pm0.6}$ & \underline{$65.2{\scriptstyle\pm0.9}$} & $53.9{\scriptstyle\pm0.8}$ & $63.7{\scriptstyle\pm0.7}$ \\
  & LB & $\mathbf{68.3}{\scriptstyle\pm0.4}$ & $56.2{\scriptstyle\pm1.8}$ & \underline{$66.2{\scriptstyle\pm0.7}$} & $64.7{\scriptstyle\pm0.9}$ & $64.3{\scriptstyle\pm0.8}$ & $66.1{\scriptstyle\pm0.6}$ & $60.5{\scriptstyle\pm0.8}$ & $66.0{\scriptstyle\pm1.6}$ \\
  & GB & $53.9{\scriptstyle\pm0.4}$ & $47.5{\scriptstyle\pm0.2}$ & $\mathbf{57.1}{\scriptstyle\pm1.8}$ & $52.9{\scriptstyle\pm0.7}$ & $52.6{\scriptstyle\pm0.6}$ & $53.8{\scriptstyle\pm1.1}$ & $47.0{\scriptstyle\pm0.6}$ & \underline{$55.0{\scriptstyle\pm0.6}$} \\
\midrule
\multirow{3}{*}{2Wiki} & SR & $\mathbf{39.1}{\scriptstyle\pm0.8}$ & $19.8{\scriptstyle\pm0.3}$ & $26.6{\scriptstyle\pm1.4}$ & $34.1{\scriptstyle\pm0.8}$ & $34.9{\scriptstyle\pm0.8}$ & \underline{$36.0{\scriptstyle\pm1.1}$} & $21.6{\scriptstyle\pm0.3}$ & $35.6{\scriptstyle\pm0.9}$ \\
  & LB & $\mathbf{41.0}{\scriptstyle\pm1.4}$ & $23.8{\scriptstyle\pm0.2}$ & $29.3{\scriptstyle\pm0.9}$ & \underline{$40.5{\scriptstyle\pm0.4}$} & $38.8{\scriptstyle\pm0.6}$ & $39.4{\scriptstyle\pm2.0}$ & $21.2{\scriptstyle\pm0.6}$ & $38.9{\scriptstyle\pm2.1}$ \\
  & GB & $20.6{\scriptstyle\pm1.0}$ & $13.5{\scriptstyle\pm0.9}$ & $21.0{\scriptstyle\pm0.2}$ & $\mathbf{26.3}{\scriptstyle\pm0.4}$ & $21.2{\scriptstyle\pm0.3}$ & \underline{$25.9{\scriptstyle\pm1.0}$} & $17.3{\scriptstyle\pm0.3}$ & $22.6{\scriptstyle\pm1.2}$ \\
\midrule
\multirow{3}{*}{HotpotQA} & SR & $\mathbf{42.1}{\scriptstyle\pm1.6}$ & $31.7{\scriptstyle\pm2.3}$ & $36.6{\scriptstyle\pm0.7}$ & $37.7{\scriptstyle\pm1.7}$ & $37.4{\scriptstyle\pm0.8}$ & \underline{$40.4{\scriptstyle\pm0.4}$} & $31.7{\scriptstyle\pm0.1}$ & $40.3{\scriptstyle\pm1.0}$ \\
  & LB & $\mathbf{40.8}{\scriptstyle\pm0.7}$ & $32.5{\scriptstyle\pm0.7}$ & $38.8{\scriptstyle\pm1.1}$ & \underline{$40.0{\scriptstyle\pm0.2}$} & $36.8{\scriptstyle\pm0.9}$ & $26.5{\scriptstyle\pm1.6}$ & $31.1{\scriptstyle\pm0.9}$ & $39.3{\scriptstyle\pm0.6}$ \\
  & GB & $23.6{\scriptstyle\pm0.6}$ & $23.0{\scriptstyle\pm0.7}$ & $31.4{\scriptstyle\pm1.6}$ & $\mathbf{34.1}{\scriptstyle\pm0.5}$ & $30.2{\scriptstyle\pm0.6}$ & $32.2{\scriptstyle\pm1.6}$ & $24.5{\scriptstyle\pm1.1}$ & \underline{$33.0{\scriptstyle\pm1.6}$} \\
\midrule
\multirow{3}{*}{MuSiQue} & SR & $8.2{\scriptstyle\pm0.4}$ & $5.3{\scriptstyle\pm0.6}$ & $8.1{\scriptstyle\pm1.2}$ & $7.1{\scriptstyle\pm0.6}$ & $\mathbf{9.7}{\scriptstyle\pm0.8}$ & $8.8{\scriptstyle\pm2.3}$ & $5.2{\scriptstyle\pm0.9}$ & \underline{$9.3{\scriptstyle\pm1.1}$} \\
  & LB & \underline{$12.1{\scriptstyle\pm1.2}$} & $7.3{\scriptstyle\pm2.1}$ & $10.9{\scriptstyle\pm1.1}$ & $7.5{\scriptstyle\pm1.6}$ & $\mathbf{12.3}{\scriptstyle\pm0.7}$ & $11.1{\scriptstyle\pm1.5}$ & $6.3{\scriptstyle\pm1.1}$ & $10.2{\scriptstyle\pm1.2}$ \\
  & GB & $\mathbf{11.9}{\scriptstyle\pm0.1}$ & $5.5{\scriptstyle\pm1.8}$ & $10.1{\scriptstyle\pm0.4}$ & $7.1{\scriptstyle\pm1.8}$ & \underline{$11.6{\scriptstyle\pm1.5}$} & $10.1{\scriptstyle\pm1.2}$ & $5.8{\scriptstyle\pm0.9}$ & $8.3{\scriptstyle\pm1.1}$ \\
\bottomrule[1.2pt]
\end{tabular}%
}
\caption{Multi-run results for Llama-3.1-8B (mean $\pm$ std over 3 runs). Methods: NR=Naive Retrieval, LB=Logit-based, GB=Generation-based. For each row, the best-performing result is shown in bold and the second-best result is underlined.}
\label{tab:multi_run_llama_variation_cols}
\end{table*}

%% file: Tables/multi_run_qwq.tex
\begin{table*}[t]
\centering
\small
\renewcommand{\arraystretch}{1.15}
\setlength{\tabcolsep}{4.2pt}
\resizebox{\textwidth}{!}{%
\begin{tabular}{llcccccccc}
\toprule[1.2pt]
\textbf{Dataset} & \textbf{Mtd.} &
\textbf{Orig.} & \textbf{Human} & \textbf{Form.} & \textbf{Read.} &
\textbf{Decl.} & \textbf{Impr.} & \textbf{Spell.} & \textbf{Gram.} \\
\midrule
\multicolumn{10}{l}{\textbf{\textit{QwQ-32B}}} \\
\midrule

\multirow{3}{*}{SQuAD} &NR & $\mathbf{28.8}{\scriptstyle\pm0.3}$ & $18.3{\scriptstyle\pm0.8}$ & \underline{$22.5{\scriptstyle\pm0.4}$} & $18.1{\scriptstyle\pm1.0}$ & $19.0{\scriptstyle\pm1.1}$ & $17.7{\scriptstyle\pm1.3}$ & $17.1{\scriptstyle\pm2.5}$ & $18.7{\scriptstyle\pm1.0}$ \\
  & LB & $19.0{\scriptstyle\pm0.3}$ & $15.8{\scriptstyle\pm2.1}$ & \underline{$25.4{\scriptstyle\pm0.7}$} & $18.0{\scriptstyle\pm0.3}$ & $\mathbf{28.4}{\scriptstyle\pm1.1}$ & $18.2{\scriptstyle\pm1.4}$ & $16.2{\scriptstyle\pm1.1}$ & $18.8{\scriptstyle\pm0.3}$ \\
  & GB & $\mathbf{38.2}{\scriptstyle\pm0.7}$ & $28.8{\scriptstyle\pm1.5}$ & $29.5{\scriptstyle\pm0.5}$ & \underline{$32.1{\scriptstyle\pm1.6}$} & \underline{$33.2{\scriptstyle\pm3.1}$} & $31.6{\scriptstyle\pm5.1}$ & $29.2{\scriptstyle\pm7.4}$ & \underline{$32.1{\scriptstyle\pm3.0}$} \\
\midrule
\multirow{3}{*}{NQ} &NR & $42.3{\scriptstyle\pm1.2}$ & $37.9{\scriptstyle\pm1.4}$ & $44.1{\scriptstyle\pm0.8}$ & \underline{$45.1{\scriptstyle\pm1.9}$} & $43.4{\scriptstyle\pm0.3}$ & $44.7{\scriptstyle\pm0.8}$ & $35.0{\scriptstyle\pm1.8}$ & $\mathbf{45.7}{\scriptstyle\pm0.7}$ \\
  & LB & $33.1{\scriptstyle\pm1.0}$ & $28.7{\scriptstyle\pm0.2}$ & $35.8{\scriptstyle\pm0.2}$ & $36.0{\scriptstyle\pm0.5}$ & $\mathbf{37.5}{\scriptstyle\pm1.0}$ & $36.4{\scriptstyle\pm1.2}$ & $29.6{\scriptstyle\pm1.7}$ & $\mathbf{37.5}{\scriptstyle\pm0.5}$ \\
  & GB & $53.5{\scriptstyle\pm8.6}$ & $41.2{\scriptstyle\pm10.7}$ & $49.9{\scriptstyle\pm0.4}$ & \underline{$67.8{\scriptstyle\pm0.7}$} & $67.4{\scriptstyle\pm0.7}$ & $\mathbf{68.0}{\scriptstyle\pm0.4}$ & $59.6{\scriptstyle\pm0.3}$ & $52.4{\scriptstyle\pm0.6}$ \\
\midrule
\multirow{3}{*}{TriviaQA} &NR & $\mathbf{65.9}{\scriptstyle\pm1.1}$ & $54.0{\scriptstyle\pm0.9}$ & $62.1{\scriptstyle\pm1.1}$ & $62.9{\scriptstyle\pm0.8}$ & $63.1{\scriptstyle\pm0.3}$ & \underline{$65.5{\scriptstyle\pm1.1}$} & $57.9{\scriptstyle\pm0.6}$ & $65.3{\scriptstyle\pm1.1}$ \\
  & LB & $\mathbf{70.2}{\scriptstyle\pm1.4}$ & $62.7{\scriptstyle\pm1.8}$ & $66.9{\scriptstyle\pm1.0}$ & $62.4{\scriptstyle\pm0.8}$ & $67.7{\scriptstyle\pm0.1}$ & $68.4{\scriptstyle\pm1.4}$ & $63.8{\scriptstyle\pm1.1}$ & \underline{$69.3{\scriptstyle\pm0.8}$} \\
  & GB & $\mathbf{77.6}{\scriptstyle\pm2.3}$ & $69.2{\scriptstyle\pm1.1}$ & $72.8{\scriptstyle\pm1.1}$ & $72.0{\scriptstyle\pm1.3}$ & \underline{$74.6{\scriptstyle\pm0.3}$} & $73.8{\scriptstyle\pm0.3}$ & $72.0{\scriptstyle\pm0.5}$ & $76.0{\scriptstyle\pm0.7}$ \\
\midrule
\multirow{3}{*}{2Wiki} &NR & $28.6{\scriptstyle\pm1.6}$ & $22.5{\scriptstyle\pm0.6}$ & $19.8{\scriptstyle\pm2.2}$ & $25.9{\scriptstyle\pm2.1}$ & \underline{$27.0{\scriptstyle\pm0.3}$} & $\mathbf{28.9}{\scriptstyle\pm0.7}$ & $16.7{\scriptstyle\pm1.2}$ & $26.1{\scriptstyle\pm2.3}$ \\
  & LB & \underline{$31.8{\scriptstyle\pm1.3}$} & $22.8{\scriptstyle\pm1.7}$ & $26.4{\scriptstyle\pm0.3}$ & $\mathbf{32.6}{\scriptstyle\pm0.5}$ & $30.4{\scriptstyle\pm0.9}$ & $30.1{\scriptstyle\pm0.6}$ & $22.3{\scriptstyle\pm0.9}$ & $29.7{\scriptstyle\pm1.0}$ \\
  & GB & $62.2{\scriptstyle\pm1.7}$ & $57.5{\scriptstyle\pm0.4}$ & $59.4{\scriptstyle\pm0.6}$ & $\mathbf{63.9}{\scriptstyle\pm1.3}$ & $63.4{\scriptstyle\pm2.3}$ & \underline{$63.7{\scriptstyle\pm0.6}$} & $49.0{\scriptstyle\pm1.6}$ & $60.9{\scriptstyle\pm1.3}$ \\
\midrule
\multirow{3}{*}{HotpotQA} &NR & $\mathbf{28.7}{\scriptstyle\pm1.0}$ & $21.8{\scriptstyle\pm2.3}$ & $20.3{\scriptstyle\pm0.8}$ & $24.0{\scriptstyle\pm2.0}$ & \underline{$26.9{\scriptstyle\pm1.3}$} & $25.1{\scriptstyle\pm0.8}$ & $21.1{\scriptstyle\pm1.3}$ & $24.4{\scriptstyle\pm0.9}$ \\
  & LB & \underline{$33.2{\scriptstyle\pm0.4}$} & $28.0{\scriptstyle\pm1.8}$ & $30.1{\scriptstyle\pm0.9}$ & $31.5{\scriptstyle\pm1.4}$ & $31.1{\scriptstyle\pm0.7}$ & $31.3{\scriptstyle\pm1.1}$ & $26.5{\scriptstyle\pm0.7}$ & $\mathbf{33.4}{\scriptstyle\pm0.3}$ \\
  & GB & $\mathbf{60.2}{\scriptstyle\pm0.6}$ & $52.5{\scriptstyle\pm0.6}$ & $55.4{\scriptstyle\pm1.2}$ & $58.0{\scriptstyle\pm0.9}$ & $57.4{\scriptstyle\pm0.8}$ & \underline{$58.6{\scriptstyle\pm1.4}$} & $51.7{\scriptstyle\pm0.7}$ & $57.8{\scriptstyle\pm0.8}$ \\
\midrule
\multirow{3}{*}{MuSiQue} &NR & \underline{$6.5{\scriptstyle\pm1.5}$} & $4.3{\scriptstyle\pm1.3}$ & $5.8{\scriptstyle\pm0.9}$ & \underline{$6.5{\scriptstyle\pm1.1}$} & $\mathbf{7.3}{\scriptstyle\pm0.8}$ & $5.5{\scriptstyle\pm0.9}$ & $3.9{\scriptstyle\pm0.5}$ & $5.1{\scriptstyle\pm1.0}$ \\
  & LB & $8.9{\scriptstyle\pm0.3}$ & $7.0{\scriptstyle\pm1.5}$ & $8.0{\scriptstyle\pm0.5}$ & $8.5{\scriptstyle\pm1.9}$ & \underline{$9.5{\scriptstyle\pm0.5}$} & $\mathbf{9.7}{\scriptstyle\pm0.2}$ & $5.0{\scriptstyle\pm1.2}$ & $7.3{\scriptstyle\pm0.1}$ \\
  & GB & $\mathbf{33.8}{\scriptstyle\pm0.4}$ & $27.0{\scriptstyle\pm0.3}$ & $28.4{\scriptstyle\pm0.6}$ & $29.2{\scriptstyle\pm0.8}$ & $31.2{\scriptstyle\pm1.1}$ & \underline{$32.0{\scriptstyle\pm1.6}$} & $24.0{\scriptstyle\pm0.3}$ & $25.4{\scriptstyle\pm0.2}$ \\
\bottomrule[1.2pt]
\end{tabular}
}
\caption{Multi-run results for QwQ-32B (mean $\pm$ std over 3 runs). Methods: NR=Naive RAG, LB=Logit-based, GB=Generation-based.For each row, the best-performing result is shown in bold and the second-best result is underlined.}
\label{tab:multi_run_qwq_variation_cols}
\end{table*}

%% file: Tables/human_instruction.tex
\begin{table*}[htbp]
\centering
\begin{tabular}{@{}p{\linewidth}@{}}
\toprule[1.7pt]
\textbf{Human Annotation Instruction}\\
\midrule

\textbf{Task Overview}
\begin{enumerate}[leftmargin=2em, itemsep=0.2em]
  \item You will be given a query (question) and its corresponding answer.
  \item Your task is to rewrite the given query — imagine how you would ask the same question if you were speaking to a large language model (like ChatGPT).
\end{enumerate}

\vspace{0.6em}
\textbf{What You Should Do}
\begin{enumerate}[leftmargin=2em, itemsep=0.2em]

  \item Understand the original query.
  \begin{itemize}[leftmargin=2em, label=$\circ$, itemsep=0.1em]
    \item Identify what the user is trying to find out.
    \item Grasp the intent and main topic clearly.
  \end{itemize}

  \item Rewrite the query in a new way while keeping the same meaning.
  \begin{itemize}[leftmargin=2em, label=$\circ$, itemsep=0.1em]
    \item You may change the tone, phrasing, or structure.
    \item Information loss, addition, or simplification is acceptable.
    \item The goal is to make the rewritten query similar in meaning but different in expression.
    \item Make sure that your rewritten query can still be answered appropriately with the same answer as the original query.
  \end{itemize}

  \item Incorporate your personal style.
  \begin{itemize}[leftmargin=2em, label=$\circ$, itemsep=0.1em]
    \item You can make the question sound more casual, detailed, concise, or clear.
    \item Different writing styles and tones are encouraged.
  \end{itemize}

  \item Avoid copying the original query.
  \begin{itemize}[leftmargin=2em, label=$\circ$, itemsep=0.1em]
    \item Do not simply rephrase it with only minor surface changes.
    \item The rewritten query should feel genuinely reworded and natural.
  \end{itemize}

  \item Use the answer only for context.
  \begin{itemize}[leftmargin=2em, label=$\circ$, itemsep=0.1em]
    \item The answer helps you understand what the original query intended to ask.
    \item You should not change your rewrite based on the answer’s content.
    \item Your task is to rewrite the question, not the answer.
  \end{itemize}

\end{enumerate}

\\[-0.3em]
\bottomrule[1.7pt]
\end{tabular}
\caption{Human Annotation Instruction: Query Rewriting Task.}
\label{tab:human_instruction}
\end{table*}

%% file: Tables/robustness_accuracy_full_llama_appendix.tex
\begin{table*}[t]
\centering
\small
\renewcommand{\arraystretch}{0.8}
\resizebox{\textwidth}{!}{
\begin{tabular}{%
p{1.2cm}  p{1.2cm}
cccccc c}
\toprule[1.5pt]

\multirow{2}{*}{\textbf{Dataset}}  & \multirow{2}{*}{\textbf{Method}}
& \multicolumn{6}{c}{\textbf{Similarity} $\uparrow$} & \multirow{2}{*}{\textbf{Diversity} $\downarrow$} \\
\cmidrule(lr){3-8}
& &  Form. & Read. & Decl. & Impr. & Spell. & Gram. & \\

\midrule
\multicolumn{9}{c}{\textbf{Llama-3.1-8B}} \\
\midrule

\multirow{4}{*}{\textbf{SQuAD}} & WR & \Sim{0.568} & \Sim{0.580} & \Sim{0.598} & \textbf{\Sim{0.635}} & \Sim{0.538} & \Sim{0.584} & \textbf{\Div{0.459}} \\
& NR & \Sim{0.509} & \Sim{0.552} & \Sim{0.606} & \textbf{\Sim{0.631}} & \Sim{0.466} & \Sim{0.578} & \textbf{\Div{0.499}} \\
& LB & \Sim{0.551} & \Sim{0.539} & \Sim{0.593} & \textbf{\Sim{0.634}} & \Sim{0.552} & \Sim{0.599} & \textbf{\Div{0.469}} \\
& GB & \Sim{0.426} & \Sim{0.408} & \textbf{\Sim{0.433}} & \Sim{0.430} & \Sim{0.411} & \Sim{0.414} & \textbf{\Div{0.585}} \\
\cmidrule(lr){1-9}
\multirow{4}{*}{\textbf{NQ}} & WR & \Sim{0.672} & \Sim{0.670} & \Sim{0.652} & \textbf{\Sim{0.712}} & \Sim{0.610} & \Sim{0.675} & \textbf{\Div{0.373}} \\
& NR & \Sim{0.311} & \Sim{0.483} & \textbf{\Sim{0.499}} & \Sim{0.494} & \Sim{0.425} & \Sim{0.321} & \textbf{\Div{0.621}} \\
& LB & \Sim{0.706} & \Sim{0.686} & \Sim{0.708} & \textbf{\Sim{0.729}} & \Sim{0.626} & \Sim{0.724} & \textbf{\Div{0.346}} \\
& GB & \Sim{0.382} & \textbf{\Sim{1.000}} & \textbf{\Sim{1.000}} & \textbf{\Sim{1.000}} & \textbf{\Sim{1.000}} & \Sim{0.435} & \textbf{\Div{0.347}} \\
\cmidrule(lr){1-9}
\multirow{4}{*}{\textbf{TriviaQA}} & WR & \Sim{0.767} & \Sim{0.773} & \Sim{0.762} & \textbf{\Sim{0.787}} & \Sim{0.734} & \Sim{0.763} & \textbf{\Div{0.268}} \\
& NR & \Sim{0.700} & \Sim{0.716} & \Sim{0.725} & \textbf{\Sim{0.758}} & \Sim{0.648} & \Sim{0.738} & \textbf{\Div{0.327}} \\
& LB & \Sim{0.781} & \textbf{\Sim{0.799}} & \Sim{0.769} & \Sim{0.798} & \Sim{0.727} & \Sim{0.785} & \textbf{\Div{0.259}} \\
& GB & \Sim{0.528} & \Sim{0.537} & \Sim{0.543} & \Sim{0.547} & \Sim{0.504} & \textbf{\Sim{0.549}} & \textbf{\Div{0.467}} \\
\cmidrule(lr){1-9}
\multirow{4}{*}{\textbf{2WIKI}} & WR & \Sim{0.561} & \Sim{0.595} & \Sim{0.630} & \textbf{\Sim{0.660}} & \Sim{0.520} & \Sim{0.593} & \textbf{\Div{0.442}} \\
& NR & \Sim{0.568} & \Sim{0.628} & \Sim{0.644} & \textbf{\Sim{0.651}} & \Sim{0.530} & \Sim{0.641} & \textbf{\Div{0.432}} \\
& LB & \Sim{0.560} & \Sim{0.628} & \Sim{0.649} & \textbf{\Sim{0.655}} & \Sim{0.537} & \Sim{0.627} & \textbf{\Div{0.414}} \\
& GB & \Sim{0.423} & \textbf{\Sim{0.458}} & \Sim{0.300} & \Sim{0.403} & \Sim{0.313} & \Sim{0.385} & \textbf{\Div{0.667}} \\
\cmidrule(lr){1-9}
\multirow{4}{*}{\textbf{HotpotQA}} & WR & \Sim{0.536} & \Sim{0.566} & \Sim{0.547} & \textbf{\Sim{0.580}} & \Sim{0.509} & \Sim{0.557} & \textbf{\Div{0.484}} \\
& NR & \Sim{0.532} & \Sim{0.542} & \Sim{0.546} & \Sim{0.543} & \Sim{0.500} & \textbf{\Sim{0.564}} & \textbf{\Div{0.500}} \\
& LB & \Sim{0.594} & \textbf{\Sim{0.628}} & \Sim{0.591} & \Sim{0.625} & \Sim{0.557} & \Sim{0.617} & \textbf{\Div{0.415}} \\
& GB & \Sim{0.479} & \textbf{\Sim{0.502}} & \Sim{0.481} & \Sim{0.483} & \Sim{0.456} & \Sim{0.489} & \textbf{\Div{0.523}} \\
\cmidrule(lr){1-9}
\multirow{4}{*}{\textbf{MuSiQue}} & WR & \Sim{0.469} & \Sim{0.489} & \Sim{0.491} & \textbf{\Sim{0.512}} & \Sim{0.423} & \Sim{0.480} & \textbf{\Div{0.560}} \\
& NR & \Sim{0.427} & \Sim{0.454} & \textbf{\Sim{0.471}} & \Sim{0.459} & \Sim{0.385} & \Sim{0.452} & \textbf{\Div{0.588}} \\
& LB & \Sim{0.513} & \Sim{0.480} & \Sim{0.502} & \textbf{\Sim{0.528}} & \Sim{0.428} & \Sim{0.481} & \textbf{\Div{0.537}} \\
& GB & \Sim{0.424} & \Sim{0.426} & \Sim{0.419} & \textbf{\Sim{0.438}} & \Sim{0.389} & \Sim{0.392} & \textbf{\Div{0.576}} \\

\bottomrule[1.5pt]
\end{tabular}
}
\caption{
\textbf{Answer similarity to the original query and diversity across perturbation types (excluding human rewrites).}We compare w/o retrieval (WR), naive RAG (NR), logit-based (LB), and generation-based (GB).
Darker shading indicates more desirable outcomes: higher Similarity and lower Diversity.}
\label{tab:robustness_simdiv_nohuman_llama}
\end{table*}

%% file: Tables/robustness_accuracy_full_qwq_appendix.tex
\begin{table*}[t]
\centering
\small
\renewcommand{\arraystretch}{0.8}
\resizebox{\textwidth}{!}{
\begin{tabular}{%
p{1.2cm}  p{1.2cm}
cccccc c}
\toprule[1.5pt]

\multirow{2}{*}{\textbf{Dataset}}  & \multirow{2}{*}{\textbf{Method}}
& \multicolumn{6}{c}{\textbf{Similarity} $\uparrow$} & \multirow{2}{*}{\textbf{Diversity} $\downarrow$} \\
\cmidrule(lr){3-8}
& &  Form. & Read. & Decl. & Impr. & Spell. & Gram. & \\

\midrule
\multicolumn{9}{c}{\textbf{QwQ-32B}} \\
\midrule

\multirow{4}{*}{\textbf{SQuAD}} & WR & \Sim{0.566} & \Sim{0.576} & \Sim{0.602} & \textbf{\Sim{0.634}} & \Sim{0.549} & \Sim{0.593} & \textbf{\Div{0.438}} \\
& NR & \Sim{0.511} & \Sim{0.570} & \Sim{0.625} & \textbf{\Sim{0.649}} & \Sim{0.488} & \Sim{0.599} & \textbf{\Div{0.475}} \\
& LB & \Sim{0.573} & \Sim{0.579} & \Sim{0.615} & \textbf{\Sim{0.631}} & \Sim{0.544} & \Sim{0.584} & \textbf{\Div{0.440}} \\
& GB & \Sim{0.552} & \Sim{0.558} & \Sim{0.579} & \textbf{\Sim{0.589}} & \Sim{0.552} & \Sim{0.579} & \textbf{\Div{0.459}} \\
\cmidrule(lr){1-9}
\multirow{4}{*}{\textbf{NQ}} & WR & \Sim{0.660} & \Sim{0.634} & \Sim{0.671} & \textbf{\Sim{0.678}} & \Sim{0.583} & \Sim{0.673} & \textbf{\Div{0.373}} \\
& NR & \Sim{0.367} & \Sim{0.518} & \Sim{0.375} & \textbf{\Sim{0.520}} & \Sim{0.483} & \Sim{0.389} & \textbf{\Div{0.566}} \\
& LB & \Sim{0.344} & \Sim{0.304} & \Sim{0.321} & \Sim{0.309} & \Sim{0.305} & \textbf{\Sim{0.353}} & \textbf{\Div{0.662}} \\
& GB & \Sim{0.438} & \Sim{0.769} & \Sim{0.783} & \textbf{\Sim{0.797}} & \Sim{0.739} & \Sim{0.442} & \textbf{\Div{0.428}} \\
\cmidrule(lr){1-9}
\multirow{4}{*}{\textbf{TriviaQA}} & WR & \Sim{0.743} & \Sim{0.735} & \textbf{\Sim{0.759}} & \Sim{0.759} & \Sim{0.699} & \Sim{0.752} & \textbf{\Div{0.288}} \\
& NR & \Sim{0.698} & \Sim{0.738} & \Sim{0.735} & \textbf{\Sim{0.759}} & \Sim{0.671} & \Sim{0.747} & \textbf{\Div{0.309}} \\
& LB & \Sim{0.770} & \Sim{0.775} & \Sim{0.785} & \Sim{0.779} & \Sim{0.730} & \textbf{\Sim{0.788}} & \textbf{\Div{0.255}} \\
& GB & \Sim{0.608} & \Sim{0.588} & \Sim{0.603} & \textbf{\Sim{0.626}} & \Sim{0.580} & \Sim{0.604} & \textbf{\Div{0.395}} \\
\cmidrule(lr){1-9}
\multirow{4}{*}{\textbf{2WIKI}} & WR & \Sim{0.528} & \Sim{0.554} & \Sim{0.589} & \textbf{\Sim{0.591}} & \Sim{0.481} & \Sim{0.483} & \textbf{\Div{0.531}} \\
& NR & \Sim{0.443} & \Sim{0.500} & \Sim{0.520} & \textbf{\Sim{0.531}} & \Sim{0.422} & \Sim{0.514} & \textbf{\Div{0.547}} \\
& LB & \Sim{0.500} & \Sim{0.545} & \Sim{0.551} & \textbf{\Sim{0.565}} & \Sim{0.462} & \Sim{0.519} & \textbf{\Div{0.511}} \\
& GB & \Sim{0.520} & \Sim{0.557} & \Sim{0.554} & \textbf{\Sim{0.559}} & \Sim{0.499} & \Sim{0.548} & \textbf{\Div{0.465}} \\
\cmidrule(lr){1-9}
\multirow{4}{*}{\textbf{HotpotQA}} & WR & \Sim{0.359} & \textbf{\Sim{0.402}} & \Sim{0.385} & \Sim{0.401} & \Sim{0.333} & \Sim{0.394} & \textbf{\Div{0.639}} \\
& NR & \Sim{0.377} & \textbf{\Sim{0.414}} & \Sim{0.408} & \Sim{0.401} & \Sim{0.357} & \Sim{0.397} & \textbf{\Div{0.624}} \\
& LB & \Sim{0.444} & \Sim{0.452} & \textbf{\Sim{0.468}} & \Sim{0.458} & \Sim{0.406} & \Sim{0.444} & \textbf{\Div{0.568}} \\
& GB & \Sim{0.582} & \Sim{0.585} & \Sim{0.581} & \textbf{\Sim{0.608}} & \Sim{0.563} & \Sim{0.580} & \textbf{\Div{0.429}} \\
\cmidrule(lr){1-9}
\multirow{4}{*}{\textbf{MuSiQue}} & WR & \Sim{0.394} & \Sim{0.390} & \textbf{\Sim{0.408}} & \Sim{0.405} & \Sim{0.371} & \Sim{0.408} & \textbf{\Div{0.629}} \\
& NR & \Sim{0.387} & \Sim{0.377} & \Sim{0.396} & \textbf{\Sim{0.405}} & \Sim{0.374} & \Sim{0.383} & \textbf{\Div{0.634}} \\
& LB & \Sim{0.395} & \Sim{0.400} & \Sim{0.414} & \textbf{\Sim{0.422}} & \Sim{0.380} & \Sim{0.404} & \textbf{\Div{0.610}} \\
& GB & \Sim{0.593} & \textbf{\Sim{0.607}} & \Sim{0.606} & \Sim{0.607} & \Sim{0.589} & \Sim{0.582} & \textbf{\Div{0.422}} \\

\bottomrule[1.5pt]
\end{tabular}
}
\vspace{-0.1cm}
\caption{
\textbf{Answer similarity to the original query and diversity across perturbation types (excluding human rewrites).} We compare w/o retrieval (WR), naive RAG (NR), logit-based (LB), and generation-based (GB).
Darker shading indicates more desirable outcomes: higher Similarity and lower Diversity.}
\label{tab:robustness_simdiv_nohuman_qwq}
\end{table*}

%% file: Tables/robustness_accuracy_human_full_llama_appendix.tex
\begin{table*}[t]
\centering
\small
\resizebox{\textwidth}{!}{
\begin{tabular}{%
p{1.2cm}  p{1.2cm}
ccccccc c}
\toprule[1.5pt]

\multirow{2}{*}{\textbf{Dataset}}  & \multirow{2}{*}{\textbf{Method}}
& \multicolumn{7}{c}{\textbf{Similarity} $\uparrow$} & \multirow{2}{*}{\textbf{Diversity} $\downarrow$} \\
\cmidrule(lr){3-9}
& & Human & Form. & Read. & Decl. & Impr. & Spell. & Gram. & \\

\midrule
\multicolumn{10}{c}{\textbf{Llama-3.1-8B}} \\
\midrule

\multirow{4}{*}{\textbf{SQuAD}} & WR & \Sim{0.542} & \Sim{0.579} & \Sim{0.610} & \Sim{0.644} & \textbf{\Sim{0.653}} & \Sim{0.542} & \Sim{0.600} & \textbf{\Div{0.446}} \\
& NR & \Sim{0.505} & \Sim{0.580} & \Sim{0.544} & \Sim{0.613} & \textbf{\Sim{0.622}} & \Sim{0.479} & \Sim{0.607} & \textbf{\Div{0.503}} \\
& LB & \Sim{0.527} & \Sim{0.557} & \Sim{0.557} & \Sim{0.599} & \textbf{\Sim{0.626}} & \Sim{0.566} & \Sim{0.584} & \textbf{\Div{0.469}} \\
& GB & \Sim{0.447} & \Sim{0.416} & \Sim{0.421} & \Sim{0.423} & \textbf{\Sim{0.449}} & \Sim{0.427} & \Sim{0.449} & \textbf{\Div{0.583}} \\
\cmidrule(lr){1-10}
\multirow{4}{*}{\textbf{NQ}} & WR & \Sim{0.624} & \Sim{0.698} & \Sim{0.646} & \Sim{0.683} & \textbf{\Sim{0.748}} & \Sim{0.586} & \Sim{0.692} & \textbf{\Div{0.385}} \\
& NR & \Sim{0.292} & \Sim{0.300} & \Sim{0.470} & \textbf{\Sim{0.507}} & \Sim{0.500} & \Sim{0.385} & \Sim{0.319} & \textbf{\Div{0.616}} \\
& LB & \Sim{0.642} & \Sim{0.674} & \Sim{0.662} & \Sim{0.661} & \Sim{0.686} & \Sim{0.577} & \textbf{\Sim{0.691}} & \textbf{\Div{0.370}} \\
& GB & \Sim{0.356} & \Sim{0.382} & \textbf{\Sim{1.000}} & \textbf{\Sim{1.000}} & \textbf{\Sim{1.000}} & \textbf{\Sim{1.000}} & \Sim{0.435} & \textbf{\Div{0.416}} \\
\cmidrule(lr){1-10}
\multirow{4}{*}{\textbf{TriviaQA}} & WR & \Sim{0.660} & \Sim{0.734} & \Sim{0.753} & \Sim{0.741} & \textbf{\Sim{0.755}} & \Sim{0.672} & \Sim{0.700} & \textbf{\Div{0.331}} \\
& NR & \Sim{0.634} & \Sim{0.673} & \Sim{0.691} & \Sim{0.690} & \textbf{\Sim{0.743}} & \Sim{0.635} & \Sim{0.725} & \textbf{\Div{0.370}} \\
& LB & \Sim{0.697} & \Sim{0.735} & \textbf{\Sim{0.776}} & \Sim{0.718} & \Sim{0.729} & \Sim{0.708} & \Sim{0.735} & \textbf{\Div{0.314}} \\
& GB & \Sim{0.475} & \Sim{0.499} & \Sim{0.458} & \Sim{0.541} & \textbf{\Sim{0.555}} & \Sim{0.460} & \Sim{0.538} & \textbf{\Div{0.517}} \\
\cmidrule(lr){1-10}
\multirow{4}{*}{\textbf{2WIKI}} & WR & \Sim{0.540} & \Sim{0.516} & \Sim{0.567} & \Sim{0.628} & \textbf{\Sim{0.628}} & \Sim{0.511} & \Sim{0.607} & \textbf{\Div{0.480}} \\
& NR & \Sim{0.488} & \Sim{0.516} & \Sim{0.585} & \textbf{\Sim{0.641}} & \Sim{0.592} & \Sim{0.453} & \Sim{0.615} & \textbf{\Div{0.473}} \\
& LB & \Sim{0.550} & \Sim{0.541} & \textbf{\Sim{0.617}} & \Sim{0.616} & \Sim{0.610} & \Sim{0.502} & \Sim{0.590} & \textbf{\Div{0.451}} \\
& GB & \Sim{0.279} & \Sim{0.423} & \textbf{\Sim{0.458}} & \Sim{0.300} & \Sim{0.403} & \Sim{0.313} & \Sim{0.385} & \textbf{\Div{0.647}} \\
\cmidrule(lr){1-10}
\multirow{4}{*}{\textbf{HotpotQA}} & WR & \Sim{0.514} & \Sim{0.544} & \Sim{0.545} & \Sim{0.555} & \textbf{\Sim{0.592}} & \Sim{0.511} & \Sim{0.565} & \textbf{\Div{0.482}} \\
& NR & \Sim{0.524} & \Sim{0.568} & \Sim{0.578} & \Sim{0.574} & \Sim{0.583} & \Sim{0.520} & \textbf{\Sim{0.616}} & \textbf{\Div{0.465}} \\
& LB & \Sim{0.543} & \Sim{0.630} & \Sim{0.617} & \Sim{0.586} & \Sim{0.634} & \Sim{0.588} & \textbf{\Sim{0.652}} & \textbf{\Div{0.401}} \\
& GB & \Sim{0.462} & \Sim{0.477} & \Sim{0.484} & \Sim{0.452} & \Sim{0.465} & \Sim{0.489} & \textbf{\Sim{0.515}} & \textbf{\Div{0.507}} \\
\cmidrule(lr){1-10}
\multirow{4}{*}{\textbf{MuSiQue}} & WR & \Sim{0.459} & \Sim{0.417} & \Sim{0.511} & \Sim{0.471} & \textbf{\Sim{0.525}} & \Sim{0.403} & \Sim{0.454} & \textbf{\Div{0.564}} \\
& NR & \Sim{0.407} & \Sim{0.415} & \textbf{\Sim{0.476}} & \Sim{0.433} & \Sim{0.452} & \Sim{0.382} & \Sim{0.400} & \textbf{\Div{0.600}} \\
& LB & \Sim{0.504} & \Sim{0.528} & \textbf{\Sim{0.578}} & \Sim{0.486} & \Sim{0.576} & \Sim{0.436} & \Sim{0.487} & \textbf{\Div{0.551}} \\
& GB & \Sim{0.360} & \Sim{0.423} & \textbf{\Sim{0.442}} & \Sim{0.380} & \Sim{0.415} & \Sim{0.372} & \Sim{0.370} & \textbf{\Div{0.590}} \\

\bottomrule[1.5pt]
\end{tabular}
}
\caption{
\textbf{Answer similarity to the original query and diversity across perturbation types (including human rewrites).} We compare w/o retrieval (WR), naive RAG (NR), logit-based (LB), and generation-based (GB).
Darker shading indicates more desirable outcomes: higher Similarity and lower Diversity.}
\label{tab:robustness_simdiv_withhuman_llama}
\end{table*}

%% file: Tables/robustness_accuracy_human_full_qwq_appendix.tex
\begin{table*}[t]
\centering
\small
\resizebox{\textwidth}{!}{
\begin{tabular}{%
p{1.2cm}  p{1.2cm}
ccccccc c}
\toprule[1.5pt]

\multirow{2}{*}{\textbf{Dataset}}  & \multirow{2}{*}{\textbf{Method}}
& \multicolumn{7}{c}{\textbf{Similarity} $\uparrow$} & \multirow{2}{*}{\textbf{Diversity} $\downarrow$} \\
\cmidrule(lr){3-9}
& & Human & Form. & Read. & Decl. & Impr. & Spell. & Gram. & \\

\midrule
\multicolumn{10}{c}{\textbf{QwQ-32B}} \\
\midrule

\multirow{4}{*}{\textbf{SQuAD}} & WR & \Sim{0.585} & \Sim{0.567} & \Sim{0.608} & \Sim{0.630} & \textbf{\Sim{0.642}} & \Sim{0.583} & \Sim{0.605} & \textbf{\Div{0.448}} \\
& NR & \Sim{0.556} & \Sim{0.514} & \Sim{0.619} & \Sim{0.655} & \textbf{\Sim{0.664}} & \Sim{0.518} & \Sim{0.584} & \textbf{\Div{0.459}} \\
& LB & \Sim{0.565} & \Sim{0.608} & \Sim{0.590} & \Sim{0.608} & \textbf{\Sim{0.635}} & \Sim{0.577} & \Sim{0.599} & \textbf{\Div{0.435}} \\
& GB & \Sim{0.544} & \Sim{0.549} & \Sim{0.560} & \Sim{0.596} & \textbf{\Sim{0.601}} & \Sim{0.575} & \Sim{0.588} & \textbf{\Div{0.461}} \\
\cmidrule(lr){1-10}
\multirow{4}{*}{\textbf{NQ}} & WR & \Sim{0.594} & \Sim{0.631} & \Sim{0.612} & \Sim{0.649} & \Sim{0.635} & \Sim{0.573} & \textbf{\Sim{0.660}} & \textbf{\Div{0.374}} \\
& NR & \Sim{0.370} & \Sim{0.358} & \textbf{\Sim{0.517}} & \Sim{0.374} & \Sim{0.510} & \Sim{0.461} & \Sim{0.374} & \textbf{\Div{0.551}} \\
& LB & \Sim{0.330} & \Sim{0.348} & \Sim{0.294} & \Sim{0.319} & \Sim{0.346} & \Sim{0.300} & \textbf{\Sim{0.387}} & \textbf{\Div{0.628}} \\
& GB & \Sim{0.438} & \Sim{0.438} & \Sim{0.769} & \Sim{0.783} & \textbf{\Sim{0.797}} & \Sim{0.739} & \Sim{0.442} & \textbf{\Div{0.454}} \\
\cmidrule(lr){1-10}
\multirow{4}{*}{\textbf{TriviaQA}} & WR & \Sim{0.684} & \textbf{\Sim{0.718}} & \Sim{0.678} & \Sim{0.696} & \Sim{0.705} & \Sim{0.661} & \Sim{0.687} & \textbf{\Div{0.336}} \\
& NR & \Sim{0.646} & \Sim{0.649} & \Sim{0.683} & \Sim{0.687} & \textbf{\Sim{0.720}} & \Sim{0.597} & \Sim{0.664} & \textbf{\Div{0.363}} \\
& LB & \Sim{0.697} & \Sim{0.730} & \Sim{0.712} & \textbf{\Sim{0.745}} & \Sim{0.693} & \Sim{0.665} & \Sim{0.717} & \textbf{\Div{0.307}} \\
& GB & \Sim{0.548} & \Sim{0.551} & \Sim{0.531} & \Sim{0.539} & \textbf{\Sim{0.582}} & \Sim{0.546} & \Sim{0.558} & \textbf{\Div{0.440}} \\
\cmidrule(lr){1-10}
\multirow{4}{*}{\textbf{2WIKI}} & WR & \Sim{0.560} & \Sim{0.542} & \Sim{0.556} & \Sim{0.623} & \textbf{\Sim{0.634}} & \Sim{0.519} & \Sim{0.570} & \textbf{\Div{0.496}} \\
& NR & \Sim{0.511} & \Sim{0.492} & \Sim{0.545} & \textbf{\Sim{0.579}} & \Sim{0.569} & \Sim{0.482} & \Sim{0.551} & \textbf{\Div{0.516}} \\
& LB & \Sim{0.518} & \Sim{0.528} & \Sim{0.570} & \Sim{0.573} & \textbf{\Sim{0.612}} & \Sim{0.522} & \Sim{0.572} & \textbf{\Div{0.486}} \\
& GB & \Sim{0.529} & \Sim{0.542} & \Sim{0.534} & \Sim{0.568} & \textbf{\Sim{0.598}} & \Sim{0.520} & \Sim{0.558} & \textbf{\Div{0.463}} \\
\cmidrule(lr){1-10}
\multirow{4}{*}{\textbf{HotpotQA}} & WR & \Sim{0.356} & \Sim{0.394} & \textbf{\Sim{0.455}} & \Sim{0.412} & \Sim{0.416} & \Sim{0.368} & \Sim{0.421} & \textbf{\Div{0.636}} \\
& NR & \Sim{0.379} & \Sim{0.366} & \Sim{0.405} & \Sim{0.398} & \textbf{\Sim{0.412}} & \Sim{0.364} & \Sim{0.381} & \textbf{\Div{0.621}} \\
& LB & \Sim{0.420} & \Sim{0.467} & \Sim{0.417} & \textbf{\Sim{0.486}} & \Sim{0.476} & \Sim{0.428} & \Sim{0.468} & \textbf{\Div{0.560}} \\
& GB & \Sim{0.616} & \Sim{0.595} & \Sim{0.596} & \Sim{0.598} & \textbf{\Sim{0.629}} & \Sim{0.573} & \Sim{0.612} & \textbf{\Div{0.426}} \\
\cmidrule(lr){1-10}
\multirow{4}{*}{\textbf{MuSiQue}} & WR & \Sim{0.395} & \Sim{0.384} & \Sim{0.388} & \textbf{\Sim{0.411}} & \Sim{0.402} & \Sim{0.372} & \Sim{0.395} & \textbf{\Div{0.633}} \\
& NR & \Sim{0.363} & \textbf{\Sim{0.427}} & \Sim{0.396} & \Sim{0.425} & \Sim{0.404} & \Sim{0.354} & \Sim{0.382} & \textbf{\Div{0.627}} \\
& LB & \Sim{0.397} & \textbf{\Sim{0.416}} & \Sim{0.371} & \Sim{0.403} & \Sim{0.408} & \Sim{0.389} & \Sim{0.390} & \textbf{\Div{0.619}} \\
& GB & \Sim{0.584} & \Sim{0.582} & \Sim{0.604} & \textbf{\Sim{0.627}} & \Sim{0.603} & \Sim{0.595} & \Sim{0.573} & \textbf{\Div{0.424}} \\

\bottomrule[1.5pt]
\end{tabular}
}
\vspace{-0.1cm}
\caption{
\textbf{Answer similarity to the original query and diversity across perturbation types (including human rewrites).} We compare w/o retrieval (WR), naive RAG (NR), logit-based (LB), and generation-based (GB).
Darker shading indicates more desirable outcomes: higher Similarity and lower Diversity.}
\label{tab:robustness_simdiv_withhuman_qwq}
\end{table*}

%% file: Tables/Robustness_efficiency_full_llama_appendix.tex
\begin{table*}[t]
\centering
\small
\renewcommand{\arraystretch}{0.8}
\resizebox{\textwidth}{!}{
\begin{tabular}{%
p{1.7cm} p{1.5cm} p{1cm}
ccccccc}
\toprule[1.5pt]

\multirow{2}{*}{\textbf{Dataset}} & \multirow{2}{*}{\textbf{Target}} & \multirow{2}{*}{\textbf{Method}}
& \multicolumn{6}{c}{\textbf{RE} $\downarrow$} & \multirow{2}{*}{\textbf{CVR} $\uparrow$} \\
\cmidrule(lr){4-9}
& & & Form. & Read. & Decl. & Impr. & Spell. & Gram. & \\

\midrule
\multicolumn{10}{c}{\textbf{Llama-3.1-8B}} \\
\midrule

\multirow{4}{*}{\textbf{SQuAD}} & \multirow{2}{*}{\textbf{Retriever}} & LB & \RE{0.472} & \RE{0.592} & \RE{0.416} & \RE{0.338} & \RE{0.431} & \RE{0.439} & \CVR{0.734} \\
&  & GB & \textbf{\RE{0.092}} & \textbf{\RE{0.094}} & \textbf{\RE{0.086}} & \textbf{\RE{0.090}} & \textbf{\RE{0.100}} & \textbf{\RE{0.092}} & \CVR{0.937} \\
& \multirow{2}{*}{\textbf{LLM}} & LB & \RE{0.441} & \RE{0.592} & \RE{0.396} & \RE{0.313} & \RE{0.416} & \RE{0.411} & \CVR{0.767} \\
&  & GB & \textbf{\RE{0.056}} & \textbf{\RE{0.059}} & \textbf{\RE{0.048}} & \textbf{\RE{0.055}} & \textbf{\RE{0.061}} & \textbf{\RE{0.051}} & \CVR{0.963} \\
\cmidrule(lr){1-10}
\multirow{4}{*}{\textbf{NQ}} & \multirow{2}{*}{\textbf{Retriever}} & LB & \RE{0.520} & \RE{0.485} & \RE{0.576} & \RE{0.465} & \RE{0.578} & \RE{0.456} & \CVR{0.702} \\
&  & GB & \textbf{\RE{0.096}} & \textbf{\RE{0.000}} & \textbf{\RE{0.000}} & \textbf{\RE{0.000}} & \textbf{\RE{0.000}} & \textbf{\RE{0.096}} & \CVR{0.949} \\
& \multirow{2}{*}{\textbf{LLM}} & LB & \RE{0.499} & \RE{0.462} & \RE{0.541} & \RE{0.428} & \RE{0.545} & \RE{0.440} & \CVR{0.750} \\
&  & GB & \textbf{\RE{0.038}} & \textbf{\RE{0.000}} & \textbf{\RE{0.000}} & \textbf{\RE{0.000}} & \textbf{\RE{0.000}} & \textbf{\RE{0.042}} & \CVR{0.973} \\
\cmidrule(lr){1-10}
\multirow{4}{*}{\textbf{TriviaQA}} & \multirow{2}{*}{\textbf{Retriever}} & LB & \RE{0.452} & \RE{0.442} & \RE{0.493} & \RE{0.436} & \RE{0.536} & \RE{0.471} & \CVR{0.712} \\
&  & GB & \textbf{\RE{0.137}} & \textbf{\RE{0.120}} & \textbf{\RE{0.121}} & \textbf{\RE{0.122}} & \textbf{\RE{0.124}} & \textbf{\RE{0.122}} & \CVR{0.917} \\
& \multirow{2}{*}{\textbf{LLM}} & LB & \RE{0.463} & \RE{0.420} & \RE{0.472} & \RE{0.410} & \RE{0.527} & \RE{0.466} & \CVR{0.747} \\
&  & GB & \textbf{\RE{0.054}} & \textbf{\RE{0.047}} & \textbf{\RE{0.048}} & \textbf{\RE{0.046}} & \textbf{\RE{0.047}} & \textbf{\RE{0.047}} & \CVR{0.956} \\
\cmidrule(lr){1-10}
\multirow{4}{*}{\textbf{2WIKI}} & \multirow{2}{*}{\textbf{Retriever}} & LB & \RE{0.853} & \RE{0.859} & \RE{0.455} & \RE{0.846} & \RE{0.928} & \RE{0.941} & \CVR{0.710} \\
&  & GB & \textbf{\RE{0.468}} & \textbf{\RE{0.416}} & \textbf{\RE{0.402}} & \textbf{\RE{0.542}} & \textbf{\RE{0.496}} & \textbf{\RE{0.435}} & \CVR{0.742} \\
& \multirow{2}{*}{\textbf{LLM}} & LB & \RE{1.001} & \RE{1.001} & \RE{0.427} & \RE{1.007} & \RE{1.093} & \RE{1.130} & \CVR{0.742} \\
&  & GB & \textbf{\RE{0.447}} & \textbf{\RE{0.404}} & \textbf{\RE{0.401}} & \textbf{\RE{0.520}} & \textbf{\RE{0.477}} & \textbf{\RE{0.418}} & \CVR{0.745} \\
\cmidrule(lr){1-10}
\multirow{4}{*}{\textbf{HotpotQA}} & \multirow{2}{*}{\textbf{Retriever}} & LB & \textbf{\RE{0.486}} & \textbf{\RE{0.461}} & \textbf{\RE{0.428}} & \textbf{\RE{0.440}} & \textbf{\RE{0.487}} & \textbf{\RE{0.428}} & \CVR{0.747} \\
&  & GB & \RE{1.914} & \RE{1.992} & \RE{1.850} & \RE{1.848} & \RE{1.924} & \RE{1.825} & \CVR{0.687} \\
& \multirow{2}{*}{\textbf{LLM}} & LB & \textbf{\RE{0.490}} & \textbf{\RE{0.439}} & \textbf{\RE{0.421}} & \textbf{\RE{0.427}} & \textbf{\RE{0.502}} & \textbf{\RE{0.418}} & \CVR{0.761} \\
&  & GB & \RE{1.560} & \RE{1.626} & \RE{1.525} & \RE{1.517} & \RE{1.565} & \RE{1.495} & \CVR{0.694} \\
\cmidrule(lr){1-10}
\multirow{4}{*}{\textbf{MuSiQue}} & \multirow{2}{*}{\textbf{Retriever}} & LB & \textbf{\RE{0.404}} & \textbf{\RE{0.417}} & \textbf{\RE{0.382}} & \textbf{\RE{0.345}} & \textbf{\RE{0.399}} & \textbf{\RE{0.340}} & \CVR{0.753} \\
&  & GB & \RE{0.521} & \RE{0.525} & \RE{0.533} & \RE{0.446} & \RE{0.574} & \RE{0.510} & \CVR{0.691} \\
& \multirow{2}{*}{\textbf{LLM}} & LB & \textbf{\RE{0.365}} & \textbf{\RE{0.398}} & \textbf{\RE{0.353}} & \textbf{\RE{0.323}} & \textbf{\RE{0.388}} & \textbf{\RE{0.318}} & \CVR{0.763} \\
&  & GB & \RE{0.502} & \RE{0.502} & \RE{0.519} & \RE{0.429} & \RE{0.550} & \RE{0.484} & \CVR{0.698} \\
\bottomrule[1.5pt]
\end{tabular}
}
\caption{
\textbf{Computation robustness across query variations (Llama-3.1-8B; no human rewrites).}
Darker shading indicates more desirable outcomes: lower RE and higher CVR.}
\label{tab:robustness_efficiency_llama_nohuman}
\end{table*}

%% file: Tables/Robustness_efficiency_full_qwq_appendix.tex
\begin{table*}[t]
\centering
\small

\resizebox{\textwidth}{!}{
\begin{tabular}{%
p{2cm} p{1.5cm} p{1cm}
ccccccc}
\toprule[1.5pt]

\multirow{2}{*}{\textbf{Dataset}} & \multirow{2}{*}{\textbf{Target}} & \multirow{2}{*}{\textbf{Method}}
& \multicolumn{6}{c}{\textbf{RE} $\downarrow$} & \multirow{2}{*}{\textbf{CVR} $\uparrow$} \\
\cmidrule(lr){4-9}
& & & Form. & Read. & Decl. & Impr. & Spell. & Gram. & \\

\midrule
\multicolumn{10}{c}{\textbf{QwQ-32B}} \\
\midrule

\multirow{4}{*}{\textbf{SQuAD}} & \multirow{2}{*}{\textbf{Retriever}} & LB & \RE{0.494} & \RE{0.450} & \RE{0.405} & \RE{0.415} & \RE{0.495} & \RE{0.441} & \CVR{0.740} \\
&  & GB & \textbf{\RE{0.360}} & \textbf{\RE{0.322}} & \textbf{\RE{0.298}} & \textbf{\RE{0.315}} & \textbf{\RE{0.361}} & \textbf{\RE{0.330}} & \CVR{0.762} \\
& \multirow{2}{*}{\textbf{LLM}} & LB & \RE{0.477} & \RE{0.466} & \RE{0.389} & \RE{0.399} & \RE{0.491} & \RE{0.424} & \CVR{0.760} \\
&  & GB & \textbf{\RE{0.265}} & \textbf{\RE{0.230}} & \textbf{\RE{0.225}} & \textbf{\RE{0.238}} & \textbf{\RE{0.250}} & \textbf{\RE{0.243}} & \CVR{0.858} \\
\cmidrule(lr){1-10}
\multirow{4}{*}{\textbf{NQ}} & \multirow{2}{*}{\textbf{Retriever}} & LB & \RE{0.551} & \RE{0.586} & \RE{0.525} & \RE{0.513} & \RE{0.673} & \RE{0.520} & \CVR{0.689} \\
&  & GB & \textbf{\RE{0.411}} & \textbf{\RE{0.402}} & \textbf{\RE{0.374}} & \textbf{\RE{0.416}} & \textbf{\RE{0.439}} & \textbf{\RE{0.452}} & \CVR{0.721} \\
& \multirow{2}{*}{\textbf{LLM}} & LB & \RE{0.546} & \RE{0.598} & \RE{0.511} & \RE{0.504} & \RE{0.679} & \RE{0.516} & \CVR{0.724} \\
&  & GB & \textbf{\RE{0.379}} & \textbf{\RE{0.375}} & \textbf{\RE{0.358}} & \textbf{\RE{0.372}} & \textbf{\RE{0.411}} & \textbf{\RE{0.424}} & \CVR{0.820} \\
\cmidrule(lr){1-10}
\multirow{4}{*}{\textbf{TriviaQA}} & \multirow{2}{*}{\textbf{Retriever}} & LB & \RE{0.581} & \RE{0.563} & \RE{0.495} & \RE{0.504} & \RE{0.587} & \RE{0.546} & \CVR{0.697} \\
&  & GB & \textbf{\RE{0.362}} & \textbf{\RE{0.377}} & \textbf{\RE{0.347}} & \textbf{\RE{0.374}} & \textbf{\RE{0.349}} & \textbf{\RE{0.387}} & \CVR{0.721} \\
& \multirow{2}{*}{\textbf{LLM}} & LB & \RE{0.564} & \RE{0.565} & \RE{0.470} & \RE{0.509} & \RE{0.601} & \RE{0.530} & \CVR{0.727} \\
&  & GB & \textbf{\RE{0.329}} & \textbf{\RE{0.373}} & \textbf{\RE{0.331}} & \textbf{\RE{0.348}} & \textbf{\RE{0.324}} & \textbf{\RE{0.348}} & \CVR{0.818} \\
\cmidrule(lr){1-10}
\multirow{4}{*}{\textbf{2WIKI}} & \multirow{2}{*}{\textbf{Retriever}} & LB & \textbf{\RE{0.560}} & \textbf{\RE{0.502}} & \textbf{\RE{0.446}} & \textbf{\RE{0.443}} & \textbf{\RE{0.565}} & \textbf{\RE{0.511}} & \CVR{0.719} \\
&  & GB & \RE{0.788} & \RE{0.808} & \RE{0.780} & \RE{0.825} & \RE{0.787} & \RE{0.884} & \CVR{0.689} \\
& \multirow{2}{*}{\textbf{LLM}} & LB & \textbf{\RE{0.547}} & \textbf{\RE{0.481}} & \textbf{\RE{0.426}} & \textbf{\RE{0.430}} & \textbf{\RE{0.565}} & \textbf{\RE{0.493}} & \CVR{0.747} \\
&  & GB & \RE{1.078} & \RE{1.142} & \RE{1.092} & \RE{1.203} & \RE{1.049} & \RE{1.219} & \CVR{0.716} \\
\cmidrule(lr){1-10}
\multirow{4}{*}{\textbf{HotpotQA}} & \multirow{2}{*}{\textbf{Retriever}} & LB & \textbf{\RE{0.589}} & \textbf{\RE{0.620}} & \textbf{\RE{0.580}} & \textbf{\RE{0.590}} & \textbf{\RE{0.671}} & \textbf{\RE{0.613}} & \CVR{0.707} \\
&  & GB & \RE{0.806} & \RE{0.774} & \RE{0.788} & \RE{0.818} & \RE{0.894} & \RE{0.663} & \CVR{0.654} \\
& \multirow{2}{*}{\textbf{LLM}} & LB & \textbf{\RE{0.516}} & \textbf{\RE{0.520}} & \textbf{\RE{0.487}} & \textbf{\RE{0.504}} & \textbf{\RE{0.605}} & \textbf{\RE{0.505}} & \CVR{0.743} \\
&  & GB & \RE{0.968} & \RE{0.920} & \RE{0.902} & \RE{1.000} & \RE{1.078} & \RE{0.778} & \CVR{0.695} \\
\cmidrule(lr){1-10}
\multirow{4}{*}{\textbf{MuSiQue}} & \multirow{2}{*}{\textbf{Retriever}} & LB & \textbf{\RE{0.632}} & \textbf{\RE{0.660}} & \textbf{\RE{0.598}} & \textbf{\RE{0.607}} & \textbf{\RE{0.702}} & \textbf{\RE{0.603}} & \CVR{0.711} \\
&  & GB & \RE{0.811} & \RE{0.890} & \RE{0.767} & \RE{0.883} & \RE{0.845} & \RE{0.881} & \CVR{0.644} \\
& \multirow{2}{*}{\textbf{LLM}} & LB & \textbf{\RE{0.538}} & \textbf{\RE{0.572}} & \textbf{\RE{0.537}} & \textbf{\RE{0.515}} & \textbf{\RE{0.607}} & \textbf{\RE{0.523}} & \CVR{0.740} \\
&  & GB & \RE{1.032} & \RE{1.168} & \RE{0.984} & \RE{1.203} & \RE{1.100} & \RE{1.117} & \CVR{0.679} \\
\bottomrule[1.5pt]
\end{tabular}
}
\vspace{-0.1cm}
\caption{
\textbf{Computation robustness across query variations (QwQ-32B; no human rewrites).}
Darker shading indicates more desirable outcomes: lower RE and higher CVR.}
\label{tab:robustness_efficiency_qwq_nohuman}
\end{table*}

%% file: Tables/Robustness_efficiency_human_full_llama_appendix.tex
\begin{table*}[t]
\centering
\small
\resizebox{\textwidth}{!}{
\begin{tabular}{%
p{2cm} p{1.5cm} p{1cm}
cccccccc}
\toprule[1.5pt]

\multirow{2}{*}{\textbf{Dataset}} & \multirow{2}{*}{\textbf{Target}} & \multirow{2}{*}{\textbf{Method}}
& \multicolumn{7}{c}{\textbf{RE} $\downarrow$} & \multirow{2}{*}{\textbf{CVR} $\uparrow$} \\
\cmidrule(lr){4-10}
& & & Human & Form. & Read. & Decl. & Impr. & Spell. & Gram. & \\

\midrule
\multicolumn{11}{c}{\textbf{Llama-3.1-8B}} \\
\midrule

\multirow{4}{*}{\textbf{SQuAD}} & \multirow{2}{*}{\textbf{Retriever}} & LB & \RE{0.415} & \RE{0.452} & \RE{0.577} & \RE{0.423} & \RE{0.344} & \RE{0.460} & \RE{0.450} & \CVR{0.748} \\
&  & GB & \textbf{\RE{0.105}} & \textbf{\RE{0.090}} & \textbf{\RE{0.100}} & \textbf{\RE{0.105}} & \textbf{\RE{0.090}} & \textbf{\RE{0.120}} & \textbf{\RE{0.120}} & \CVR{0.946} \\
& \multirow{2}{*}{\textbf{LLM}} & LB & \RE{0.407} & \RE{0.418} & \RE{0.587} & \RE{0.398} & \RE{0.306} & \RE{0.414} & \RE{0.414} & \CVR{0.778} \\
&  & GB & \textbf{\RE{0.061}} & \textbf{\RE{0.055}} & \textbf{\RE{0.060}} & \textbf{\RE{0.065}} & \textbf{\RE{0.048}} & \textbf{\RE{0.068}} & \textbf{\RE{0.067}} & \CVR{0.969} \\
\cmidrule(lr){1-11}
\multirow{4}{*}{\textbf{NQ}} & \multirow{2}{*}{\textbf{Retriever}} & LB & \RE{0.549} & \RE{0.495} & \RE{0.487} & \RE{0.562} & \RE{0.481} & \RE{0.551} & \RE{0.425} & \CVR{0.696} \\
&  & GB & \textbf{\RE{0.077}} & \textbf{\RE{0.096}} & \textbf{\RE{0.000}} & \textbf{\RE{0.000}} & \textbf{\RE{0.000}} & \textbf{\RE{0.000}} & \textbf{\RE{0.096}} & \CVR{0.942} \\
& \multirow{2}{*}{\textbf{LLM}} & LB & \RE{0.542} & \RE{0.485} & \RE{0.475} & \RE{0.546} & \RE{0.410} & \RE{0.498} & \RE{0.427} & \CVR{0.739} \\
&  & GB & \textbf{\RE{0.029}} & \textbf{\RE{0.038}} & \textbf{\RE{0.000}} & \textbf{\RE{0.000}} & \textbf{\RE{0.000}} & \textbf{\RE{0.000}} & \textbf{\RE{0.042}} & \CVR{0.969} \\
\cmidrule(lr){1-11}
\multirow{4}{*}{\textbf{TriviaQA}} & \multirow{2}{*}{\textbf{Retriever}} & LB & \RE{0.492} & \RE{0.508} & \RE{0.457} & \RE{0.500} & \RE{0.397} & \RE{0.535} & \RE{0.472} & \CVR{0.709} \\
&  & GB & \textbf{\RE{0.065}} & \textbf{\RE{0.095}} & \textbf{\RE{0.105}} & \textbf{\RE{0.090}} & \textbf{\RE{0.080}} & \textbf{\RE{0.090}} & \textbf{\RE{0.095}} & \CVR{0.924} \\
& \multirow{2}{*}{\textbf{LLM}} & LB & \RE{0.496} & \RE{0.563} & \RE{0.466} & \RE{0.537} & \RE{0.378} & \RE{0.532} & \RE{0.468} & \CVR{0.747} \\
&  & GB & \textbf{\RE{0.024}} & \textbf{\RE{0.039}} & \textbf{\RE{0.044}} & \textbf{\RE{0.041}} & \textbf{\RE{0.030}} & \textbf{\RE{0.037}} & \textbf{\RE{0.041}} & \CVR{0.959} \\
\cmidrule(lr){1-11}
\multirow{4}{*}{\textbf{2WIKI}} & \multirow{2}{*}{\textbf{Retriever}} & LB & \RE{0.745} & \RE{0.862} & \RE{0.870} & \RE{0.510} & \RE{1.000} & \RE{0.890} & \RE{0.960} & \CVR{0.717} \\
&  & GB & \textbf{\RE{0.506}} & \textbf{\RE{0.468}} & \textbf{\RE{0.416}} & \textbf{\RE{0.402}} & \textbf{\RE{0.542}} & \textbf{\RE{0.496}} & \textbf{\RE{0.435}} & \CVR{0.740} \\
& \multirow{2}{*}{\textbf{LLM}} & LB & \RE{0.827} & \RE{1.065} & \RE{1.041} & \RE{0.526} & \RE{1.259} & \RE{1.126} & \RE{1.195} & \CVR{0.747} \\
&  & GB & \textbf{\RE{0.480}} & \textbf{\RE{0.447}} & \textbf{\RE{0.404}} & \textbf{\RE{0.401}} & \textbf{\RE{0.520}} & \textbf{\RE{0.477}} & \textbf{\RE{0.418}} & \CVR{0.743} \\
\cmidrule(lr){1-11}
\multirow{4}{*}{\textbf{HotpotQA}} & \multirow{2}{*}{\textbf{Retriever}} & LB & \textbf{\RE{0.426}} & \textbf{\RE{0.495}} & \textbf{\RE{0.429}} & \textbf{\RE{0.350}} & \textbf{\RE{0.413}} & \textbf{\RE{0.496}} & \textbf{\RE{0.415}} & \CVR{0.747} \\
&  & GB & \RE{1.962} & \RE{1.895} & \RE{1.928} & \RE{1.627} & \RE{1.697} & \RE{1.805} & \RE{1.832} & \CVR{0.684} \\
& \multirow{2}{*}{\textbf{LLM}} & LB & \textbf{\RE{0.426}} & \textbf{\RE{0.479}} & \textbf{\RE{0.396}} & \textbf{\RE{0.328}} & \textbf{\RE{0.381}} & \textbf{\RE{0.498}} & \textbf{\RE{0.401}} & \CVR{0.757} \\
&  & GB & \RE{1.668} & \RE{1.524} & \RE{1.556} & \RE{1.326} & \RE{1.389} & \RE{1.481} & \RE{1.527} & \CVR{0.693} \\
\cmidrule(lr){1-11}
\multirow{4}{*}{\textbf{MuSiQue}} & \multirow{2}{*}{\textbf{Retriever}} & LB & \textbf{\RE{0.284}} & \textbf{\RE{0.275}} & \textbf{\RE{0.377}} & \textbf{\RE{0.448}} & \textbf{\RE{0.343}} & \textbf{\RE{0.296}} & \textbf{\RE{0.219}} & \CVR{0.751} \\
&  & GB & \RE{0.597} & \RE{0.568} & \RE{0.544} & \RE{0.554} & \RE{0.438} & \RE{0.586} & \RE{0.585} & \CVR{0.704} \\
& \multirow{2}{*}{\textbf{LLM}} & LB & \textbf{\RE{0.268}} & \textbf{\RE{0.262}} & \textbf{\RE{0.331}} & \textbf{\RE{0.410}} & \textbf{\RE{0.307}} & \textbf{\RE{0.274}} & \textbf{\RE{0.221}} & \CVR{0.761} \\
&  & GB & \RE{0.528} & \RE{0.510} & \RE{0.504} & \RE{0.491} & \RE{0.415} & \RE{0.542} & \RE{0.505} & \CVR{0.710} \\

\bottomrule[1.5pt]
\end{tabular}
}
\vspace{-0.1cm}
\caption{
\textbf{Computation robustness across query variations (Llama-3.1-8B; including human rewrites).}
Darker shading indicates more desirable outcomes: lower RE and higher CVR.}
\label{tab:robustness_efficiency_llama}
\end{table*}

%% file: Tables/Robustness_efficiency_human_full_qwq_appendix.tex
\begin{table*}[t]
\centering
\small
\resizebox{\textwidth}{!}{
\begin{tabular}{%
p{2cm} p{1.5cm} p{1cm}
cccccccc}
\toprule[1.5pt]

\multirow{2}{*}{\textbf{Dataset}} & \multirow{2}{*}{\textbf{Target}} & \multirow{2}{*}{\textbf{Method}}
& \multicolumn{7}{c}{\textbf{RE} $\downarrow$} & \multirow{2}{*}{\textbf{CVR} $\uparrow$} \\
\cmidrule(lr){4-10}
& & & Human & Form. & Read. & Decl. & Impr. & Spell. & Gram. & \\

\midrule
\multicolumn{11}{c}{\textbf{QwQ-32B}} \\
\midrule

\multirow{4}{*}{\textbf{SQuAD}} & \multirow{2}{*}{\textbf{Retriever}} & LB & \RE{0.439} & \RE{0.500} & \RE{0.568} & \RE{0.446} & \RE{0.404} & \RE{0.451} & \RE{0.547} & \CVR{0.741} \\
&  & GB & \textbf{\RE{0.370}} & \textbf{\RE{0.295}} & \textbf{\RE{0.280}} & \textbf{\RE{0.310}} & \textbf{\RE{0.240}} & \textbf{\RE{0.335}} & \textbf{\RE{0.290}} & \CVR{0.766} \\
& \multirow{2}{*}{\textbf{LLM}} & LB & \RE{0.461} & \RE{0.468} & \RE{0.597} & \RE{0.426} & \RE{0.389} & \RE{0.417} & \RE{0.494} & \CVR{0.757} \\
&  & GB & \textbf{\RE{0.285}} & \textbf{\RE{0.252}} & \textbf{\RE{0.233}} & \textbf{\RE{0.242}} & \textbf{\RE{0.210}} & \textbf{\RE{0.273}} & \textbf{\RE{0.242}} & \CVR{0.859} \\
\cmidrule(lr){1-11}
\multirow{4}{*}{\textbf{NQ}} & \multirow{2}{*}{\textbf{Retriever}} & LB & \RE{0.565} & \RE{0.485} & \RE{0.462} & \RE{0.533} & \RE{0.558} & \RE{0.662} & \RE{0.537} & \CVR{0.692} \\
&  & GB & \textbf{\RE{0.404}} & \textbf{\RE{0.411}} & \textbf{\RE{0.402}} & \textbf{\RE{0.374}} & \textbf{\RE{0.416}} & \textbf{\RE{0.439}} & \textbf{\RE{0.452}} & \CVR{0.715} \\
& \multirow{2}{*}{\textbf{LLM}} & LB & \RE{0.554} & \RE{0.456} & \RE{0.492} & \RE{0.510} & \RE{0.555} & \RE{0.669} & \RE{0.533} & \CVR{0.725} \\
&  & GB & \textbf{\RE{0.370}} & \textbf{\RE{0.379}} & \textbf{\RE{0.375}} & \textbf{\RE{0.358}} & \textbf{\RE{0.372}} & \textbf{\RE{0.411}} & \textbf{\RE{0.424}} & \CVR{0.816} \\
\cmidrule(lr){1-11}
\multirow{4}{*}{\textbf{TriviaQA}} & \multirow{2}{*}{\textbf{Retriever}} & LB & \RE{0.488} & \RE{0.510} & \RE{0.438} & \RE{0.538} & \RE{0.514} & \RE{0.600} & \RE{0.557} & \CVR{0.700} \\
&  & GB & \textbf{\RE{0.455}} & \textbf{\RE{0.405}} & \textbf{\RE{0.420}} & \textbf{\RE{0.430}} & \textbf{\RE{0.395}} & \textbf{\RE{0.440}} & \textbf{\RE{0.440}} & \CVR{0.716} \\
& \multirow{2}{*}{\textbf{LLM}} & LB & \RE{0.473} & \RE{0.459} & \textbf{\RE{0.438}} & \RE{0.483} & \RE{0.530} & \RE{0.590} & \RE{0.486} & \CVR{0.727} \\
&  & GB & \textbf{\RE{0.472}} & \textbf{\RE{0.417}} & \RE{0.499} & \textbf{\RE{0.481}} & \textbf{\RE{0.407}} & \textbf{\RE{0.448}} & \textbf{\RE{0.448}} & \CVR{0.811} \\
\cmidrule(lr){1-11}
\multirow{4}{*}{\textbf{2WIKI}} & \multirow{2}{*}{\textbf{Retriever}} & LB & \textbf{\RE{0.407}} & \textbf{\RE{0.531}} & \textbf{\RE{0.486}} & \textbf{\RE{0.461}} & \textbf{\RE{0.437}} & \textbf{\RE{0.546}} & \textbf{\RE{0.510}} & \CVR{0.720} \\
&  & GB & \RE{0.746} & \RE{0.686} & \RE{0.643} & \RE{0.760} & \RE{0.757} & \RE{0.751} & \RE{0.794} & \CVR{0.679} \\
& \multirow{2}{*}{\textbf{LLM}} & LB & \textbf{\RE{0.437}} & \textbf{\RE{0.506}} & \textbf{\RE{0.480}} & \textbf{\RE{0.459}} & \textbf{\RE{0.425}} & \textbf{\RE{0.548}} & \textbf{\RE{0.502}} & \CVR{0.749} \\
&  & GB & \RE{0.911} & \RE{0.939} & \RE{0.942} & \RE{1.083} & \RE{1.130} & \RE{0.996} & \RE{1.135} & \CVR{0.708} \\
\cmidrule(lr){1-11}
\multirow{4}{*}{\textbf{HotpotQA}} & \multirow{2}{*}{\textbf{Retriever}} & LB & \textbf{\RE{0.500}} & \textbf{\RE{0.497}} & \textbf{\RE{0.552}} & \textbf{\RE{0.508}} & \textbf{\RE{0.451}} & \textbf{\RE{0.570}} & \textbf{\RE{0.465}} & \CVR{0.706} \\
&  & GB & \RE{1.023} & \RE{0.928} & \RE{0.712} & \RE{0.910} & \RE{0.901} & \RE{0.877} & \RE{0.716} & \CVR{0.632} \\
& \multirow{2}{*}{\textbf{LLM}} & LB & \textbf{\RE{0.446}} & \textbf{\RE{0.462}} & \textbf{\RE{0.448}} & \textbf{\RE{0.440}} & \textbf{\RE{0.411}} & \textbf{\RE{0.575}} & \textbf{\RE{0.392}} & \CVR{0.746} \\
&  & GB & \RE{1.146} & \RE{1.092} & \RE{0.776} & \RE{0.976} & \RE{1.105} & \RE{1.083} & \RE{0.874} & \CVR{0.678} \\
\cmidrule(lr){1-11}
\multirow{4}{*}{\textbf{MuSiQue}} & \multirow{2}{*}{\textbf{Retriever}} & LB & \textbf{\RE{0.652}} & \textbf{\RE{0.629}} & \textbf{\RE{0.657}} & \RE{0.602} & \RE{0.689} & \textbf{\RE{0.655}} & \textbf{\RE{0.602}} & \CVR{0.719} \\
&  & GB & \RE{0.770} & \RE{0.797} & \RE{0.767} & \textbf{\RE{0.548}} & \textbf{\RE{0.642}} & \RE{0.713} & \RE{0.709} & \CVR{0.645} \\
& \multirow{2}{*}{\textbf{LLM}} & LB & \textbf{\RE{0.557}} & \textbf{\RE{0.501}} & \textbf{\RE{0.537}} & \textbf{\RE{0.512}} & \textbf{\RE{0.579}} & \textbf{\RE{0.570}} & \textbf{\RE{0.510}} & \CVR{0.745} \\
&  & GB & \RE{0.888} & \RE{0.908} & \RE{0.916} & \RE{0.584} & \RE{0.718} & \RE{0.767} & \RE{0.719} & \CVR{0.679} \\
\bottomrule[1.5pt]
\end{tabular}
}
\vspace{-0.1cm}
\caption{
\textbf{Computation robustness across query variations (QwQ-32B; including human rewrites).}
Darker shading indicates more desirable outcomes: lower RE and higher CVR.}
\label{tab:robustness_efficiency_qwq}
\end{table*}